\documentclass[10pt,twocolumn,letterpaper]{article}

\usepackage[pagenumbers]{cvpr} 

\usepackage{graphicx}
\usepackage{amsmath}
\usepackage{amssymb}
\usepackage{booktabs}
\usepackage{algorithm}
\usepackage{algorithmic}
\usepackage{bbm}
\usepackage{bm}
\usepackage{amsmath}
\newtheorem{assumption}{Assumption}
\newtheorem{thm}{\bf Theorem}
\newtheorem{lemm}{\bf Lemma}[section]

\newtheorem{coro}{\bf Corollary}
\newenvironment{proof}{{\noindent\it Proof.}\quad}{\hfill\it \\ \rightline{End Proof.} \par}
\allowdisplaybreaks[4]
\usepackage{multirow}
\usepackage{booktabs}
\usepackage{float}
\usepackage{stfloats}

\usepackage[pagebackref,breaklinks,colorlinks]{hyperref}

\usepackage[capitalize]{cleveref}
\crefname{section}{Sec.}{Secs.}
\Crefname{section}{Section}{Sections}
\Crefname{table}{Table}{Tables}
\crefname{table}{Tab.}{Tabs.}

\def\cvprPaperID{10609} 
\def\confName{CVPR}
\def\confYear{2023}

\begin{document}

\title{GradMA: A Gradient-Memory-based Accelerated Federated Learning with Alleviated Catastrophic Forgetting}

\author{Kangyang Luo, Xiang Li\thanks{Corresponding author}, Yunshi Lan, Ming Gao\\
East China Normal University\\
Shanghai, China\\
{\tt\small 52205901003@stu.ecnu.edu.cn, \{xiangli, yslan, mgao\}@dase.ecnu.edu.cn}}
\maketitle

\begin{abstract}
Federated Learning~(FL) has emerged as a de facto machine learning area and received rapid increasing research interests from the community.
However, catastrophic forgetting caused by data heterogeneity and partial participation poses distinctive challenges for FL, which are detrimental to the performance.
To tackle the problems, 
we propose a new FL approach~(namely GradMA), which takes inspiration from continual learning to simultaneously correct the server-side and worker-side update directions as well as take full advantage of server's rich computing and memory resources. 
Furthermore, we elaborate a memory reduction strategy to enable GradMA to accommodate FL with a large scale of workers.
We then analyze convergence of GradMA theoretically under the smooth non-convex setting and show that its convergence rate achieves a linear speed up w.r.t the increasing number of sampled active workers. 
At last, our extensive experiments on various image classification tasks show that GradMA achieves significant performance gains in accuracy and communication efficiency compared to SOTA baselines. 
We provide our code here:
\href{https://github.com/lkyddd/GradMA}{https://github.com/lkyddd/GradMA}.

\end{abstract}

\section{Introduction}
\label{sec:intro}

Federated Learning~(FL)~\cite{McMahan2017Communication, li2020federated} is a privacy-preserving distributed machine learning scheme in which workers jointly participate in the collaborative training of a centralized model by sharing model information~(parameters or updates) rather than their private datasets. 
In recent years, FL has shown its potential to facilitate real-world applications, which falls broadly into two categories~\cite{kairouz2021advances}: the \textit{cross-silo} FL and the \textit{cross-device} FL. 
The \textit{cross-silo} FL corresponds to a relatively small number of reliable workers, usually organizations, such as healthcare facilities~\cite{jiang2022harmofl} and financial institutions~\cite{yang2019ffd}, etc.  %
In contrast, for the \textit{cross-device} FL, the number of workers can be very huge and unreliable, such as mobile devices~\cite{McMahan2017Communication}, IoT~\cite{nguyen2021federated} and autonomous driving cars~\cite{li2021privacy}, among others. 
In this paper, we focus on \textit{cross-device} FL.

The privacy-preserving and communication-efficient properties of the \textit{cross-device} FL make it promising, but it also confronts practical challenges arising from data heterogeneity~(i.e., non-iid data distribution across workers) and partial participation~\cite{li2019convergence, karimireddy2020scaffold, yang2021achieving, gu2021fast}. 
Specifically, the datasets held by real-world workers are generated locally according to their individual circumstances, resulting in the distribution of data on different workers being not identical.
Moreover, owing to the flexibility of worker participation in many scenarios~(e.g., IoT and mobile devices), workers can join or leave the FL system at will, thus making the set of active workers random and time-varying across communication rounds. 
Note that we consider a worker participates or is active at round $t$~(i.e., the index of the communication round) if it is able to complete the computation task and send back model information at the end of round $t$.

The above-mentioned challenges mainly bring catastrophic forgetting~(CF)~\cite{mccloskey1989catastrophic, shoham2019overcoming, xu2022acceleration} to FL. 
In a typical FL process, represented by FedAvg~\cite{McMahan2017Communication}, a server updates the centralized model by iteratively aggregating the model information from workers that generally is trained over several steps locally before being sent to the server. 
On the one hand, due to data heterogeneity, the model is updated on private data in local training, which is prone to overfit the current knowledge and forget the previous experience, thus leading to CF~\cite{huang2022learn}. 
In other words, the updates of the local models are prone to drift and diverge increasingly from the update of the centralized model~\cite{karimireddy2020scaffold}.
This can seriously deteriorate the performance of the centralized model. 
To ameliorate this issue, a variety of existing efforts regularize the objectives of the local models to align the centralized optimization objective~\cite{li2020federated1,  karimireddy2020scaffold, Acar2021Federated, li2021model, kim2022multi}. 
On the other hand, the server can only aggregate model information from active workers per communication round caused by partial participation.
In this case, many existing works directly discard~\cite{McMahan2017Communication, li2020federated, karimireddy2020scaffold, Acar2021Federated, Karimireddy2020Mime, yang2021achieving} or implicitly utilize~\cite{hsu2019measuring, reddi2020adaptive}, by means of momentum, the information provided by workers who have participated in the training but dropped out in the current communication round~(i.e., stragglers).
This results the centralized model, which tends to forget the experience of the stragglers, thus inducing CF.
In doing so, the convergence of popular FL approaches~(e.g., FedAvg) can be seriously slowed down by stragglers.
Moreover, all above approaches solely aggregate the collected information by averaging in the server, ignoring the server’s rich computing and memory resources that could be potentially harnessed to boost the performance of FL~\cite{zhang2022fine}.

In this paper, to alleviate CF caused by data heterogeneity and stragglers, we bring forward a new FL approach, dubbed as GradMA~(\underline{\textbf{Grad}}ient-\underline{\textbf{M}}emory-based \underline{\textbf{A}}ccelerated Federated Learning), which takes inspiration from continual learning~(CL)~\cite{yoon2019scalable, kirkpatrick2017overcoming, lopez2017gradient, farajtabar2020orthogonal, saha2021gradient} to simultaneously correct the server-side and worker-side update directions and fully utilize the rich computing and memory resources of the server.
Concretely, motivated by the success of GEM~\cite{lopez2017gradient} and  OGD~\cite{farajtabar2020orthogonal}, two
memory-based CL methods, we invoke quadratic programming~(QP) and memorize updates to correct the update directions. 
On the worker side, GradMA harnesses the gradients of the local model in the previous step and the centralized model, and the parameters difference between the local model in the current step and the centralized model as constraints of QP to adaptively correct the gradient of the local model.
Furthermore, 
we maintain a memory state to memorize accumulated update of each worker on the server side. 
GradMA then explicitly takes the memory state to constrain QP to augment the momentum~(i.e., the update direction) of the centralized model. 
Here, we need the server to allocate memory space to store memory state.
However, it may be not feasible in FL scenarios with a large size of workers, which can increase the storage cost and the burden of computing QP largely.
Therefore, we carefully craft a memory reduction strategy to alleviate the said limitations.
In addition, we theoretically analyze the convergence of GradMA in the smooth non-convex setting.

To sum up, we highlight our contributions as follows:
\begin{itemize}
    \item  We formulate a novel FL approach GradMA, which aims to simultaneously correct the server-side and worker-side update directions and fully harness the server's rich computing and memory resources. Meanwhile, we tailor a memory reduction strategy for GradMA to reduce the scale of QP and memory cost.
    
    \item For completeness, we analyze the convergence of GradMA theoretically in the smooth non-convex setting.
    As a result, the convergence result of GradMA achieves the linear speed up as the number of selected active workers increases.

    \item We conduct extensive experiments on four commonly used image classification datasets~(i.e., MNIST, CIFAR-10, CIFAR-100 and Tiny-Imagenet) to show that GradMA is highly competitive compared with other state-of-the-art baselines.
    Meanwhile, ablation studies demonstrate efficacy and indispensability for core modules and key parameters.
\end{itemize}


\section{Related Work}

\textbf{FL with Data Heterogeneity.}~FedAvg, the classic distributed learning framework for FL, is first proposed by McMahan et al.~\cite{McMahan2017Communication}. 
Although FedAvg provides a practical and simple solution for aggregation, it still suffers performance deterioration when the data among workers is non-iid ~\cite{li2020federated}.
Shortly thereafter, a panoply of modifications for FedAvg have been proposed to handle said issue. 
For example, 
FedProx~\cite{li2020federated1} constrains local updates 
via adding a proximal term to the local objectives. 
Scaffold~\cite{karimireddy2020scaffold} uses 
control variate
to augment the local updates.
FedDyn~\cite{Acar2021Federated} dynamically regularizes the objectives of workers to align global and local objectives.
Moon~\cite{li2021model} corrects the local training by conducting contrastive learning in model-level. 
Meanwhile, there exists another line of works to improve the global performance of FL through performing knowledge distillation~\cite{lin2020ensemble, yao2021local, zhu2021data, zhang2022fine, kim2022multi} on the server side or worker side. 
FedMLB~\cite{kim2022multi} architecturally regularizes the local objectives via online knowledge distillation. 
However, other approaches incur additional communication overhead~\cite{yao2021local, zhu2021data} or pseudo data~\cite{lin2020ensemble, zhang2022fine}. 
Going beyond the aforementioned approaches, FL with momentum is an effective way to tackle worker drift problem caused by data heterogeneity and accelerate the convergence. 
Specifically, on the server side,  FedAvgM~\cite{hsu2019measuring} maintains a momentum buffer, whereas FedADAM~\cite{reddi2020adaptive} and FedAMS~\cite{wang2022communication} both adopt  adaptive gradient-descent methods to speed up training.
FedCM~\cite{xu2021fedcm} keeps a state, carrying global information broadcasted by the server, on the worker side to address data heterogeneity issue.
DOMO~\cite{xu2022coordinating} and Mime~\cite{Karimireddy2020Mime} maintain momentum buffers on both server side and worker side to improve the training performance.

\textbf{FL with Partial Participation.}
In addition to data heterogeneity issue, another key hurdle to FL stems from partial participation.
The causes for partial participation can be roughly classified into two categories. 
One is the difference in the computing power and communication speed of different workers. 
A natural way to cope with this situation is to allow asynchronous updates~\cite{xie2019asynchronous, avdiukhin2021federated, yang2022anarchic}.
The other is the different availability mode, 
in which workers can abort the training midway~(i.e., stragglers)~\cite{kairouz2021advances}.
To do so, 
many approaches may collect information from only a subset of workers to update the centralized model~\cite{McMahan2017Communication, li2020federated, karimireddy2020scaffold, Acar2021Federated, hsu2019measuring, Karimireddy2020Mime, yang2021achieving}. 
However, the server in the mentioned approaches simply ignores and discards the information of the stragglers, which can lead to other problems such as under-utilization of computation and memory~\cite{zhang2022fine}, slower convergence~\cite{li2020federated}, and biased/unfair use of workers' information~\cite{kairouz2021advances}.
Recently, MIFA~\cite{gu2021fast} corrects the gradient bias by exploiting the memorized latest updates of all workers, which avoids excessive delays caused by inactive workers and mitigates CF to some extent.

\textbf{Continual Learning.}
CL is a training paradigm that focuses on scenarios with a continuously changing class distribution of each task and aims at overcoming CF. 
Existing works for CL can be roughly divided into three branches: expansion-based methods~\cite{yoon2019scalable, li2019learn}, regularization-based methods~\cite{kirkpatrick2017overcoming,wang2021training} and memory-based methods~\cite{lopez2017gradient, farajtabar2020orthogonal, saha2021gradient}.
Note that unlike CL, we focus on alleviating CF in distributed data, not sequential data.
There are a handful of recent studies that consider FL with CL.
For example, FedWeIT~\cite{yoon2021federated} focuses on sequential data. FedCurv~\cite{shoham2019overcoming} trains objectives based on all-reduce protocol. FedReg~\cite{xu2022acceleration} and FCCL~\cite{huang2022learn} require generated pseudo data and public data, respectively.



\section{Preliminaries}
This section defines the objective function for FL and introduces QP.

In practice, FL is designed to minimize the empirical risk over data distributed across multiple workers without compromising local data. 
The following optimization problem is often considered:
\begin{equation}
\min_{\bm{x}\in \mathbbm{R}^{d}} f(\bm{x})=\frac{1}{N}\sum_{i=1}^{N} \left[f_{i}(\bm{x})=\frac{1}{n_i}\sum_{r=1}^{n_i}F_i(\bm{x};\bm{\xi}_r^{(i)})\right],
\end{equation}
where $N$ is the number of workers. Moreover, the local objective $f_{i}:\mathbbm{R}^{d} \rightarrow \mathbbm{R}$ measures the local empirical risk over data distribution $\mathcal{D}_i$, i.e., $\bm{\xi}_r^{(i)} \sim \mathcal{D}_i$, with $n_i$ samples available at $i$-th worker. %
Note that
$\mathcal{D}_i$ 
can be different among workers. In this work, we consider the typical centralized setup where $N$ workers are connected to one central server.

Next, we introduce QP, which is a fundamental optimization problem with well-established solutions and can be widely seen in the machine learning community, to correct the server-side and work-side update directions.
In this paper, 
we can model our goal via QP, which is posed in the following primal form:
\begin{equation}
    \label{prim_qp_1:}
    \begin{split}
        \min_{\Tilde{\bm{p}}} \frac{1}{2}\|\bm{p}-\Tilde{\bm{p}}\|^2 \quad {\rm s.t.} \ \langle \Tilde{\bm{p}}, \bm{M}[i]\rangle \geq 0, \forall i \in [C],
    \end{split}
\end{equation}
where $\bm{p} \in \mathbbm{R}^{d}$ and $\bm{M}\in \mathbbm{R}^{d\times C}$
~($C \in \mathbbm{N}$). 
One can see that the goal of~(\ref{prim_qp_1:}) is to seek a vector $\tilde{\bm{p}}$ that is positively correlated with $\bm{M}[i] \in \mathbbm{R}^d , \forall i \in [C]$ while being close to $\bm{p}$. By discarding the constant term $\bm{p}^\top\bm{p}$, we rewrite (\ref{prim_qp_1:}) as:
\begin{equation}
    \label{prim_qp_2:}
    \begin{split}
        &\min_{\Tilde{\bm{p}}} \frac{1}{2}\Tilde{\bm{p}}^\top\Tilde{\bm{p}}-\bm{p}^\top\Tilde{\bm{p}} \quad {\rm s.t.} \ \bm{M}^\top \Tilde{\bm{p}} \succeq \bm{0} \in \mathbbm{R}^C.
    \end{split}
\end{equation}

However, this is a QP problem on $d$ variables, which are updates of the model. 
Generally, $d$ can be enormous, resulting in $d$ being much larger than $C$.
We thus solve the dual formulation of the above QP problem:
\begin{equation}
    \label{prim_qp_3:}
    \begin{split}
        &\min_{\bm{z}} \frac{1}{2}\bm{z}^\top \bm{M}^\top \bm{M} \bm{z} + \bm{p}^\top \bm{Mz} \quad {\rm s.t.} \ \bm{z} \succeq \bm{0} \in \mathbbm{R}^C.
    \end{split}
\end{equation}

Once we solve for the optimal dual variable $\bm{z}^\star$, we can recover the optimal primal solution as $\Tilde{\bm{p}}=\bm{M}\bm{z}^\star+\bm{p}$.

\section{Proposed Approach: GradMA}
We now present the proposed FL approach GradMA, see Alg.~\ref{alg:1} for complete pseudo-code. 
Note that the communication cost of GradMA is the same as that of FedAvg.
Next, we detail core modules of GradMA, which include the memory reduction strategy~(i.e., mem\_red$()$), Worker\_Update$()$ and Server\_Update$()$ on lines 9, 12 and 16 of Alg.~\ref{alg:1}, respectively. 

\begin{algorithm}[tb]
  \caption{GradMA: A Gradient-Memory-based Accelerated Federated Learning}
  \label{alg:1}
\begin{algorithmic}[1]
  \STATE {\bfseries Input:} learning rates ($\eta_l ,\eta_g$), the number of all workers $N$, the number of sampled active workers per communication round $S$, control parameters ($\beta_1$, $\beta_2$),  synchronization interval $I$ and memory size $m$ ($S \leq m \leq \min\{d, N\}$).
  \STATE Initial state $\bm{x}_0^{(i)}=\bm{x}_0\in \mathbbm{R}^d$~($\forall i \in [N]$), $\tilde{\bm{m}}_0=\bm{0}$.
  \STATE Initial $counter = \{c(i) = 0, \forall i \in [N]$\}.
  \STATE Initial memory state $\bm{D}=\{\}$.
  \STATE $buf=\{\}$, $new\_buf=\{\}$.
  \FOR{$t=0,1,\ldots, T-1$}
        \STATE \textbf{On server:}
        \STATE Server samples a subset $\mathcal{S}_t$ with $S$ active workers and transmits $\bm{x}_t$ to $\mathcal{S}_t$.
        \STATE $counter, \bm{D}, buf, new\_buf\leftarrow$ {mem\_red} $(m, \mathcal{S}_t, $ $counter, \bm{D}, buf, new\_buf)$.
        \STATE \textbf{On workers:}
        \FOR{$i \in \mathcal{S}_t$ parallel}
            \STATE $\bm{x}_{t+1}^{(i)}=$ Worker\_Update($\bm{x}_{t}^{(i)}$, $\bm{x}_t$, $\eta_l$, $I$),
            \STATE sends $\bm{d}_{t+1}^{(i)}= \bm{x}_{t}-\bm{x}_{t+1}^{(i)}$ to server.
        \ENDFOR
        \STATE \textbf{On server:}
        \STATE  $\bm{D}, \bm{x}_{t+1}, \tilde{\bm{m}}_{t+1}=$ Server\_Update($ [\bm{d}_{t+1}^{(i)}, i\in \mathcal{S}_t]$, $\Tilde{\bm{m}}_{t}$, $\bm{D}$, $\eta_g$, $\beta_1$, $\beta_2$, $buf$, $new\_buf$).
        \STATE Sends $\bm{x}_{t+1}$ to sampled active workers in the next round.
        \STATE $new\_buf=\{\}$.
  \ENDFOR
  \STATE {\bfseries Output:} $ \bm{x}_T$
\end{algorithmic}
\end{algorithm}

\subsection{Correcting gradient for the worker side}
Throughout the local update, we leverage QP to perform correcting gradient directions, see Alg.~\ref{local_update:}. 
Here, the input of QP~(marked as QP$_l$ for distinction) is $\bm{p}\leftarrow \bm{g}_{\tau}^{(i)}$ and $\bm{M}\leftarrow \bm{G}_{\tau}^{(i)} \in \mathbbm{R}^{d\times 3}$ (line 5 of Alg.~\ref{local_update:}), and its output is the following vector $\tilde{\bm{g}}_{\tau}^{(i)}$, which is positively correlated with $\nabla f_{i}(\bm{x}_{\tau-1}^{(i)}), \nabla f_{i}(\bm{x})$ and $\bm{x}_{ \tau}^{(i)}-\bm{x}_t$ while ensuring the minimum $\|\bm{g}_{\tau}^{(i)}-\tilde{\bm{g}}_{\tau}^{(i)}\|$:
\begin{align}
    \label{tilde_g_sol:}
    &\tilde{\bm{g}}_{\tau}^{(i)}=\bm{G}_{\tau}^{(i)}\bm{z}_{\tau}^\star+\bm{g}_{\tau}^{(i)}\\
    & = z_{\tau,1}^\star\nabla f_{i}(\bm{x}_{\tau-1}^{(i)}) + z_{\tau,2}^\star\nabla f_{i}(\bm{x}) + z_{\tau,3}^\star(\bm{x}_{ \tau}^{(i)}-\bm{x}_t) + \bm{g}_{\tau}^{(i)}, \notag
\end{align}
where $\bm{z}_{\tau}^\star = [z_{\tau,1}^\star, z_{\tau,2}^\star, z_{\tau,3}^\star]^\top$ and $\bm{z}_{\tau}^\star \succeq \bm{0} \in \mathbbm{R}^3$. 
Essentially, the output of QP$_l$ is a conical combination and serves as an update direction for local training. 
Particularly, when $z_{\tau,1}^\star=0, z_{\tau,2}^\star=0$ and $z_{\tau,3}^\star>0$, Eq.~(\ref{tilde_g_sol:}) is equivalent to the local update of FedProx~\cite{li2020federated1}. 
The difference is that the control parameter $\mu$ in FedProx is a hyper-parameter, while  $z_{\tau,3}^\star$ is determined adaptively by QP$_l$.
Specifically, when $\bm{g}_{\tau}^{(i)}$ is positively correlated with $\bm{x}_{ \tau}^{(i)}-\bm{x}_t$, i.e., $\langle\bm{g}_{\tau}^{(i)}, \bm{x}_{ \tau}^{(i)}-\bm{x}_t\rangle\geq 0$, $z_{\tau,3}^\star$ is approximately equal to $0$; otherwise, $z_{\tau,3}^\star$ is greater than $0$.
In other words,
$\bm{x}_{ \tau}^{(i)}-\bm{x}_t$ acts as a hard constraint only when $\bm{g}_{\tau}^{(i)}$ is negatively correlated with $\bm{x}_{ \tau}^{(i)}-\bm{x}_t$, which makes $\tilde{\bm{g}}_{\tau}^{(i)}$ focus more on local information. 
Moreover, the calculation mechanisms for $z_{\tau,1}^\star$ and $z_{\tau,2}^\star$ are the same as that for $z_{\tau,3}^\star$.
When $z_{\tau,1}^\star>0$ and $z_{\tau,2}^\star>0$, it indicates that the update direction $\tilde{\bm{g}}_{\tau}^{(i)}$ takes into account the previous step and global information about the model, which is inspired by CL~\cite{lopez2017gradient, farajtabar2020orthogonal}.
Intuitively, Eq.~(\ref{tilde_g_sol:}) adaptively taps previous and global knowledge, thus effectively mitigating CF caused by data heterogeneity. 

\begin{algorithm}[tb]
  \caption{Worker\_Update($\bm{x}^{\prime}$, $\bm{x}$, $\eta_l$, $I$)}
  \label{local_update:}
\begin{algorithmic}[1]
    \STATE Sets $\bm{x}_{-1}^{(i)}=\bm{x}^{\prime}$, $\bm{x}_{0}^{(i)}=\bm{x}$.
    \FOR{$\tau = 0,1,\ldots, I-1$}
        \STATE $\bm{g}_{\tau}^{(i)}=\nabla f_{i}(\bm{x}_{\tau}^{(i)})$,
        \STATE $\bm{G}_{\tau}^{(i)}=[ \nabla f_{i}(\bm{x}_{\tau-1}^{(i)}), \nabla f_{i}(\bm{x}), \bm{x}_{ \tau}^{(i)}-\bm{x}_t]$,
        \STATE $\tilde{\bm{g}}_{\tau}^{(i)}={\rm QP}_l(\bm{g}_{\tau}^{(i)}, \bm{G}_{\tau}^{(i)})$,
        \STATE $\bm{x}_{\tau+1}^{(i)}=\bm{x}_{\tau}^{(i)}- \eta_{l} \tilde{\bm{g}}_{\tau}^{(i)}$.
    \ENDFOR
    \STATE {\bfseries Output:} $\bm{x}_{I}^{(i)}$.
\end{algorithmic}
\end{algorithm}

\subsection{Correcting update direction for the server side}
Now, we describe the proposed update process of the centralized model on the server side, see Alg.~\ref{server_update:} for details. 
For ease of presentation, we define the number of local updates of workers that the server can store as the memory size $m$.
To elaborate, we assume that there is enough memory space on the server such that $m=N$.
In this way, at communication round $t$, the update process can be streamlined, which takes the form:
\begin{align}
    &\bm{d}_{t+1}=\frac{1}{S}\sum_{i \in \mathcal{S}_t}\bm{d}_{t+1}^{(i)}, \bm{m}_{t+1}=\beta_1\Tilde{\bm{m}}_{t}+\bm{d}_{t+1}, \label{cal_d_:}\\
    &\bm{D}[i]\leftarrow\left\{\begin{array}{l}
            \beta_2 \bm{D}[i]+\bm{d}_{t+1}^{(i)},  i \in \mathcal{S}_t \\
            \beta_2 \bm{D}[i],  i \notin \mathcal{S}_t
            \end{array}  \right. ,\label{update_M_:}\\
    & \tilde{\bm{m}}_{t+1}={\rm QP}_g(\bm{m}_{t+1}, \bm{D}),  \bm{x}_{t+1} = \bm{x}_{t}-\eta_g \tilde{\bm{m}}_{t+1}. \label{tidle_update_m_:}
\end{align}

As shown in Eq.~(\ref{update_M_:}), we propose that the server allocates memory space to maintain a memory state $\bm{D}$, 
which is updated in a momentum-like manner to memorize the accumulated updates of all workers.
Each worker only uploads update $\bm{d}_{t+1}^{(i)}$ ($i \in [N]$) to the server, and as such the risk of data leakage is greatly reduced.
By memorizing accumulated updates of inactive workers, GradMA avoids waiting for any straggler when facing heterogeneous workers with different availability, so as to effectively overcome the adverse effects caused by partial participation.
This is different from the recently proposed MIFA~\cite{gu2021fast}~(see Alg.~\ref{MIAFA:} in Appendix~\ref{appendix_A:}), which stores the latest updates of all workers to perform averaging. 
However, such a straightforward and naive implementation of integration implicitly increases statistical heterogeneity in situations where different workers have varying data distributions, which can induce bias.

Therefore, the core idea of this paper is how to leverage memorized information to overcome the above challenge effectively. 
To tackle the challenge, 
we apply QP~(marked as QP$_g$ for distinction) to seek an update direction $\tilde{\bm{m}}_{t+1}$ that is positively correlated with buffers $\bm{D}[i] \in \mathbbm{R}^d , \forall i \in [N]$ while being close to $\bm{m}_{t+1}$. Concretely, the input of QP$_g$ is $\bm{p} \leftarrow \bm{m}_{t+1}$ and $\bm{M} \leftarrow \bm{D} \in \mathbbm{R}^{d\times N}$, and its output is $\tilde{\bm{m}}_{t+1}$~(see Eq.~(\ref{tidle_update_m_:})), which takes the form $\Tilde{\bm{m}}_{t+1}=\bm{D}\bm{z}_{t+1}^\star+\bm{m}_{t+1}$, where $\bm{z}_{t+1}^\star = [z_{t+1, 1}^\star, \cdots, z_{t+1, N}^\star]^\top \succeq \bm{0} \in \mathbbm{R}^N$ is determined adaptively by QP$_g$.
Inherently, QP$_g$ takes advantage of the accumulated updates of all workers stored on $\bm{D}$ to correct the update direction $\bm{m}_{t+1}$ and circumvents the centralized model from forgetting stragglers' knowledge, thereby alleviating CF induced by partial participation.
In particular, one can easily observe that $\tilde{\bm{m}}_{t+1}=\bm{m}_{t+1}$ holds if $m=0$ (that is, $\bm{D}=\bm{0}$). 
The update process of Alg.~\ref{server_update:} is then consistent with that of FedAvgM~\cite{hsu2019measuring} on the server side. 
Consequently, Alg.~\ref{server_update:} can be considered as an extension of FedAvgM in terms of augmenting updates through allocating memory.

\begin{algorithm}[tb]
  \caption{Server\_Update($[\bm{d}_{t+1}^{(i)}, i\in \mathcal{S}_t]$, $\Tilde{\bm{m}}_{t}$, $\bm{D}$, $\eta_g$, $\beta_1$, $\beta_2$, $buf$, $new\_buf$)}
  \label{server_update:}
\begin{algorithmic}[1]
    \STATE  $\bm{d}_{t+1}=\frac{1}{S}\sum_{i \in \mathcal{S}_t}\bm{d}_{t+1}^{(i)}$, $\bm{m}_{t+1}=\beta_1\Tilde{\bm{m}}_{t}+\bm{d}_{t+1}$.
    \FOR{$c(i) \in buf$}
        \IF{$i \in \mathcal{S}_t$}
            \STATE $\bm{D}[i] \leftarrow\left\{\begin{array}{l}
                 \beta_2 \bm{D}[i]+\bm{d}_{t+1}^{(i)},  c(i) \notin new\_buf \\
                 \bm{d}_{t+1}^{(i)},  c(i) \in new\_buf 
                \end{array}  \right.$.
        \ELSIF{$i \notin \mathcal{S}_t$}
            \STATE $\bm{D}[i] \leftarrow \beta_2 \bm{D}[i]$.
        \ENDIF
    \ENDFOR
    \STATE $\tilde{\bm{m}}_{t+1}={\rm QP}_g(\bm{m}_{t+1}, \bm{D})$,  $\bm{x}_{t+1} = \bm{x}_{t}-\eta_g \tilde{\bm{m}}_{t+1}$.
    \STATE {\bfseries Output:} $\bm{D}$, $\bm{x}_{t+1}$, $\tilde{\bm{m}}_{t+1}$.
\end{algorithmic}
\end{algorithm}

\subsection{A Practical Memory Reduction Strategy}

\begin{algorithm}[tb]
  \caption{mem\_red$(m, \mathcal{S}, c, \bm{D}, buf, new\_buf)$}
  \label{mem_red:}
\begin{algorithmic}[1]
  \FOR{$i \in \mathcal{S}$}
        \IF{$c(i) \in buf$}
            \STATE $c(i) \leftarrow c(i) + 1$.
        \ELSIF{$c(i) \notin buf$}
            \IF{$Length(buf)=m$}
                \STATE $old\_buf=\{\}$.
                \FOR{$k \in buf$}
                    \IF{$k \notin \mathcal{S}$}
                        \STATE $old\_buf \leftarrow old\_buf\cup \{c(k)\}$.
                    \ENDIF
                \ENDFOR
                \STATE Discarding $c(i^\prime)$ with the smallest value from $old\_buf$ and set $c(i^\prime)=0$.
                \STATE Discarding $\bm{D}[i^\prime]$ from memory state $\bm{D}$.
            \ENDIF
            \STATE $c(i) \leftarrow c(i)+1$.
            \STATE $buf \leftarrow buf \cup \{c(i)\}$.
            \STATE $new\_buf\leftarrow new\_buf \cup \{c(i)\}$.
        \ENDIF
    \ENDFOR
  \STATE {\bfseries Output:} $c, \bm{D}, buf, new\_buf$
\end{algorithmic}
\end{algorithm}
It is well known that in realistic FL scenarios, on the one hand, the number of workers may be large; the size of the model may be huge on the other hand, leading to large-scale QP as well as high memory demanding for server to store $\bm{D}$, which is infeasible and unnecessary in practice.

Therefore, we propose a memory reduction strategy to alleviate this deficiency, which ensures that the size of $\bm{D}$ does not exceed a pre-given $m$ and $S \leq m \leq \min \{d, N\}$, see Alg.~\ref{mem_red:} for details.
The design ethos of the memory reduction strategy is to keep as much useful information as possible in a given $m$.
Specifically, at communication round $t$, the server samples $S$ active workers and performs that
$c(i)\leftarrow c(i)+1$~($i \in \mathcal{S}$)~(lines 3 and 15 of Alg.~\ref{mem_red:}).
When the memory used is less than the given one,  $c(i) \notin buf$ of sampled active workers enter the buffers $buf$ and $new\_buf$ in turn~(lines 16-17 of Alg.~\ref{mem_red:}).
Once the memory used is equal to the given one, $c(i^\prime)$ with the smallest value in $old\_buf$ is discarded and set $c(i^\prime)=0$. 
Also, $\bm{D}[i^\prime]$ is discarded from $\bm{D}$~(lines 12-13 of Alg.~\ref{mem_red:}).

\section{Convergence Results for GradMA}
We now present a convergence analysis of  GradMA in the smooth non-convex setting. 
And the following assumptions are considered.
\begin{assumption}
\label{Global_Function_Below_Bounds:}
{\rm (Global function below bounds).} 
Set $f^*=\inf_{\bm{x}\in \mathbbm{R}^d}f(\bm{x})$ and $f^*>-\infty$.
\end{assumption}
\begin{assumption}
\label{L_smooth:}
{\rm ($L$-smooth).}
$\forall i \in [N]$,
the local functions $f_i$ are differentiable, and 
there exist constant $L>0$ such that for any $\bm{x},\bm{y} \in \mathbbm{R}^d$,
$\left \| \nabla f_i(\bm{x})-\nabla f_i(\bm{y}) \right \| \leq L \|\bm{x}-\bm{y}\|$.
\end{assumption}
\begin{assumption}
\label{Bounded_data_heterogeneity:} 
{\rm (Bounded data heterogeneity).}
The degree of heterogeneity of the data distribution across workers can be quantified as $\|\nabla f_i(\bm{x})-\nabla f(\bm{x})\|^2 \leq \rho^2$,
for any $i \in [N]$ and some constant $\rho \geq 0$.
\end{assumption}
\begin{assumption}
\label{UBE_QP_l:} 
{\rm (Bounded optimal solution error for ${\rm QP}_l$).}
Given $\bm{g}^{(i)}=\nabla f_i(\bm{x})$~(see Alg.~\ref{local_update:}),  then there exists $\varepsilon_l > 0$ such that $\|\bm{g}^{(i)}-\Tilde{\bm{g}}^{(i)}\|^2 \leq \varepsilon_{l}^2$.
\end{assumption}
\begin{assumption}
\label{UBE_QP_g:} 
{\rm (Bounded optimal solution error for ${\rm QP}_g$).}
Given $\beta_2\in [0, 1)$ and $\bm{m}$~(see Alg.~\ref{server_update:}), then there exists $\varepsilon_g > 0$ such that $\|\bm{m}-\Tilde{\bm{m}}\|^2 \leq \frac{\varepsilon_{g}^2}{1-\beta_2}$.
\end{assumption}

Assumptions \ref{Global_Function_Below_Bounds:} and \ref{L_smooth:}
are commonly used in the analysis of 
distribution learning \cite{karimireddy2020scaffold, li2020federated1, xin2021hybrid}.
Assumption~\ref{Bounded_data_heterogeneity:} quantifies inter-worker variances, i.e., data heterogeneity \cite{karimireddy2020scaffold, li2020federated1}. 
Assumptions~\ref{UBE_QP_l:} and~\ref{UBE_QP_g:} are necessary for the theoretical analysis of GradMA, which constrain the upper bound on the optimal solution errors of QP$_l$ and QP$_g$, respectively.
Intuitively, the assumptions hold if the local updates for all worker make sense~\cite{esfandiari2021cross}. 
Note that the upper bound for Assumption~\ref{UBE_QP_g:} follows an intuitive observation: more accumulated updates of workers (i.e., the larger $\beta_2$) can provide more accumulated update information for the centralized model.
Next, we state our convergence results for GradMA.
\begin{thm}
    \label{thm:alg}
    Assume Assumptions \ref{Global_Function_Below_Bounds:}-\ref{UBE_QP_g:} exist. Let $\eta_l \leq \frac{1}{160^{0.5}LI}$, $\eta_g\eta_l\leq\frac{(1-\beta_1)^2S(N-1)}{IL(\beta_1S(N-1)+4N(S-1))}$ and $ 320I^2\eta_l^2L^2+\frac{64I\eta_g\eta_lL(1+40I^2\eta_l^2L^2)}{(1-\beta_1)^2}\frac{N-S}{S(N-1)}\leq 1$. For all $t \in [0,\cdots, T-1]$,
    the following relationship generated by Alg.~\ref{alg:1} holds:
    \begin{align}
    &\frac{1}{T}\sum_{t=0}^{T-1}\mathbbm{E}\left[\left\|\nabla f(\bm{x}_{t})\right\|^2\right] \overset{}{\leq} 
         \frac{8(1-\beta_1)(f(\bm{x}_{0}) -f^\star)}{I\eta_g\eta_lT} \notag\\
    & \quad \quad \quad \quad \quad \quad \quad \quad \quad \quad \quad \quad + C_1 \varepsilon_l^2 + C_2 \varepsilon_g^2 + C_3 \rho^2, \notag
    \end{align}
where the expectation $\mathbbm{E}$ is w.r.t the sampled active workers per communication round, and $C_1=8+320I^2\eta_l^2L^2+ \frac{64I\eta_g\eta_lL(1+40I^2\eta_l^2L^2)}{(1-\beta_1)^2}\frac{N-S}{S(N-1)}$, $C_2 =\frac{20  \eta_g L }{(1-\beta_1)^2(1-\beta_2)I\eta_l} + \frac{8}{(1-\beta_2)I^2\eta_l^2}$, $C_3 = C_1 - 8$.
\end{thm}
A detailed proof of Theorem~\ref{thm:alg} is presented in the Appendix~\ref{appendix_C:}. 
\begin{coro}
\label{coro_1:}
Assume Assumptions \ref{Global_Function_Below_Bounds:}-\ref{UBE_QP_g:} exist. 
We set  $\eta_l=\frac{1}{T^{0.5}LI}$, $\eta_g=\frac{S^{0.5}}{I^{0.5}}$, $\varepsilon_l=\frac{1}{T^{0.5}}$ and $\varepsilon_g=\frac{I^{0.25}}{T^{0.75}S^{0.25}L}$.
For $T\geq\max \left\{160, \frac{(\beta_1S(N-1)+4N(S-1))^2}{I^2(1-\beta_1)^4S(N-1)^2}, \frac{(b+(b^2+1280)^{0.5})^2}{4}\right\}$ where $b=\frac{128(N-S)}{(1-\beta_1)^2I^{0.5}S^{0.5}(N-1)}$ in Theorem~\ref{thm:alg}, we have:
\begin{align*}
&\frac{1}{T}\sum_{t=0}^{T-1} \mathbbm{E}\left[\|\nabla f(\bm{x}_t)\|^2\right] =\mathcal{O}\left(\frac{I^{0.5}}{S^{0.5}T^{0.5}}+\frac{1}{T}\right).
\end{align*}
\end{coro}

An immediate observation from Corollary~\ref{coro_1:} is that GradMA can achieve the linear speed up as the number of sampled active workers $S$ increases. 
This convergence rate matches the well-known best result in FL approaches in literature~\cite{yang2021achieving} under the smooth non-convex setting.

\section{Empirical Study}
In this section, we empirically investigate GradMA on four datasets~(MNIST~\cite{lecun1998gradient}, 
CIFAR-10, CIFAR-100~\cite{Krizhevsky2009Learning} and Tiny-Imagenet\footnote{http://cs231n.stanford.edu/tiny-imagenet-200.zip}) commonly used for image classification tasks. 

\subsection{Experimental Setup}
To gauge the effectiveness of Worker\_Update$()$ and Server\_Update$()$, we perform ablation study of \textbf{GradMA}. 
For this purpose, we design Alg.~\ref{GradMA-W:}~(marked as \textbf{GradMA-W}) and Alg.~\ref{GradMA-S:}~(marked as \textbf{GradMA-S}), as specified in Appendix~\ref{appendix_A:}.
Meanwhile, we compare other baselines, including \textbf{FedAvg}~\cite{McMahan2017Communication}, 
\textbf{FedProx}~\cite{li2020federated1}, 
\textbf{MOON}~\cite{li2021model}, 
\textbf{FedMLB}~\cite{kim2022multi},
\textbf{Scaffold}~\cite{karimireddy2020scaffold}, 
\textbf{FedDyn}~\cite{Acar2021Federated}, 
\textbf{MimeLite}~\cite{Karimireddy2020Mime},
\textbf{MIFA}~\cite{gu2021fast}
and slow-momentum variants of FedAvg, FedProx, MIFA, MOON and FedMLB~(i.e., \textbf{FedAvgM}~\cite{hsu2019measuring}, \textbf{FedProxM}, \textbf{MIFAM}, \textbf{MOONM} and \textbf{FedMLBM}),
in terms of test accuracy and communication efficiency in different FL scenarios. 
For fairness, 
we divide the baselines into three groups based on FedAvg's improvements on the \textbf{worker side}, \textbf{server side}, or \textbf{both}. 
See Table~\ref{table_1:} and Table~\ref{table_2:} for details.
Furthermore, on top of GradMA-S, we empirically study the effect of the control parameters~($\beta_1$, $\beta_2$) and verify the effectiveness of men\_red$()$ by setting varying memory sizes $m$.

All our experiments are performed on a centralized network with $100$ workers.
And we fix synchronization interval $I=5$.
To explore the performances of the approaches,
we set up multiple different scenarios w.r.t. the number of sampled active workers $S$ per communication round and data heterogeneity.
Specifically, 
we set $S \in \{5, 10, 50\}$.  
Moreover, we use Dirichlet process $Dp(\omega)$~\cite{Acar2021Federated, zhu2021data} to strictly partition the training set of each dataset across $100$ workers.
We set $\omega \in \{0.01, 0.1, 1.0\}$. 
A visualization of the data partitions for the four datasets at varying $\omega$ values can be found in Fig.~\ref{data_par_sum_appendix:} in Appendix~\ref{appendix_B:}.
Also, the original testing set~(without partitioning) of each dataset is used to evaluate the performance of the trained centralized model.
For MNIST, a neural network (NN) with three linear hidden layers is implemented for each worker. 
We fix the total number of iterations to $2500$, i.e., $T\times I=2500$. 
For CIFAR-10~(CIFAR-100, Tiny-Imagenet), each worker runs a Lenet-5~\cite{lecun1998gradient}~(VGG-11~\cite{simonyan2014very}, Resnet20~\cite{he2016deep}) architecture. 
We fix the total number of iterations to $5000~(10000, 10000)$, i.e., $T\times I=5000~(10000, 10000)$.
Due to the space limitation, we relegate detailed hyper-parameters tuning and full experimental results to Appendix~\ref{appendix_B:}. 

\subsection{Performance Analysis}

\begin{table*}[htbp]
  \centering
  \caption{Top test accuracy~(\%) overview given different FL scenarios.}
  \resizebox{2.0\columnwidth}{!}{
    \begin{tabular}{cccc|cc|ccc|ccc|ccc}
    \toprule
    \multirow{2}[1]{*}{Alg.s} & \multicolumn{3}{c|}{MNIST+NN, $S=10$} & \multicolumn{2}{c|}{MNIST+NN, $\omega=0.01$} & \multicolumn{3}{c|}{CIFAR-10+Lenet-5, $S=10$} & \multicolumn{3}{c|}{CIFAR-100+VGG-11, $\omega=0.1$} & \multicolumn{3}{c}{Tiny-Imagenet+Resnet20, $(\omega, S)$} \\
          & $\omega=1.0$ & $\omega=0.1$ & $\omega=0.01$ & $S=5$ & $S=50$ & $\omega=1.0$ &  $\omega=0.1$ & $\omega=0.01$ & $S=5$ & $S=10$ & $S=50$ & $(0.01,5)$  & $(1.0, 5)$ & $(1.0, 10)$ \\ 
    \midrule
    FedAvg & 98.22$\pm$0.05 & 97.11$\pm$0.39 & 46.19$\pm$1.29 & 49.65$\pm$3.88 & 68.32$\pm$4.16 & 69.48$\pm$8.28 & 47.86$\pm$5.26 & \textbf{20.97$\pm$3.73} & 56.02$\pm$0.37 & 61.22$\pm$0.16 & 64.78$\pm$0.43 & 7.50$\pm$0.32 & 41.80$\pm$0.55 & 42.90$\pm$0.12 \\
    \midrule
    FedProx & 98.16$\pm$0.09 & 97.19$\pm$0.31 & 46.82$\pm$0.96 & 49.89$\pm$3.67 & 67.97$\pm$4.27 & 71.58$\pm$4.66 & 48.63$\pm$4.92 & 20.40$\pm$3.85 & 55.94$\pm$0.71 & 61.25$\pm$0.09 & 64.69$\pm$0.27 & 7.51$\pm$0.46 & 41.82$\pm$0.29 & 42.58$\pm$0.69 \\
    FedMLB & \textbf{98.31$\pm$0.06} & \textbf{97.26$\pm$0.40} & 54.53$\pm$0.39 & 57.22$\pm$3.07 & 68.44$\pm$3.26 & 69.28$\pm$6.56 & 48.99$\pm$4.94 & 20.81$\pm$3.26 & 53.80$\pm$0.16 & 59.20$\pm$0.27 & 64.06$\pm$0.30 & 7.98$\pm$0.34 & 42.83$\pm$0.13 & 43.59$\pm$0.80 \\
    MOON  & 98.18$\pm$0.12 & 97.11$\pm$0.31 & 46.26$\pm$1.35 & 50.39$\pm$5.16 & \textbf{68.75$\pm$4.50} & 71.11$\pm$7.94 & 48.84$\pm$5.16 & 19.39$\pm$3.99 & 55.37$\pm$0.34 & 60.58$\pm$0.60 & 64.48$\pm$0.42 & 7.70$\pm$0.38 & 41.68$\pm$0.22 & 42.80$\pm$0.54 \\
    Scaffold & 97.63$\pm$0.37 & 93.94$\pm$1.18 & 50.86$\pm$7.46 & 39.97$\pm$4.88 & 49.54$\pm$2.28 & 53.33$\pm$6.63 & 35.91$\pm$2.14 & 15.55$\pm$1.33 & 32.22$\pm$0.92 & 34.72$\pm$0.80 & 45.70$\pm$0.76 & 7.20$\pm$0.33 & 40.96$\pm$0.23 & 43.02$\pm$0.30 \\
    GradMA-W & 98.15$\pm$0.10 & 97.01$\pm$0.23 & \textbf{63.34$\pm$3.75} & \textbf{65.39$\pm$0.96} & 65.13$\pm$2.54 & \textbf{72.33$\pm$3.84} & \textbf{50.25$\pm$3.94} & 18.99$\pm$4.06 & \textbf{56.43$\pm$0.51} & \textbf{61.38$\pm$0.11} & \textbf{64.96$\pm$0.36} & \textbf{9.98$\pm$0.22} & \textbf{43.68$\pm$0.23} & \textbf{44.57$\pm$0.45} \\
    \midrule
    FedAvgM & 98.29$\pm$0.18 & 97.20$\pm$0.30 & 53.77$\pm$0.32 & 57.87$\pm$3.64 & 67.80$\pm$5.58 & 71.04$\pm$7.29 & 51.91$\pm$4.46 & 21.02$\pm$3.52 & 55.85$\pm$0.28 & 61.32$\pm$0.29 & 64.88$\pm$0.25 & 16.96$\pm$1.08 & 41.91$\pm$0.23 & 42.57$\pm$0.14 \\
    MIFA  & 98.02$\pm$0.12 & 96.88$\pm$0.56 & 66.92$\pm$2.53 & 56.04$\pm$3.92 & 52.84$\pm$4.89 & 71.41$\pm$5.81 & 50.60$\pm$11.87 & 23.78$\pm$2.04 & 50.37$\pm$1.02 & 58.74$\pm$0.42 & 64.71$\pm$0.31 & 8.88$\pm$0.33 & 41.42$\pm$0.22 & 42.83$\pm$0.13 \\
    MIFAM & 98.02$\pm$0.15 & 96.90$\pm$0.44 & 67.15$\pm$2.23 & 55.28$\pm$6.05 & 53.35$\pm$6.84 & 73.48$\pm$1.37 & 52.13$\pm$9.71 & 24.17$\pm$1.24 & 49.30$\pm$0.86 & 58.91$\pm$0.24 & 64.61$\pm$0.33 & 12.01$\pm$0.32 & 41.94$\pm$0.06 & 43.17$\pm$0.09 \\
    GradMA-S & \textbf{98.38$\pm$0.09} & \textbf{97.35$\pm$0.28} & \textbf{74.52$\pm$1.71} & \textbf{75.93$\pm$0.97} & \textbf{69.09$\pm$3.83} & \textbf{78.76$\pm$1.96} & \textbf{64.60$\pm$5.87} & \textbf{28.41$\pm$2.43} & \textbf{59.08$\pm$0.43} & \textbf{63.23$\pm$0.22} & \textbf{65.63$\pm$0.35} & \textbf{20.93$\pm$1.49} & \textbf{48.83$\pm$1.06} & \textbf{49.65$\pm$0.72} \\
    \midrule
    FedProxM & 98.26$\pm$0.08 & 97.13$\pm$0.34 & 54.50$\pm$0.79 & 58.59$\pm$4.58 & 69.00$\pm$4.42 & 78.00$\pm$1.61 & 51.22$\pm$5.14 & 21.80$\pm$3.72 & 55.63$\pm$0.31 & 63.15$\pm$0.12 & 64.78$\pm$0.11 & 18.30$\pm$0.79 & 37.98$\pm$0.10 & 45.27$\pm$0.19 \\
    FedMLBM & 98.26$\pm$0.16 & \textbf{97.35$\pm$0.30} & 61.12$\pm$1.48 & 64.12$\pm$4.17 & 68.78$\pm$3.28 & 73.70$\pm$4.62 & 49.90$\pm$5.82 & 21.53$\pm$2.93 & 53.91$\pm$0.78 & 60.44$\pm$0.34 & 64.85$\pm$0.18 & 17.32$\pm$0.82 & 44.62$\pm$0.32 & 45.18$\pm$0.27 \\
    MOONM & 98.21$\pm$0.13 & 97.04$\pm$0.42 & 62.34$\pm$8.91 & 57.98$\pm$5.51 & 68.82$\pm$4.43 & 73.96$\pm$4.11 & 50.06$\pm$6.14 & 20.19$\pm$3.10 & 56.01$\pm$0.25 & 62.06$\pm$0.19 & 65.37$\pm$0.17 & 16.78$\pm$0.95 & 42.43$\pm$0.39 & 42.78$\pm$0.46 \\
    Feddyn & 97.92$\pm$0.12 & 96.03$\pm$0.46 & 59.39$\pm$2.29 & 65.36$\pm$5.20 & 57.68$\pm$4.30 & 74.94$\pm$2.48 & 41.93$\pm$3.22 & 17.94$\pm$3.52 & 52.95$\pm$1.63 & 58.48$\pm$0.18 & 61.71$\pm$0.25 & 17.89$\pm$0.95 & 44.37$\pm$0.57 & 44.86$\pm$0.15 \\
    MimeLite & 98.19$\pm$0.07 & 97.10$\pm$0.31 & 54.86$\pm$13.36 & 51.04$\pm$4.15 & \textbf{69.41$\pm$4.15} & 77.98$\pm$1.48 & 53.27$\pm$1.69 & 20.73$\pm$3.33 & 58.00$\pm$0.51 & 63.29$\pm$0.49 & 64.68$\pm$0.33 & 8.29$\pm$0.29 & 41.05$\pm$0.21 & 41.56$\pm$0.18 \\
    GradMA & \textbf{98.39$\pm$0.04} & 97.34$\pm$0.35 & \textbf{77.97$\pm$1.28} & \textbf{75.51$\pm$1.94} & 66.68$\pm$3.03 & \textbf{79.92$\pm$0.59} & \textbf{65.91$\pm$5.10} & \textbf{30.81$\pm$1.78} & \textbf{59.47$\pm$0.58} & \textbf{63.49$\pm$0.47} & \textbf{65.68$\pm$0.25} & \textbf{23.52$\pm$1.32} & \textbf{49.29$\pm$0.86} & \textbf{50.54$\pm$0.56} \\
    \midrule
    \end{tabular}}
  \label{table_1:}
\end{table*}

\begin{table*}[htbp]
  \centering
  \caption{Communication rounds to reach given test accuracy $ac$ under different FL scenarios. Note that since Scaffold and MimeLite have twice as much communication load per communication round as the other approaches, we use $2\times$ to show the distinction.}
  \resizebox{2.0\columnwidth}{!}{
    \begin{tabular}{cccc|cc|ccc|ccc|ccc}
    \toprule
    \multirow{3}[4]{*}{Alg.s} & \multicolumn{3}{c|}{MNIST, $S=10$} & \multicolumn{2}{c|}{MNIST,  $\omega=0.01$} & \multicolumn{3}{c|}{CIFAR-10, $S=10$} & \multicolumn{3}{c|}{CIFAR-100, $\omega=1.0$} & \multicolumn{3}{c}{Tiny\_Imagenet+Resnet20, $(\omega, S)$} \\
\cmidrule{2-15}          & $w=1.0$ & $w=0.1$ & $w=0.01$ & $S=5$   & $S=50$  & $w=1.0$ & $w=0.1$ & $w=0.01$ & $S=5$   & $S=10$  & $S=50$  & $(0.01,5)$ & $(1.0, 5)$ & $(1.0, 10)$ \\
          & $ac=95\%$ & $ac=95\%$ & $ac=45\%$ & $ac=40\%$ & $ac=50\%$ & $ac=55\%$ & $ac=45\%$ & $ac=15\%$ & $ac=30\%$ & $ac=40\%$ & $ac=60\%$ & $ac=5\%$ & $ac=30\%$ & $ac=35\%$ \\
    \midrule
    FedAvg & 25    & 115   & 493   & 299   & 117   & 177   & 882   & 106   & 437   & 435   & 1,071  & 986   & 906   & 1,116  \\
    \midrule
    FedProx & 25    & 120   & 478   & 283   & 117   & 141   & 766   & 197   & 429   & 435   & 1,081  & 986   & 906   & 1,091  \\
    FedMLB & \textbf{22}    & 115   & 280   & 203   & \textbf{78}    & 245   & 694   & \textbf{63}    & 511   & 548   & 1,404  & \textbf{791}   & 806   & 966  \\
    MOON  & 25    & 116   & 493   & 283   & 117   & \textbf{96}    & 616   & 247   & 475   & 458   & 1,088  & 986   & 906   & 1,116  \\
    Scaffold~(2$\times$) & 26    & --    & \textbf{60}    & --    & --    & --    & --    & --    & 806   & --    & --    & 971   & 971   & 1,181  \\
    GradMA-W & 30    & \textbf{111}   & 125   & \textbf{78}    & 172   & 168   & \textbf{582}   & 136   & \textbf{389}   & \textbf{420}   & \textbf{1,027}  & 821   & \textbf{841}   & \textbf{936} \\
    \midrule
    FedAvgM & 19    & 116   & 280   & 138   & 79    & 175   & 605   & 73    & 429   & 397   & 654   & 286   & 821   & 991  \\
    MIFA  & 42    & 85    & 67    & 62    & 89    & 156   & 393   & 54    & 631   & 567   & 1,075  & 716   & 1,021  & 1,151  \\
    MIFAM & 39    & 80    & 60    & 62    & 80    & 152   & 325   & 25    & 677   & 547   & 883   & 656   & 956   & 1,076  \\
    GradMA-S & \textbf{16}    & \textbf{43}    & \textbf{38}    & \textbf{38}    & \textbf{56}    & \textbf{83}    & \textbf{101}   & \textbf{14}    & \textbf{297}   & \textbf{266}   & \textbf{579}   & \textbf{131}   & \textbf{231}   & \textbf{251} \\
    \midrule
    FedProxM & 22    & 115   & 280   & 138   & 96    & 97    & 631   & 67    & 468   & 397   & 665   & 291   & 821   & 611  \\
    FedMLBM & \textbf{16}    & 87    & 191   & 113   & \textbf{77}    & 185   & 653   & 69    & 522   & 522   & 922   & 296   & 681   & 796  \\
    MOONM & 19    & 115   & 186   & 210   & 95    & 78    & 766   & 85    & 461   & 417   & 697   & 291   & 821   & 991  \\
    Feddyn & 19    & 73    & 136   & 128   & 96    & 66    & --    & 83    & 441   & 356   & 736   & 371   & 396   & 406  \\
    MimeLite~(2$\times$) & 32    & 115   & 277   & 284   & 113   & 92    & 385   & 134   & 355   & 329   & 675   & 1,016  & 871   & 1,071  \\
    GradMA & 18    & \textbf{49}    & \textbf{37}    & \textbf{29}    & 105   & \textbf{49}    & \textbf{130}   & \textbf{11}    & \textbf{274}   & \textbf{253}   & \textbf{559}   & \textbf{146}   & \textbf{231}   & \textbf{231} \\
    \midrule
    \end{tabular}}
  \label{table_2:}
\end{table*}

\begin{figure}[t]
\centering
\includegraphics[width=0.45\textwidth]{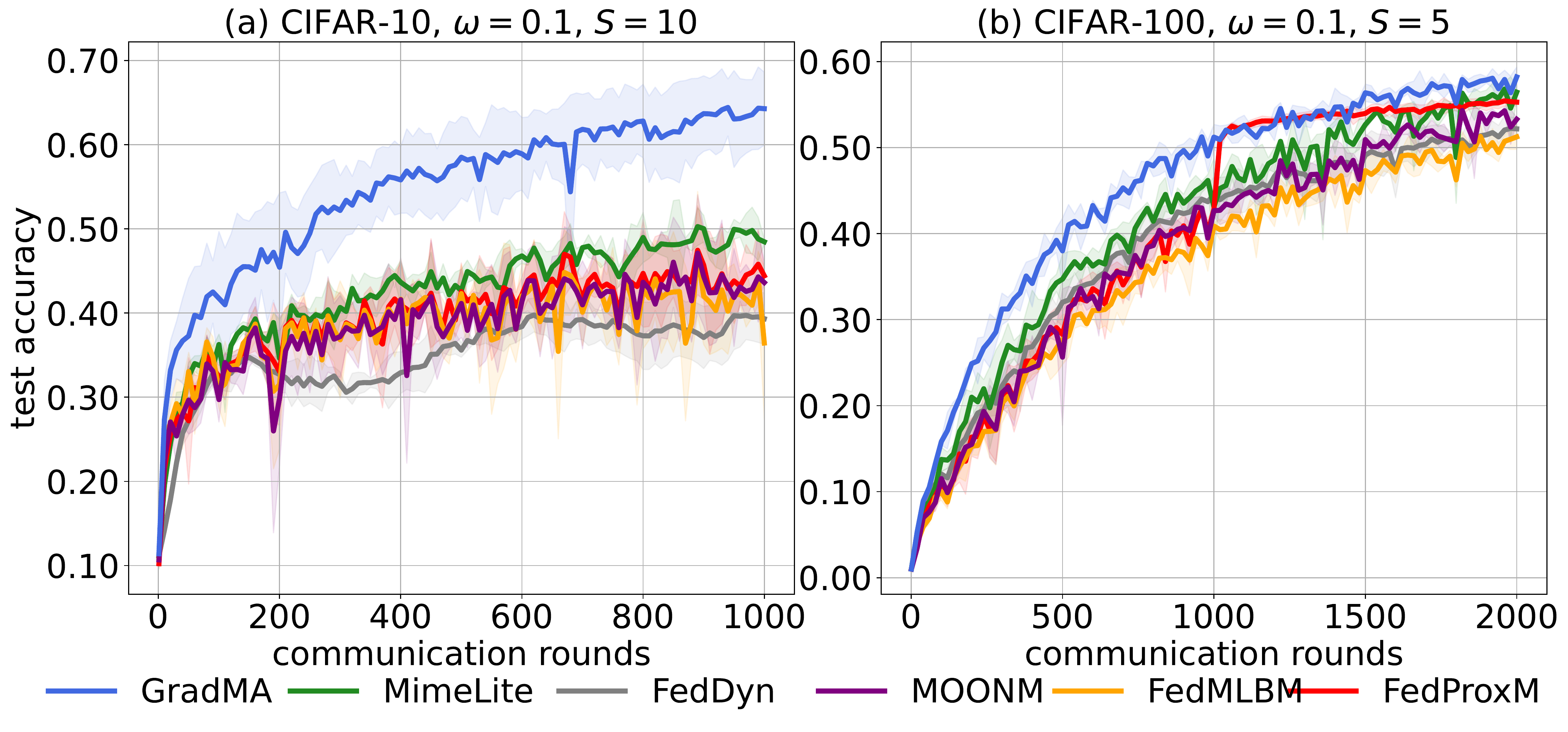} 
\caption{Test accuracy curves selected of GradMA as well as baselines over CIFAR-10 and CIFAR-100.}
\label{fig_gradma_baseline_main:}
\end{figure}

\textbf{Effects of data heterogeneity.} 
From Table~\ref{table_1:}, one can see that the performances of all approaches degrade severely with decreasing $\omega$ on MNIST,  CIFAR-10 and Tiny-Imagenet, with GradMA being the only approach that is robust while surpasses other baselines with an overwhelming margin against most scenarios.
In particular, the higher data heterogeneity, the more superior performance for GradMA.
Also, as shown in Table~\ref{table_2:} and Fig.~\ref{fig_gradma_baseline_main:}, GradMA requires much less communication rounds to reach a given test accuracy compared to baselines against most scenarios.
These results validate our idea in the sense that the advantage of GradMA comes from the effective adaptive utilization of workers' information on both the worker side and server side, which alleviates negative impacts caused by the discrepancy of data distributions among workers.

\begin{figure}[b]
\centering
\includegraphics[width=0.45\textwidth]{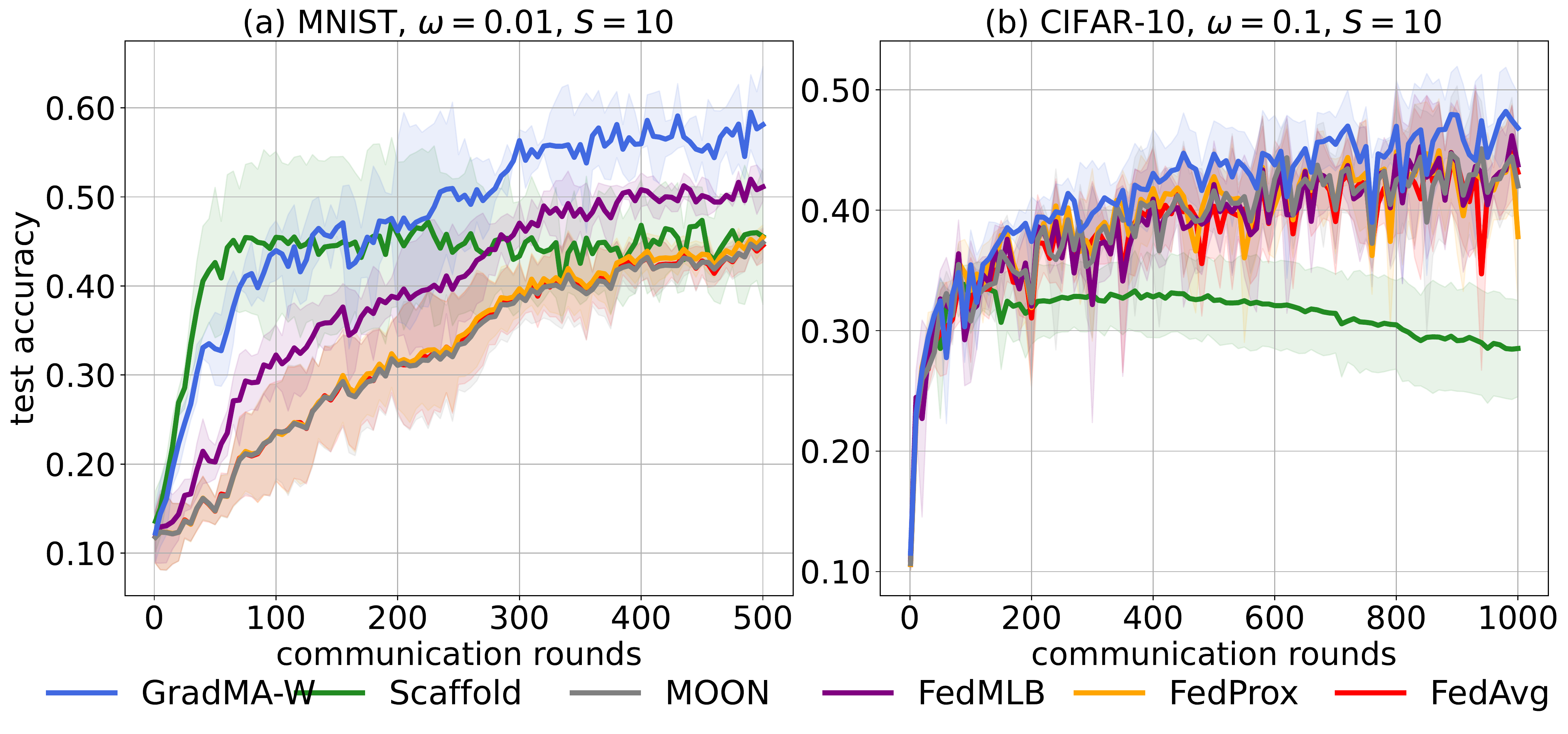} 
\caption{Test accuracy curves selected of GradMA-W as well as baselines over MNIST and CIFAR-10.}
\label{fig_gradma_w_baseline_main:}
\end{figure}

\textbf{Impacts of stragglers.} 
We explore the impacts of different $S$ on MNIST, CIFAR-100 and Tiny-Imagenet.  
A higher $S$ means more active workers upload updates per communication round.
From Table~\ref{table_1:}, we can clearly see that the performance of all approaches improves uniformly with increasing $S$ on CIFAR-100 and Tiny-Imagenet, where GradMA consistently dominates other baselines in terms of test accuracy.
Meanwhile, Fig.~\ref{fig_gradma_baseline_main:} shows that the learning efficiency of GradMA consistently outperforms other baselines~(see Appendix~\ref{appendix_B:} for more results). 
However, for MNIST, the test accuracy for most of the approaches does not intuitively improve with increasing $S$.
We conjecture that for simple classification tasks and models, the more active workers participating in training, the more prone the centralized model is to overfitting.
\begin{figure}[t]
\centering
\includegraphics[width=0.45\textwidth]{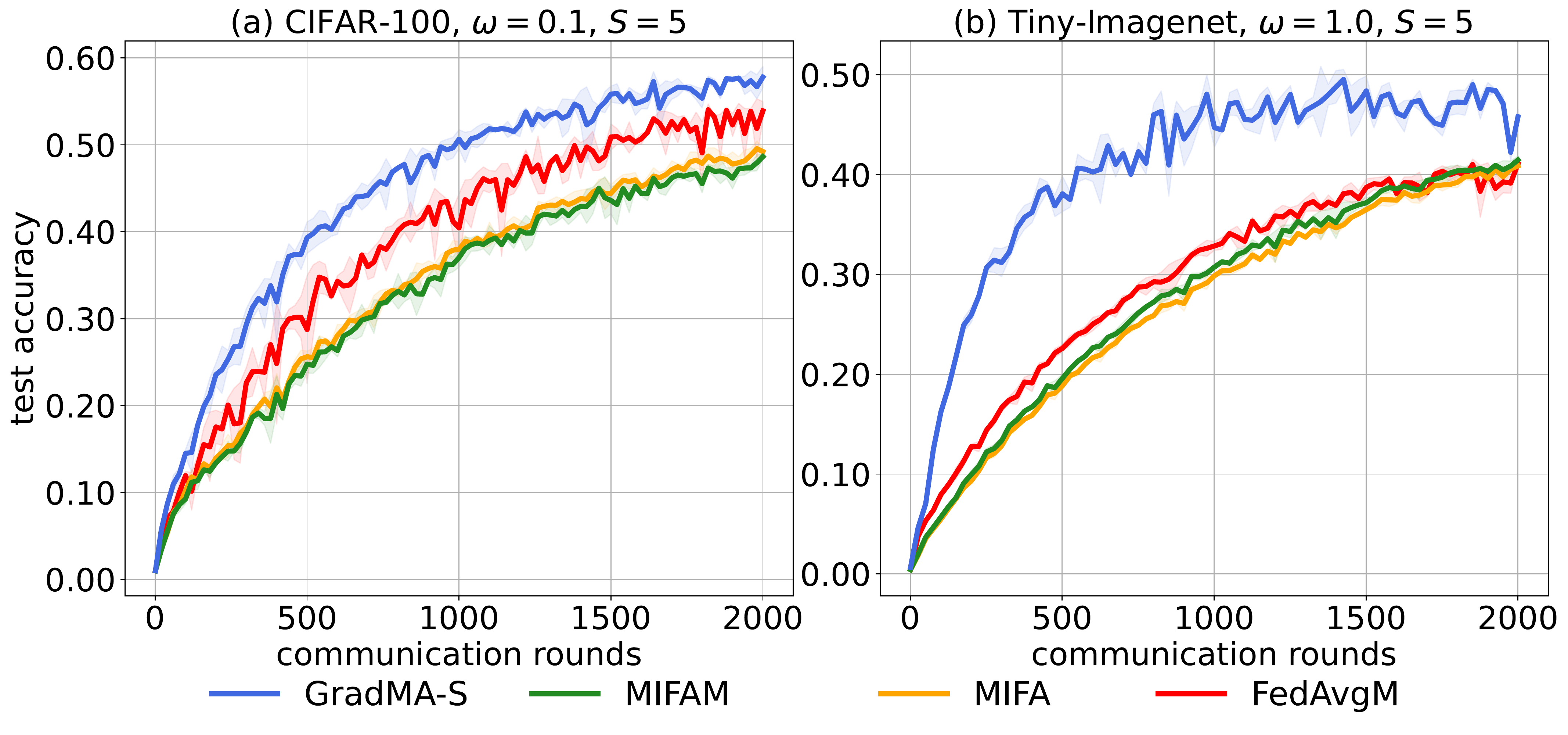} 
\caption{Test accuracy curves selected of GradMA-S as well as baselines over CIFAR-100 and Tiny-Imagenet.}
\label{fig_gradma_s_baseline_main:}
\end{figure}

\begin{figure}[t]
\centering
\includegraphics[width=0.45\textwidth]{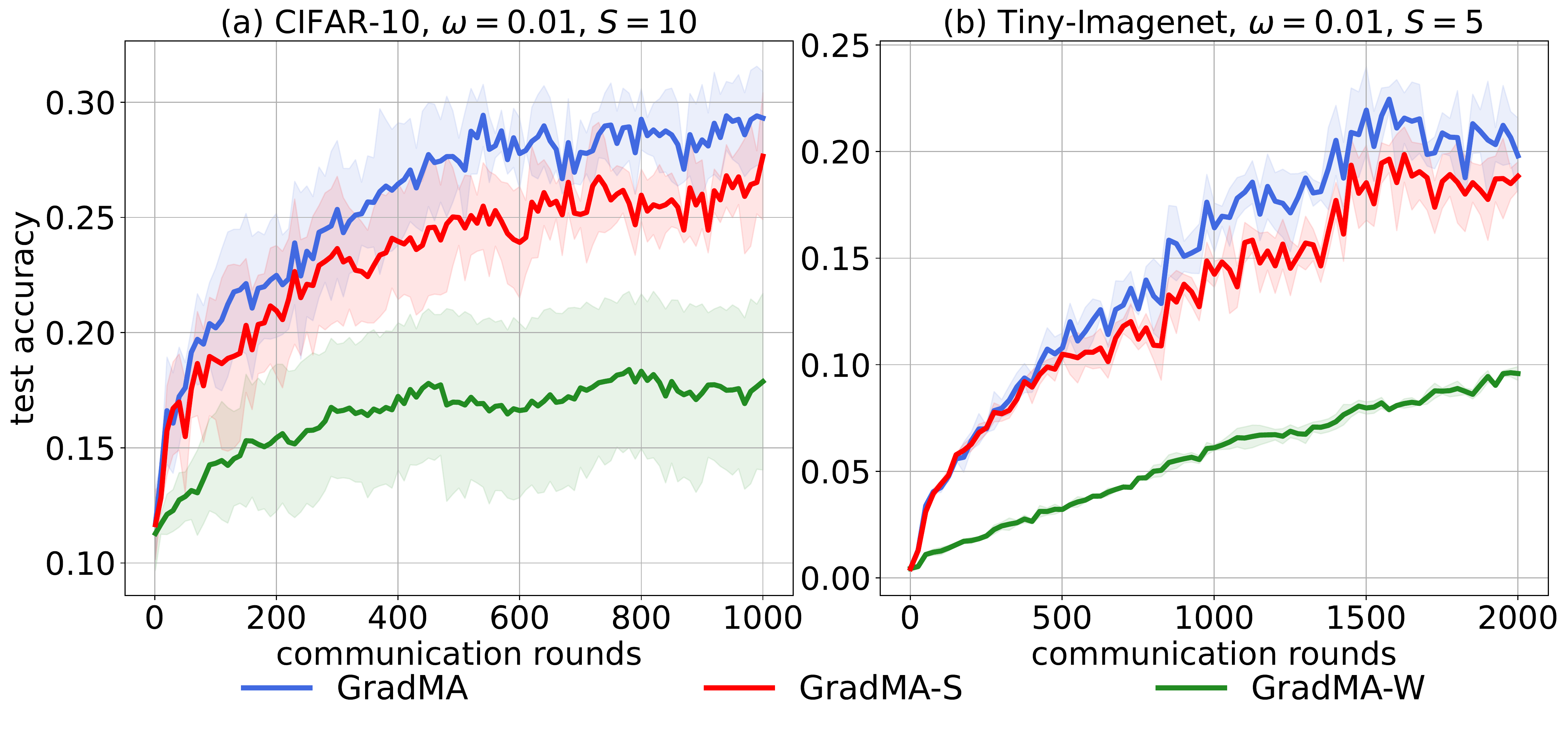} 
\caption{Test accuracy curves selected of GradMA, GradMA-S and GradMA-W over CIFAR-10 and Tiny-Imagenet.}
\label{fig_gradma_s_w_main:}
\end{figure}

\textbf{Comments on GradMA-W and GradMA-S.} 
We now discuss the empirical performances of GradMA-W and GradMA-S and observe that GradMA-S beats GradMA-W by a significant margin in different FL scenarios, and even slightly outperforms GradMA in a few cases~(see Table~\ref{table_1:} and Table~\ref{table_2:}).
To put it differently, GradMA leads GradMA-S in most FL scenarios, suggesting that the combination of Worker\_Update$()$ and Server\_Update$()$ can have a positive effect and thus improve performance.
Meanwhile, GradMA-W trumps baselines in most cases, which suggests that Worker\_Update$()$ can mitigate the issue of CF and thus augment the centralized model.
In addition, we can draw an empirical conclusion that correcting the update direction of the centralized model on the server can greatly boost accuracy compared to correcting that of the local model for each worker.
Selected learning curves shown in Fig.~\ref{fig_gradma_w_baseline_main:} and Fig.~\ref{fig_gradma_s_w_main:} verify the above statements.

Next, we further explore effects of ($\beta_1$, $\beta_2$) and $m$ on the performance of GradMA-S on MNIST and CIFAR-10. 

\begin{figure}[t]
\centering
\includegraphics[width=0.45\textwidth]{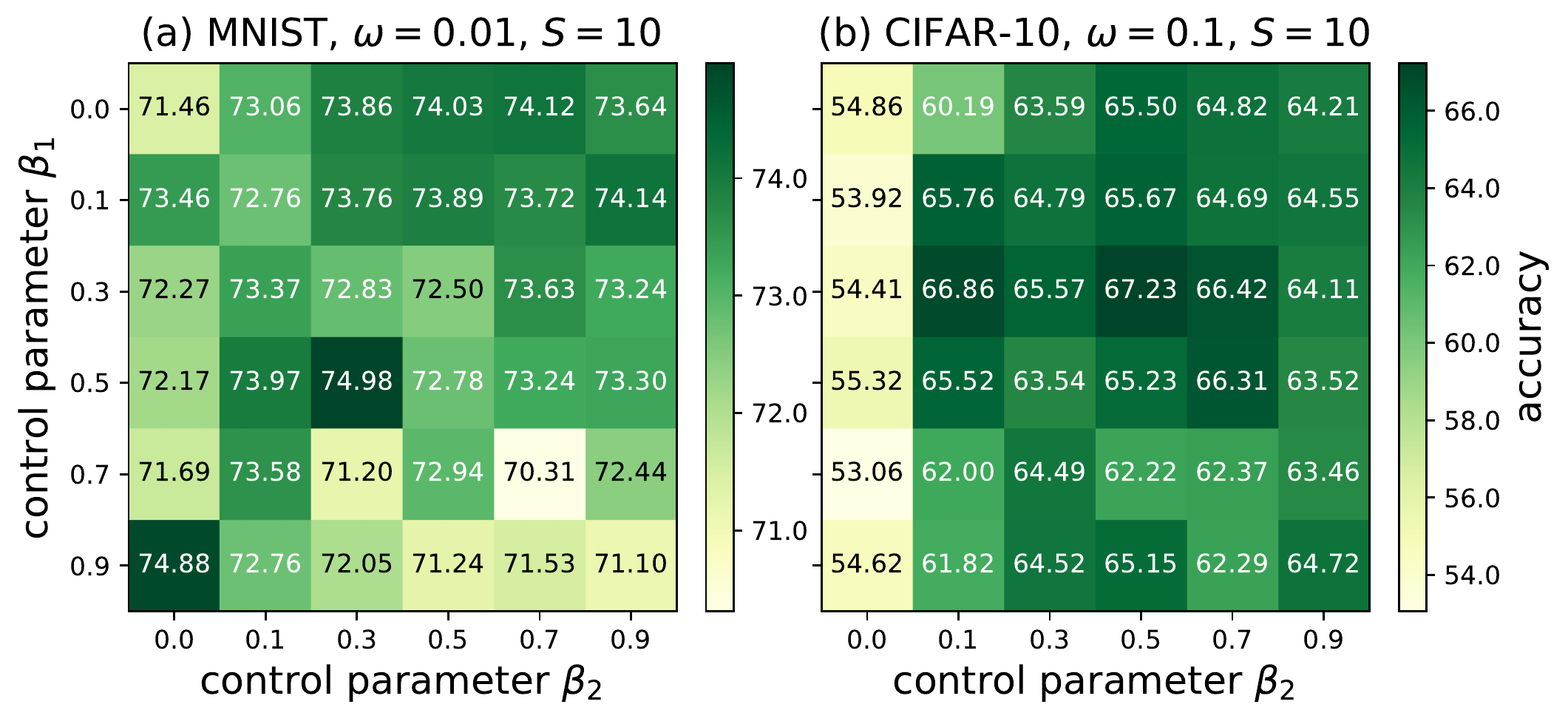} 
\caption{Top test accuracy~(\%) overview for GradMA-S with varying control parameters ($\beta_1$, $\beta_2$) on MNIST and CIFAR-10.}
\label{beta_1_beta2_main:}
\end{figure}

\textbf{Varying control parameters~($\beta_1$, $\beta_2$).}  In order to explore effects of ($\beta_1$, $\beta_2$) in more detail, we set $\beta_1, \beta_2 \in \{0.0, 0.1, 0.3, 0.5, 0.7, 0.9\}$. And we fix $m=100$ and $S=10$. 
Notice that some similarities exist between GradMA-S and MIFA~(MIFAM) when $\beta_1 = 0.0$ and $\beta_2 = 0.0$~($\beta_1 = 0.0$ and $ \beta_2 > 0.0$), i.e., they both memorize the latest updates of stragglers at the server side. 
From Table~\ref{table_1:} and Fig.~\ref{beta_1_beta2_main:}~(refer to Appendix~\ref{appendix_B:} for more results), 
we can see that GradMA-S with $\beta_2 = 0.0$ considerably beats MIFA and MIFAM regardless of value of $\beta_1$.
Furthermore, we observe that GradMA-S with $\beta_2 > 0.0$ outperforms GradMA-S with $\beta_2 = 0.0$ in most cases, and the best test accuracy is located in the region of $\beta_2 > 0.0$.
This indicates that the accumulated updates of stragglers can provide more effective update information for the centralized model to refine the performance of GradMA-S compared to the latest updates of stragglers.

\textbf{Varying memory sizes $m$.} 
\begin{figure}[t]
\centering
\includegraphics[width=0.45\textwidth]{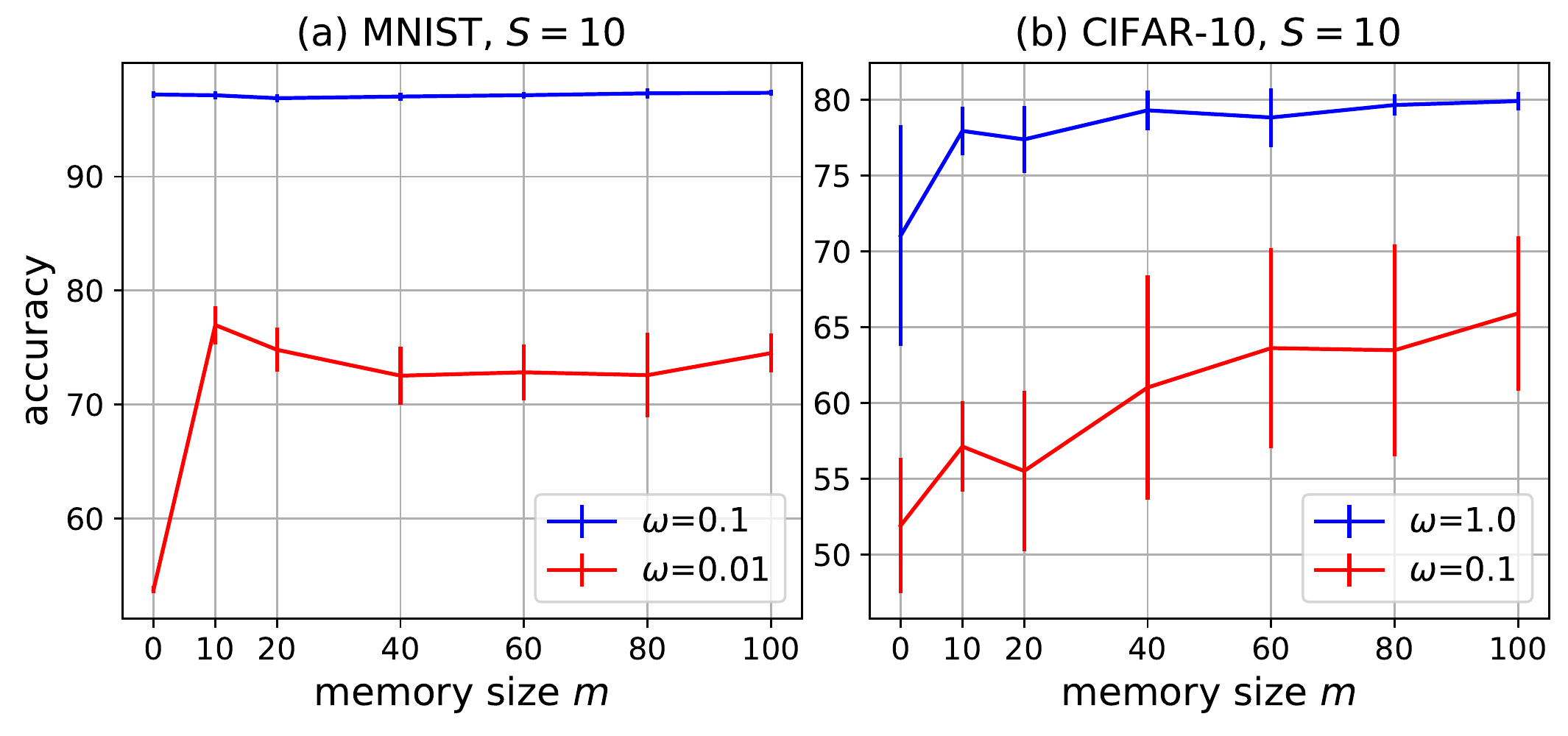} 
\caption{Top test accuracy~(\%) overview for GradMA-S with varying memory sizes $m$ on MNIST and CIFAR-10.}
\label{m_omega_main:}
\end{figure}
\begin{figure}[t]
\centering
\includegraphics[width=0.45\textwidth]{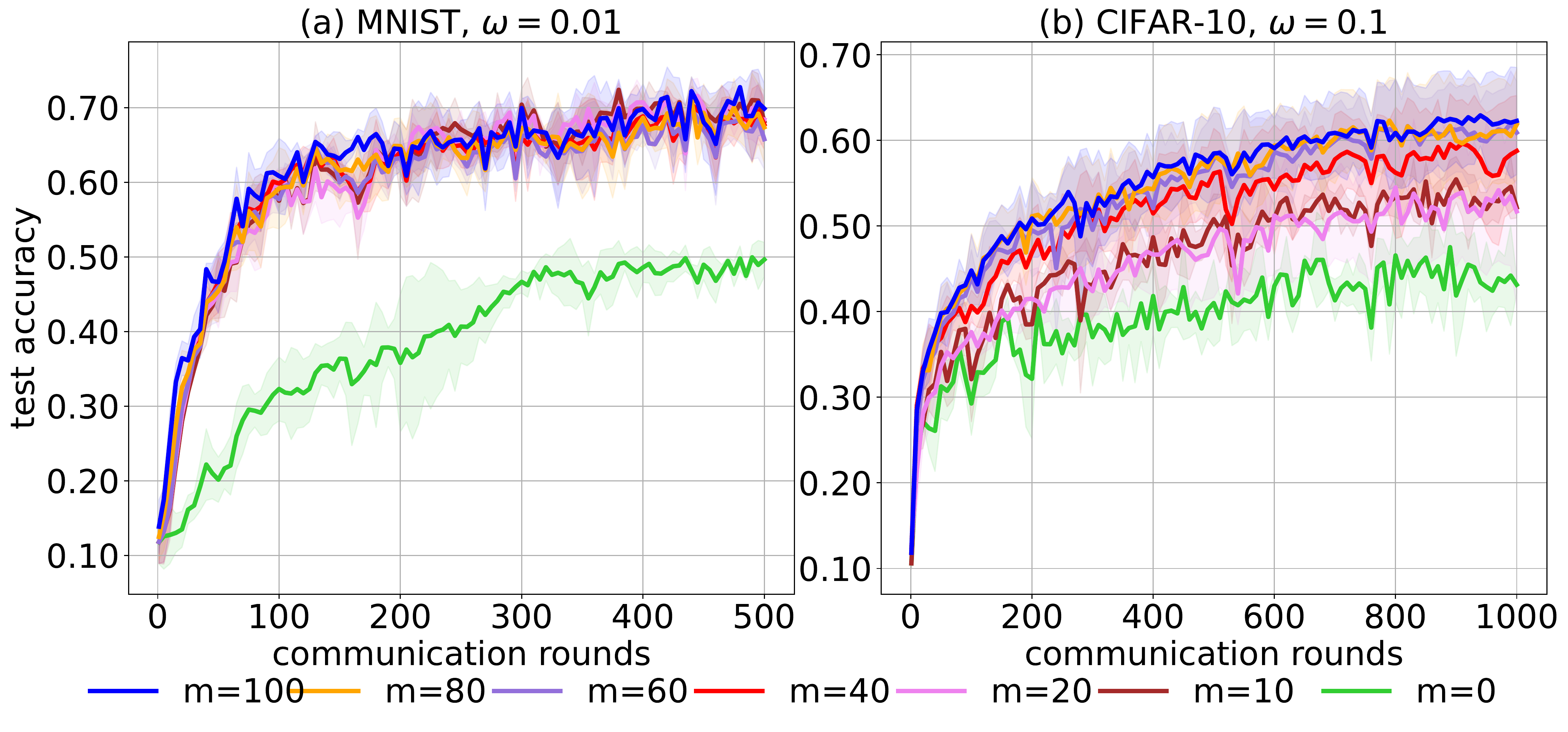} 
\caption{Test accuracy curves selected of GradMA-S with varying memory sizes $m$ on MNIST and CIFAR-10.}
\label{m_omega_comm_round_main:}
\end{figure}
In a real-world FL scenario, the memory space on the server side determines the value of the tunable parameter $m$ for GradMA-S. 
Here, we fix $S=10$ and set $m \in \{0, 10, 20, 40, 60, 80, 100\}$ to carefully look into the performance of GradMA-S with varying $m$.
Notably, GradMA-S with $m=0$ is equivalent to FedAvgM. 
From Fig.~\ref{m_omega_main:},  
FedAvgM performs comparably to GradMA-S with $m>0$ for MNIST under moderate data heterogeneity setting~(i.e., $\omega=0.1$).
In contrast, the performance of FedAvgM sharply degrades and is seriously worse than that of GradMA-S with $m>0$ under high data heterogeneity setting~(i.e., $\omega=0.01$).
Meanwhile, for CIFAR-10, GradMA-S with $m>0$ consistently surpasses FedAvgM, even under mild data heterogeneity setting~(i.e., $\omega=1.0$).
Besides, we can see that the performance of GradMA-S does not intuitively and monotonically improve with increasing $m$. 
This indicates that the quality of the memory reduction strategy is an essential ingredient affecting the performance of GradMA-S for a given $m$. 
Therefore, how to tailor a more effective memory reduction strategy is one of our future works.
From Fig.~\ref{m_omega_comm_round_main:}, the learning curves selected also echo the said statements~(see Appendix~\ref{appendix_B:} for more results).

\section{Conclusions}
In this paper, we propose a novel FL approach GradMA, which corrects the update directions of the server and workers simultaneously.
Specifically, 
on the worker side, GradMA utilizes the gradients of the local model in the previous step and the centralized model, and the parameters difference between the local model in the current round and the centralized model as constraints of QP to adaptively correct the update direction of the local model.
On the server side, GradMA takes the memorized accumulated gradients of all workers as constraints of QP to augment the update direction of the centralized model.
Meanwhile, we provide the convergence analysis theoretically of GradMA in the smooth non-convex setting.
Also, we conduct extensive experiments to verify the superiority of GradMA.

\section{Acknowledgments}
\label{sec:acknowledge}
This work has been supported by the National Natural Science Foundation of China under Grant No.U1911203, 
and the National Natural Science Foundation of China under Grant No.61977025.


{\small
\bibliographystyle{ieee_fullname}
\bibliography{egbib}
}

\onecolumn
\clearpage
\appendix
\section*{Appendix}

\subsection{\quad Pseudocodes}
\label{appendix_A:}

\begin{algorithm}[H]
  \caption{FedAvgM~(FedProxM)}
  \label{fedavgprox_slowm:}
\begin{algorithmic}[1]
  \STATE {\bfseries Input:}  learning rates ($\eta_l$,  $\eta_g$), control parameters $\mu$ and $\beta_1$, synchronization interval $I$ and the number of workers $N$.
  \STATE Initial state $\bm{x}_0^{(i)}=\bm{x}_0 \in \mathbb{R}^d$,  $\forall i \in [N]$ and $\bm{m}_0=\bm{0}$.
  \FOR{$t=0,1,\ldots, T-1$}
        \STATE \textbf{On server:}
        \STATE Server samples a subset $\mathcal{S}_t$ with $S$ active workers from $[N]$ and transmits $\bm{x}_t$ to $\mathcal{S}_t$.
        \STATE \textbf{On workers:}
        \FOR{$i \in \mathcal{S}_t$ parallel}
            \STATE Sets $\bm{x}_{t,0}^{(i)}=\bm{x}_t$.
            \FOR{$\tau = 0,1,\ldots, I-1$}
                \STATE $\bm{x}_{t,\tau+1}^{(i)}=\bm{x}_{t,\tau}^{(i)}- \eta_l \nabla f_{i}(\bm{x}_{t, \tau}^{(i)})$. (FedAvgM)
                \STATE $\bm{x}_{t,\tau+1}^{(i)}=\bm{x}_{t,\tau}^{(i)}- \eta_l( \nabla f_{i}(\bm{x}_{t, \tau}^{(i)})+\mu(\bm{x}_{t, \tau}^{(i)}-\bm{x}_{t}))$. (FedProxM)
            \ENDFOR
            \STATE Sends $\bm{d}_{t+1}^{(i)}=\bm{x}_t-\bm{x}_{t,I}^{(i)}$ to server.
        \ENDFOR
        \STATE \textbf{On server:}
        \STATE $\bm{d}_{t+1}=\frac{1}{S}\sum_{i \in \mathcal{S}_t}\bm{d}_{t+1}^{(i)}$, $\bm{m}_{t+1}=\beta_1 \bm{m}_{t} + \bm{d}_{t+1}$.
        \STATE  $\bm{x}_{t+1}=\bm{x}_{t}- \eta_g \bm{m}_{t+1}$. 
        \STATE Sends $\bm{x}_{t+1}$ to sampled active workers in the next round.
  \ENDFOR
  \STATE {\bfseries Output:} $\bm{x}_T$
\end{algorithmic}
\end{algorithm}

\begin{algorithm}[H]
  \caption{MIFAM~(MIFA, i.e., MIFAM with $\beta_1=0.0$)}
  \label{MIAFA:}
\begin{algorithmic}[1]
  \STATE {\bfseries Input:}  learning rates ($\eta_l$,  $\eta_g$), control parameter $\beta_1$, synchronization interval $I$ and the number of workers $N$.
  \STATE Initial state $\bm{x}_0^{(i)}=\bm{x}_0\in \mathbb{R}^d$, $\bm{g}_{old}^{(i)}=\bm{0}$, $\forall i \in [N]$, $\bm{d}_{0} = \frac{1}{N}\sum_{i=1}^{N} \bm{g}_{old}^{(i)}$ and $\bm{m}_0=\bm{0}$.
  \FOR{$t=0,1,\ldots, T-1$}
        \STATE \textbf{On server:}
        \STATE Server samples a subset $\mathcal{S}_t$ with $S$ active workers from $[N]$ and transmits $\bm{x}_t$ to $\mathcal{S}_t$.
        \STATE \textbf{On workers:}
        \FOR{$i \in S_t$ parallel}
            \STATE Sets $\bm{x}_{t,0}^{(i)}=\bm{x}_t$.
            \FOR{$\tau = 0,1,\ldots, I-1$}
                \STATE $\bm{x}_{t,\tau+1}^{(i)}=\bm{x}_{t,\tau}^{(i)}- \eta_l \nabla f_{i}(\bm{x}_{t, \tau}^{(i)})$.
            \ENDFOR
            \STATE Computes $\bm{g}_{t+1}^{(i)} = \bm{x}_{t}-\bm{x}_{t,I}^{(i)}$.
            \STATE Sends $\bm{d}_{t+1}^{(i)} = \bm{g}_{t+1}^{(i)} - \bm{g}_{old}^{(i)}$ to Server.
            \STATE Sets $\bm{g}_{old}^{(i)} = \bm{g}_{t+1}^{(i)}$.
        \ENDFOR
        \STATE \textbf{On server:}
        \STATE $\bm{d}_{t+1}=\bm{d}_{t}+\frac{1}{N}\sum_{i\in \mathcal{S}_t}\bm{d}_{t+1}^{(i)}$,  $\bm{m}_{t+1}=\beta_1\bm{m}_{t}+\bm{d}_{t+1}$.
        \STATE  $\bm{x}_{t+1}=\bm{x}_{t}- \eta_g\bm{m}_{t+1}$. 
        \STATE Sends $\bm{x}_{t+1}$ to sampled active workers in the next round.
  \ENDFOR
  \STATE {\bfseries Output:} $\bm{x}_T$
\end{algorithmic}
\end{algorithm}

\begin{algorithm}[t]
  \caption{GradMA-W}
  \label{GradMA-W:}
\begin{algorithmic}[1]
  \STATE {\bfseries Input:} learning rates ($\eta_l ,\eta_g$), the number of all workers $N$, the number of active workers each round $S$ and  synchronization interval $I$.
  \STATE Initial state $\bm{x}_0^{(i)}=\bm{x}_0\in \mathbbm{R}^d$, $\forall i \in [N]$.
  \FOR{$t=0,1,\ldots, T-1$}
        \STATE \textbf{On server:}
        \STATE Server samples a subset $\mathcal{S}_t$ with $S$ active workers and transmits $\bm{x}_t$ to $\mathcal{S}_t$.
        \STATE \textbf{On workers:}
        \FOR{$i \in \mathcal{S}_t$ parallel}
            \STATE $\bm{x}_{t+1}^{(i)}=$ Worker\_Update($\bm{x}_t^{(i)}$, $\bm{x}_t$, $\eta_l$, $I$),
            \STATE sends $\bm{d}_{t+1}^{(i)}= \bm{x}_{t}-\bm{x}_{t+1}^{(i)}$ to server.
        \ENDFOR
        \STATE \textbf{On server:}
        \STATE  $\bm{d}_{t+1}=\frac{1}{S}\sum_{i \in \mathcal{S}_t}\bm{d}_{t+1}^{(i)}$, $\bm{x}_{t+1}=\bm{x}_{t}-\eta_g \bm{d}_{t+1}$.
        \STATE Sends $\bm{x}_{t+1}$ to sampled active workers in the next round.
  \ENDFOR
  \STATE {\bfseries Output:} $ \bm{x}_T$
\end{algorithmic}
\end{algorithm}

\begin{algorithm}[t]
  \caption{GradMA-S}
  \label{GradMA-S:}
\begin{algorithmic}[1]
  \STATE {\bfseries Input:} learning rates ($\eta_l ,\eta_g$), the number of all workers $N$, the number of sampled active workers per communication round $S$, control parameters ($\beta_1$, $\beta_2$),  synchronization interval $I$ and memory size $m$ ($S \leq m \leq \min\{d, N\}$).
  \STATE Initial state $\bm{x}_0^{(i)}=\bm{x}_0\in \mathbbm{R}^d$, $\forall i \in [N]$, $\tilde{\bm{m}}_0=\bm{0}$.
  \STATE Initial $counter = \{ c(i) = 0\}, \forall i \in [N]$.
  \STATE Initial memory state $\bm{D}=\{\}$.
  \STATE $buf=\{\}$, $new\_buf=\{\}$.
  \FOR{$t=0,1,\ldots, T-1$}
        \STATE \textbf{On server:}
        \STATE Server samples a subset $\mathcal{S}_t$ with $S$ active workers and transmits $\bm{x}_t$ to $\mathcal{S}_t$.
        \STATE $counter, \bm{D}, buf, new\_buf\leftarrow$ {mem\_red} $(m, \mathcal{S}_t, $ $counter, \bm{D}, buf, new\_buf)$.
        \STATE \textbf{On workers:}
        \FOR{$i \in \mathcal{S}_t$ parallel}
            \STATE Sets $\bm{x}_{t,0}^{(i)}=\bm{x}_t$.
            \FOR{$\tau = 0,1,\ldots, I-1$}
                \STATE $\bm{x}_{t,\tau+1}^{(i)}=\bm{x}_{t,\tau}^{(i)}- \eta_l \nabla f_{i}(\bm{x}_{t, \tau}^{(i)})$.
            \ENDFOR
            \STATE Sends $\bm{d}_{t+1}^{(i)}= \bm{x}_{t}-\bm{x}_{t, I}^{(i)}$ to server.
        \ENDFOR
        \STATE \textbf{On server:}
        \STATE  $\bm{D}, \bm{x}_{t+1}, \tilde{\bm{m}}_{t+1}=$ Server\_Update($ [\bm{d}_{t+1}^{(i)}, i\in \mathcal{S}_t]$, $\Tilde{\bm{m}}_{t}$, $\bm{D}$, $\eta_g$, $\beta_1$, $\beta_2$, $buf$, $new\_buf$).
        \STATE Sends $\bm{x}_{t+1}$ to sampled active workers in the next round.
        \STATE $new\_buf=\{\}$.
  \ENDFOR
  \STATE {\bfseries Output:} $ \bm{x}_T$
\end{algorithmic}
\end{algorithm}

\subsection{Complete Empirical Study}
\label{appendix_B:}
\subsubsection{Experimental Setup}
To gauge the effectiveness of Worker\_Update$()$ and Server\_Update$()$, we perform ablation study of GradMA. 
For this purpose, we design Alg.~\ref{GradMA-W:}~(marked as GradMA-W) and Alg.~\ref{GradMA-S:}~(marked as GradMA-S), as specified in Appendix~\ref{appendix_A:}.
Meanwhile, we compare other baselines, including FedAvg~\cite{McMahan2017Communication}, 
FedProx~\cite{li2020federated1}, 
MOON~\cite{li2021model}, 
FedMLB~\cite{kim2022multi},
Scaffold~\cite{karimireddy2020scaffold}, 
FedDyn~\cite{Acar2021Federated}, 
MimeLite~\cite{Karimireddy2020Mime},
MIFA~\cite{gu2021fast}
and slow-momentum variants of FedAvg, FedProx, MIFA, MOON and FedMLB~(i.e., FedAvgM~\cite{hsu2019measuring}, FedProxM, MIFAM, MOONM and FedMLBM), 
in terms of test accuracy and communication efficiency in different FL scenarios. 
For fairness, 
we divide the baselines into three groups based on FedAvg's improvements on the worker side, server side, or both. 
Furthermore, on top of GradMA-S, we empirically study the effect of the control parameters~($\beta_1$, $\beta_2$) and verify the effectiveness of men\_red$()$ by setting varying memory sizes $m$.

\begin{figure}[H]
\centering
\includegraphics[width=0.8\textwidth]{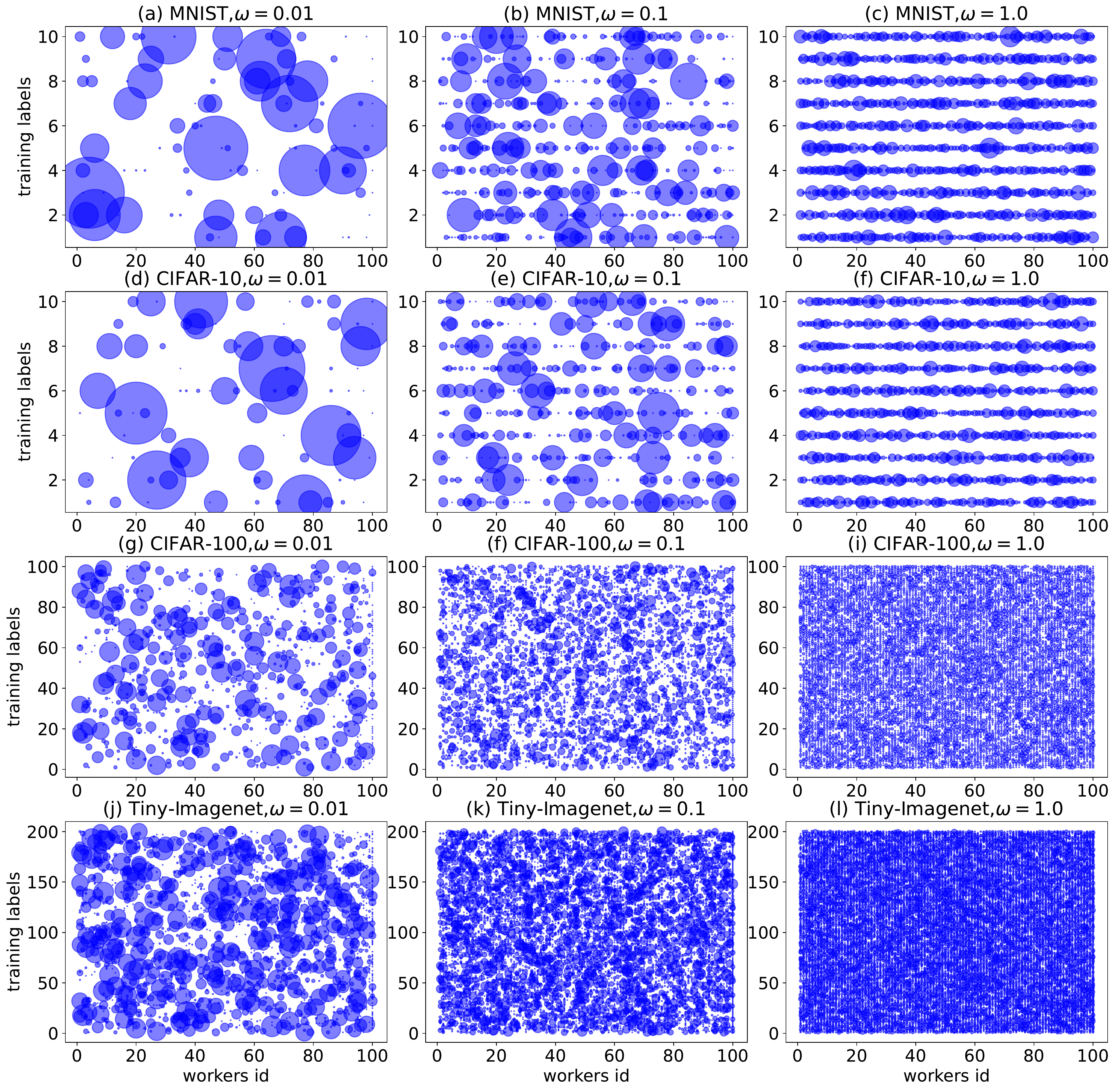} 
\caption{Data heterogeneity among workers is visualized on four datasets (MNIST, CIFAR-10, CIFAR-100 and Tiny-Imagenet), where the $x$-axis represents the workers id, the $y$-axis represents the class labels on the training set, and the size of scattered points represents the number of training samples with available labels for that worker.}
\label{data_par_sum_appendix:}
\end{figure}

All our experiments are performed on a centralized network with $100$ workers. And fix synchronization interval $I=5$.
To explore the performances of the approaches, 
we set up multiple different scenarios w.r.t. the number of sampled active workers $S$ per communication round and data heterogeneity.
Specifically, 
we set $S \in \{5, 10, 50\}$.  
Furthermore, we use Dirichlet process $Dp(\omega)$~\cite{Acar2021Federated, zhu2021data} to strictly partition the training set of each dataset across $100$ workers, where the scaling parameter $\omega$ controls the degree of data heterogeneity across workers. 
Notably, a smaller $\omega$ corresponds to higher data heterogeneity.
We set $\omega \in \{0.01, 0.1, 1.0\}$. 
A visualization of the data partitions for the four datasets at varying $\omega$ values can be found in Fig.~\ref{data_par_sum_appendix:}.
Also, the original testing set~(without partitioning) of each dataset is used to evaluate the performance of the trained centralized model.
For MNIST, a neural network (NN) with three linear hidden layers is implemented for each worker. 
We fix the total number of iterations to $2500$, i.e., $T\times I=2500$. For CIFAR-10~(CIFAR-100, Tiny-Imagenet), each worker implements a Lenet-5~\cite{lecun1998gradient}~(VGG-11~\cite{simonyan2014very}, Resnet20~\cite{he2016deep}) architecture. 
We fix the total number of iterations to $5000~(10000, 10000)$, i.e., $T\times I=5000~(10000, 10000)$.

We perform careful hyper-parameters tuning of all approaches. We set the local learning rate $\eta_l$ for each worker to $\eta_l \in \{0.001, 0.01, 0.1\}$ and the global learning rate $\eta_g$ for server to $\eta_g \in \{0.1, 1.0, 10.0\}$.  
The control parameter $\mu$ for FedProx~(FedProxM) and $\alpha$ for FedDyn are fine-tuned within $\{0.001, 0.01, 0.1\}$. 
For control parameters~($\beta_1$, $\beta_2$), we set $\beta_1, \beta_2 \in \{0.1, 0.5, 0.9\}$ unless otherwise specified.
Also, we fix memory size $m=100$ unless otherwise specified.
For the remaining tunable hyper-parameters of MOON~(MOONM) and FedMLB~(FedMLBM), we follow the settings of~\cite{li2021model} and~\cite{kim2022multi}, respectively.
For fairness, the popular SGD procedure is employed to perform local update steps for each worker.
For all experiments, we fix batch size to $64$ for all datasets. 
To ensure reliability, we report the average for each experiment over $3$ random seeds.

\subsubsection{Full Experimental Results}

\begin{figure}[H]
\centering
\includegraphics[width=1\textwidth]{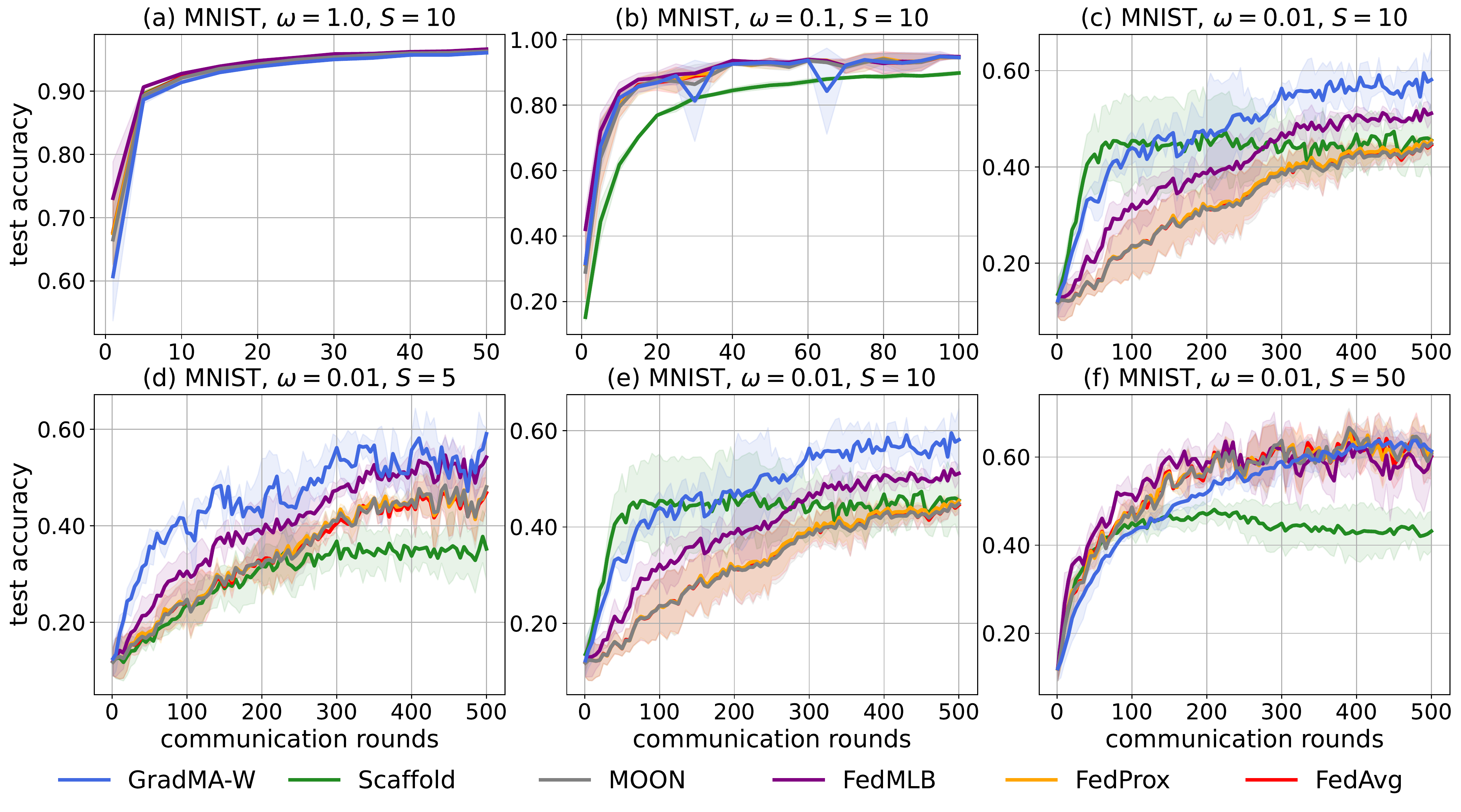} 
\caption{Full test accuracy curves for GradMA-W as well as baselines on MNIST.}
\end{figure}

\begin{figure}[H]
\centering
\includegraphics[width=1\textwidth]{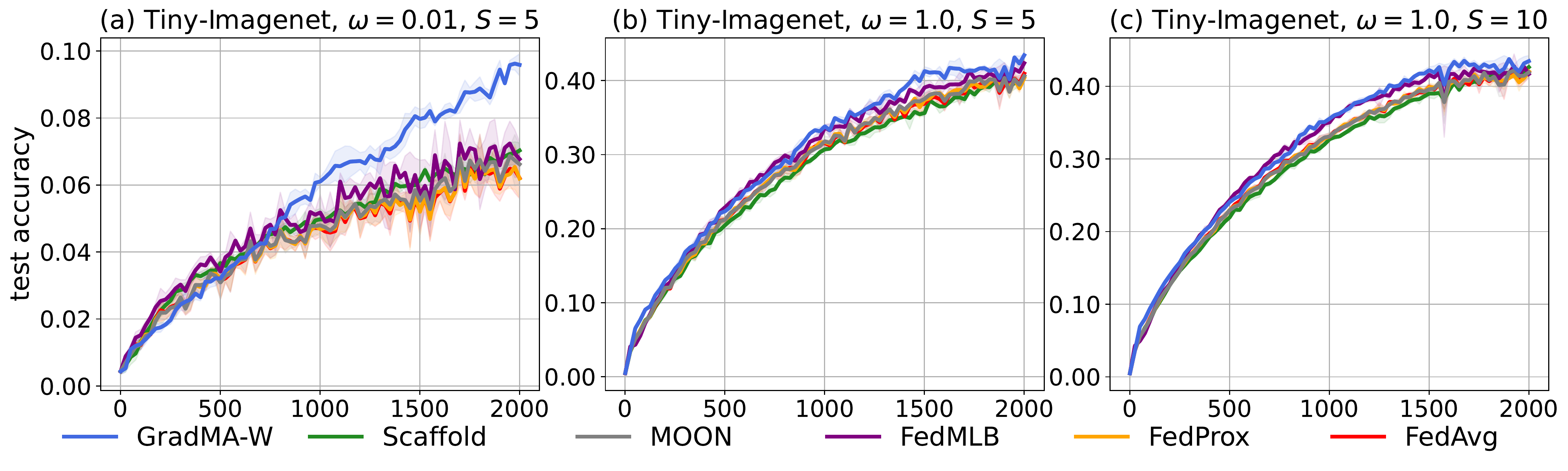} 
\caption{Full test accuracy curves for GradMA-W as well as baselines on Tiny-Imagenet.}
\end{figure}

\begin{figure}[H]
\centering
\includegraphics[width=1\textwidth]{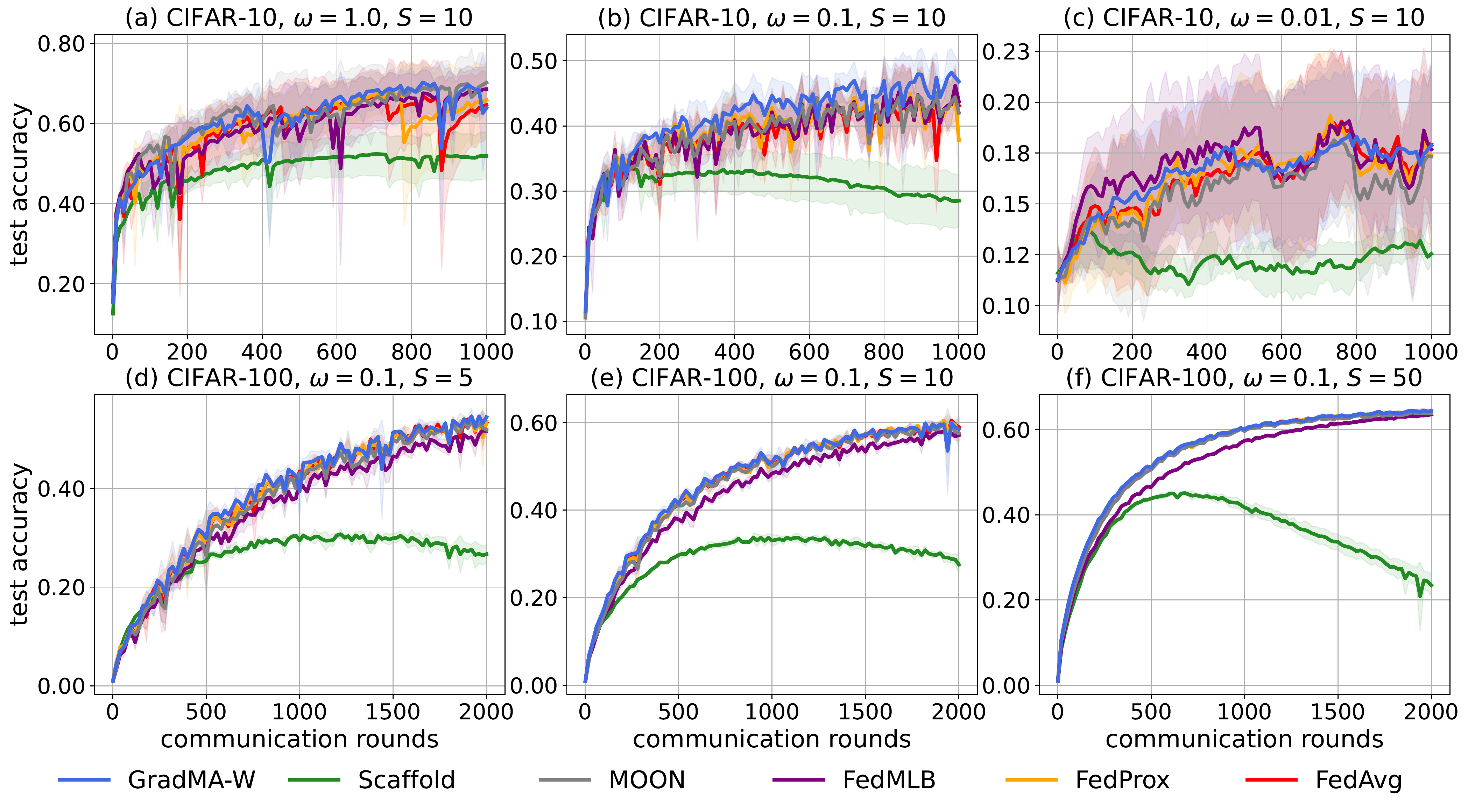} 
\caption{Full test accuracy curves for GradMA-W as well as baselines on CIFAR-10 and CIFAR-100.}
\end{figure}

\begin{figure}[H]
\centering
\includegraphics[width=1\textwidth]{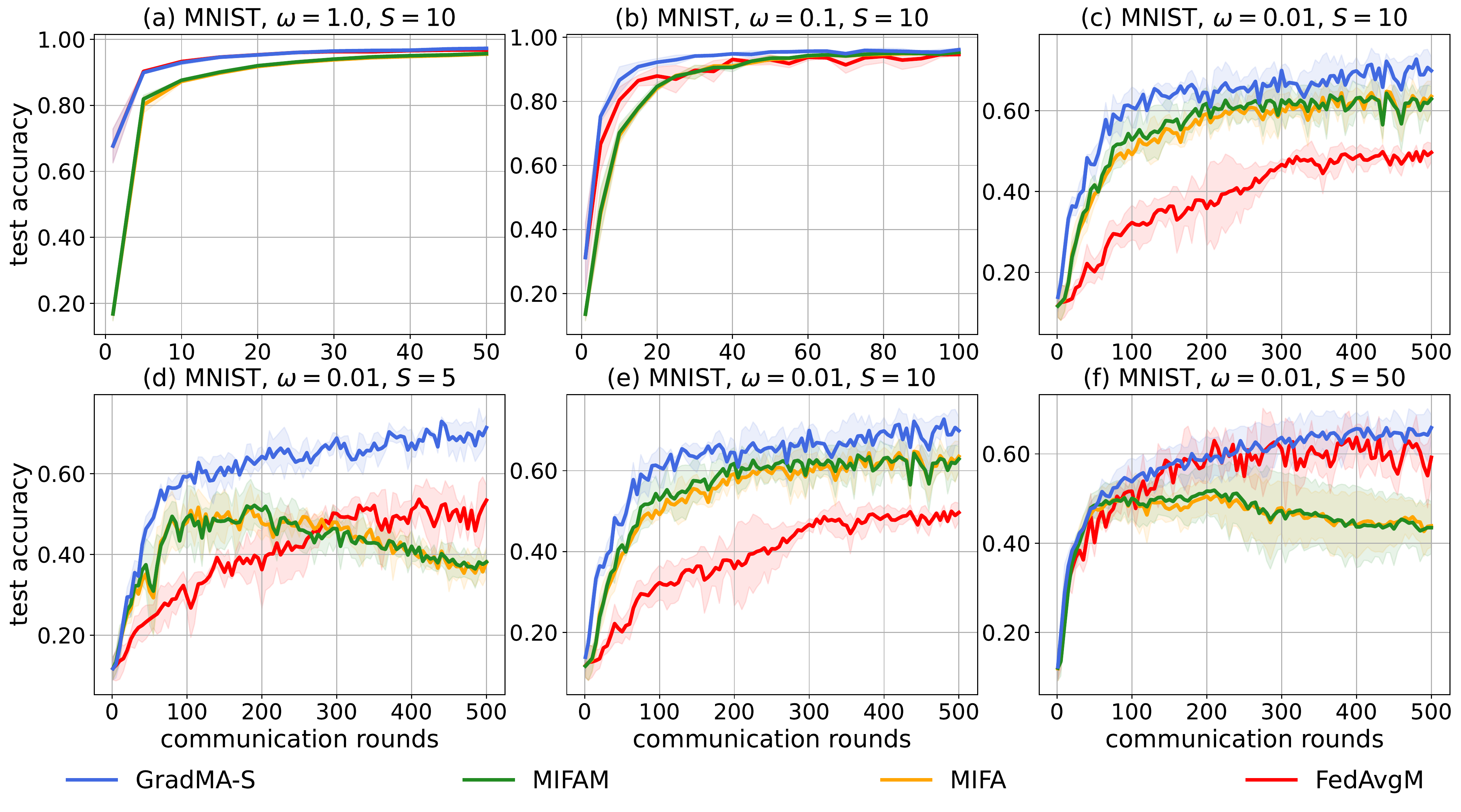} 
\caption{Full test accuracy curves for GradMA-S as well as baselines on MNIST.}
\end{figure}

\begin{figure}[H]
\centering
\includegraphics[width=1\textwidth]{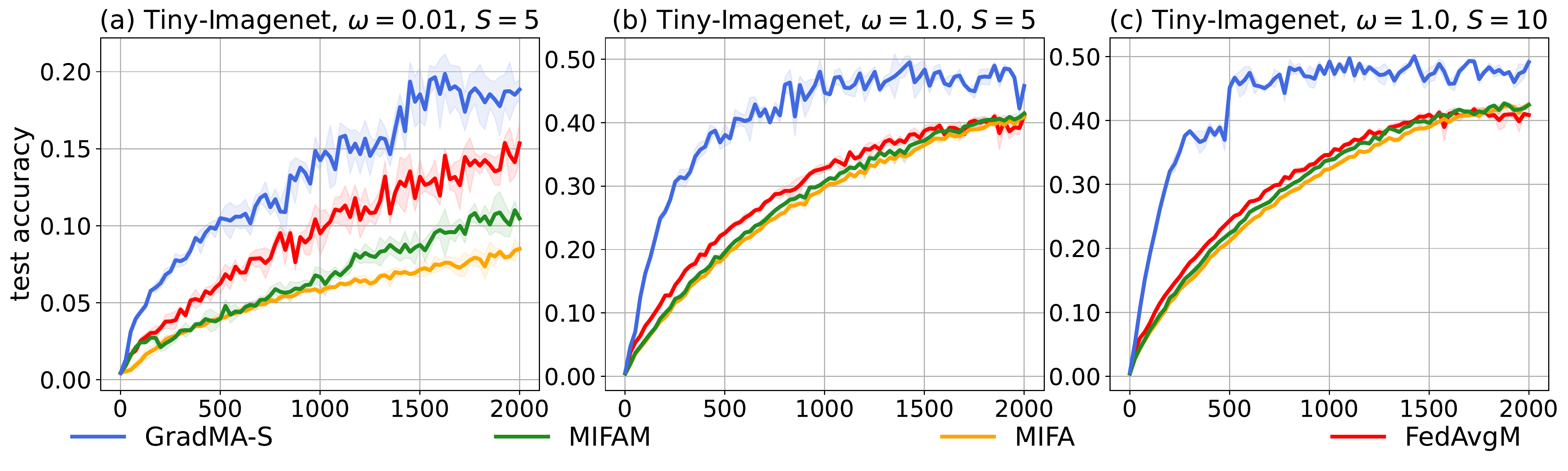} 
\caption{Full test accuracy curves for GradMA-S as well as baselines on Tiny-Imagenet.}
\end{figure}

\begin{figure}[H]
\centering
\includegraphics[width=1\textwidth]{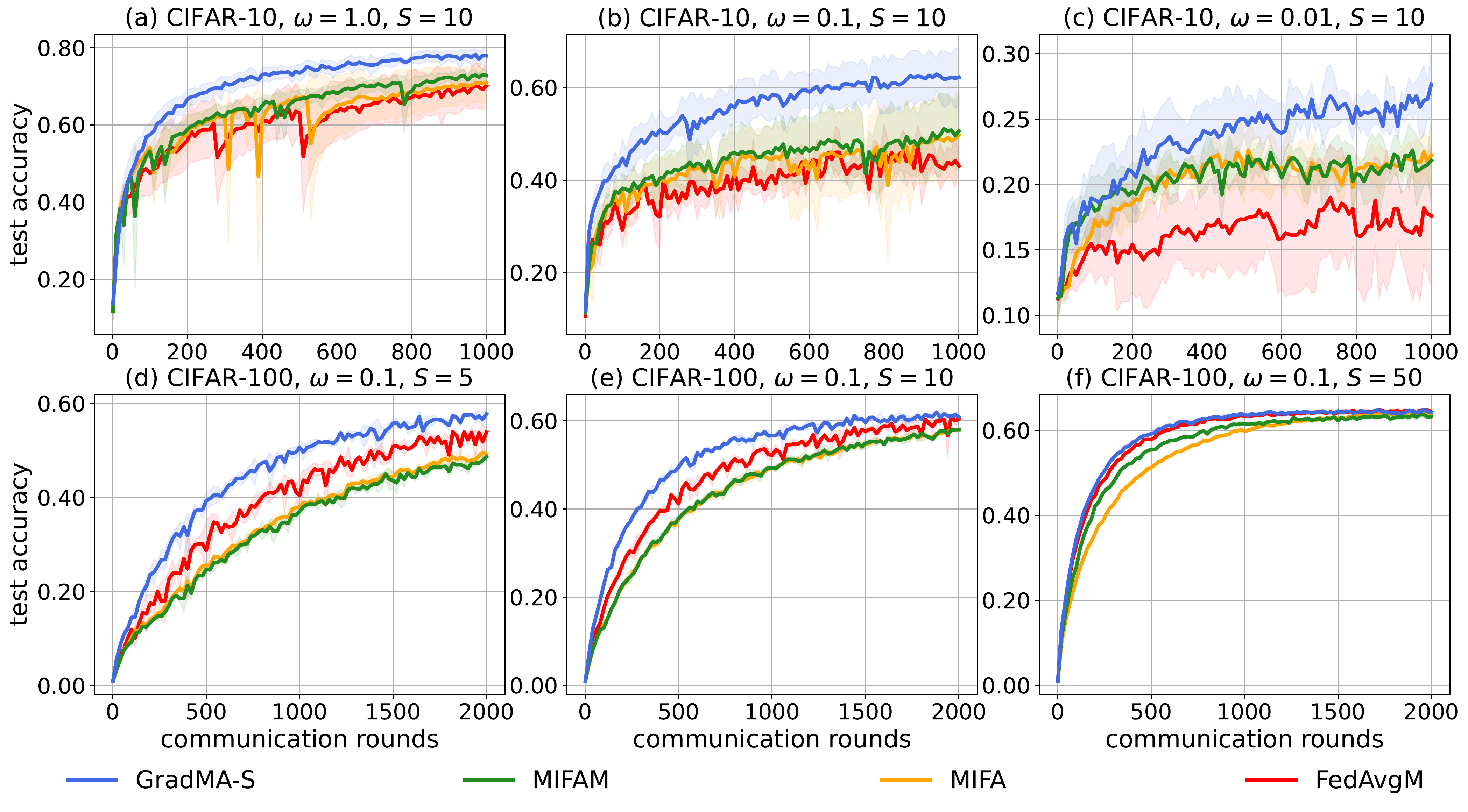} 
\caption{Full test accuracy curves for GradMA-S as well as baselines on CIFAR-10 and CIFAR-100.}
\end{figure}

\begin{figure}[H]
\centering
\includegraphics[width=1\textwidth]{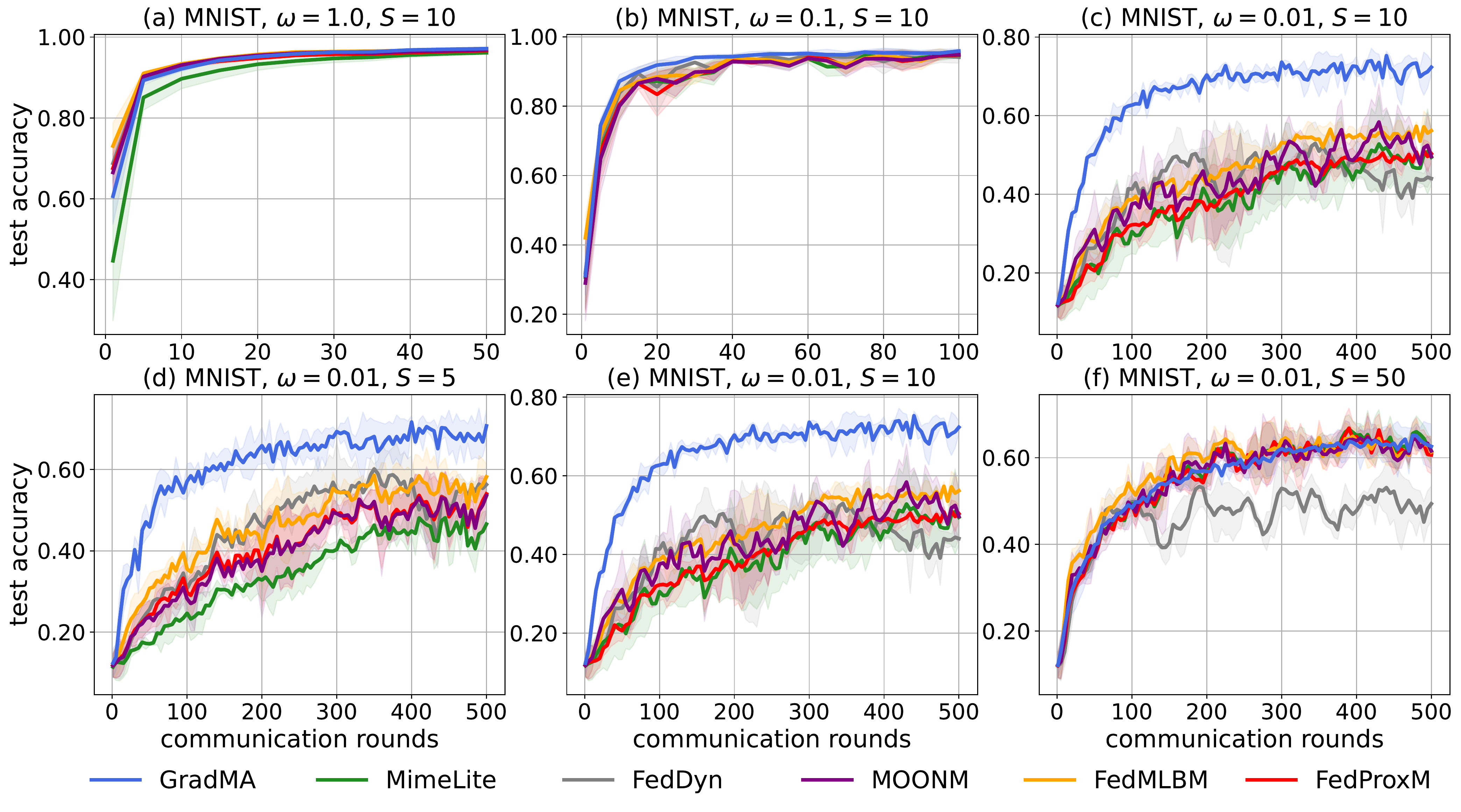} 
\caption{Full test accuracy curves for GradMA as well as baselines on MNIST.}
\end{figure}

\begin{figure}[H]
\centering
\includegraphics[width=1\textwidth]{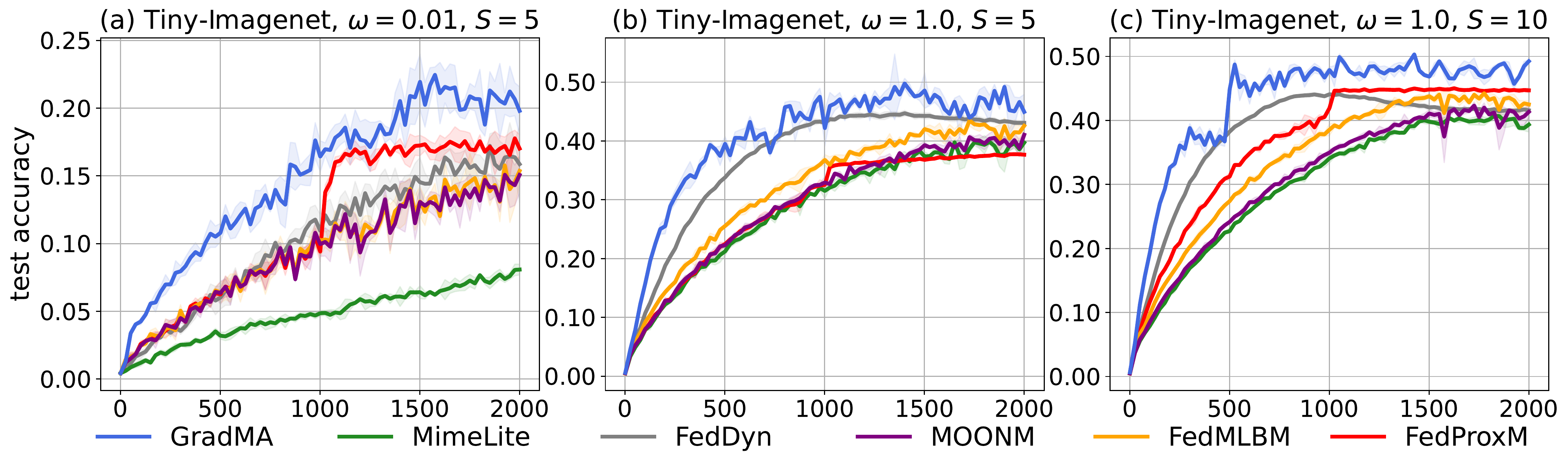} 
\caption{Full test accuracy curves for GradMA as well as baselines on Tiny-Imagenet.}
\end{figure}

\begin{figure}[H]
\centering
\includegraphics[width=1\textwidth]{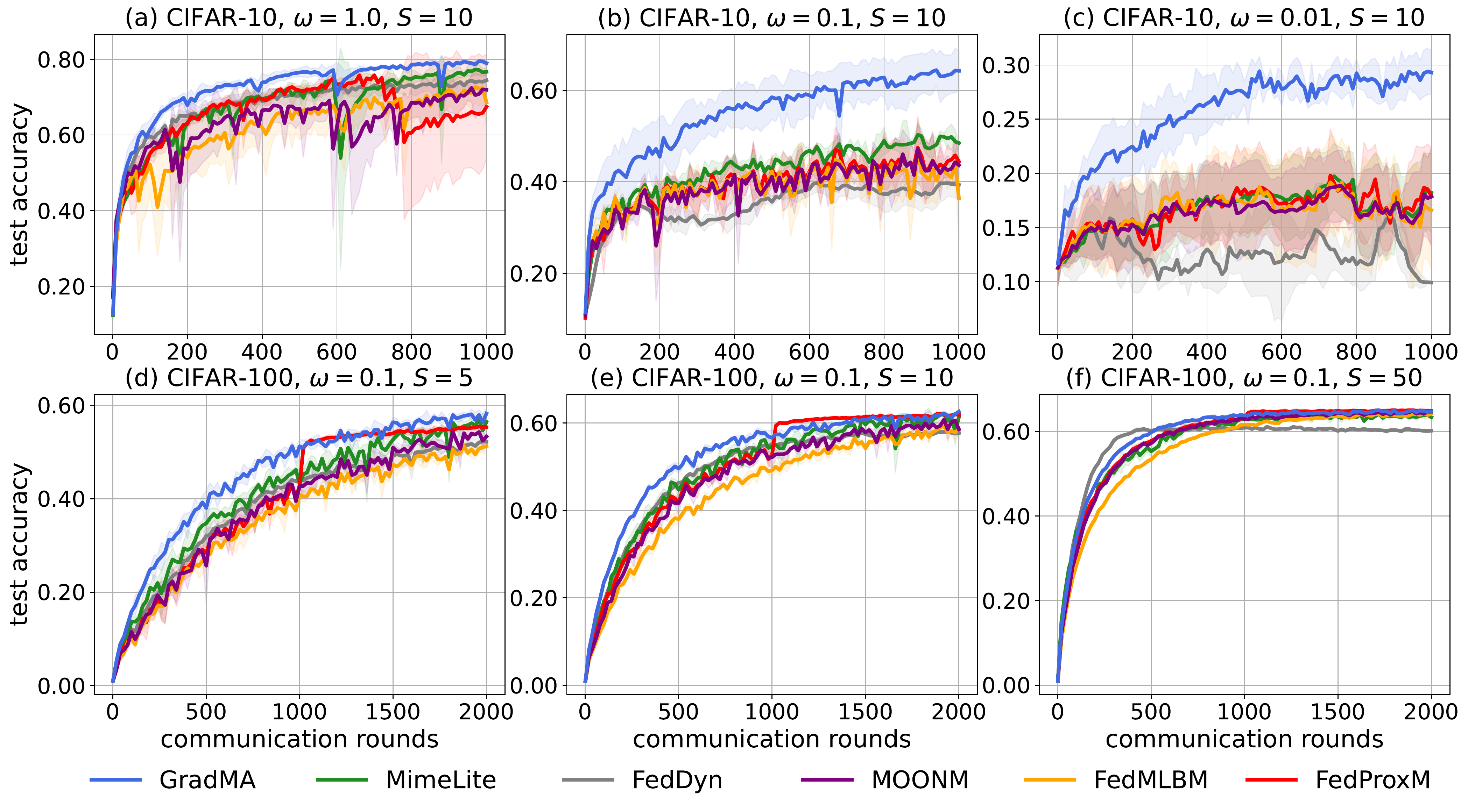} 
\caption{Full test accuracy curves for GradMA as well as baselines on CIFAR-10 and CIFAR-100.}
\end{figure}

\begin{figure}[H]
\centering
\includegraphics[width=1\textwidth]{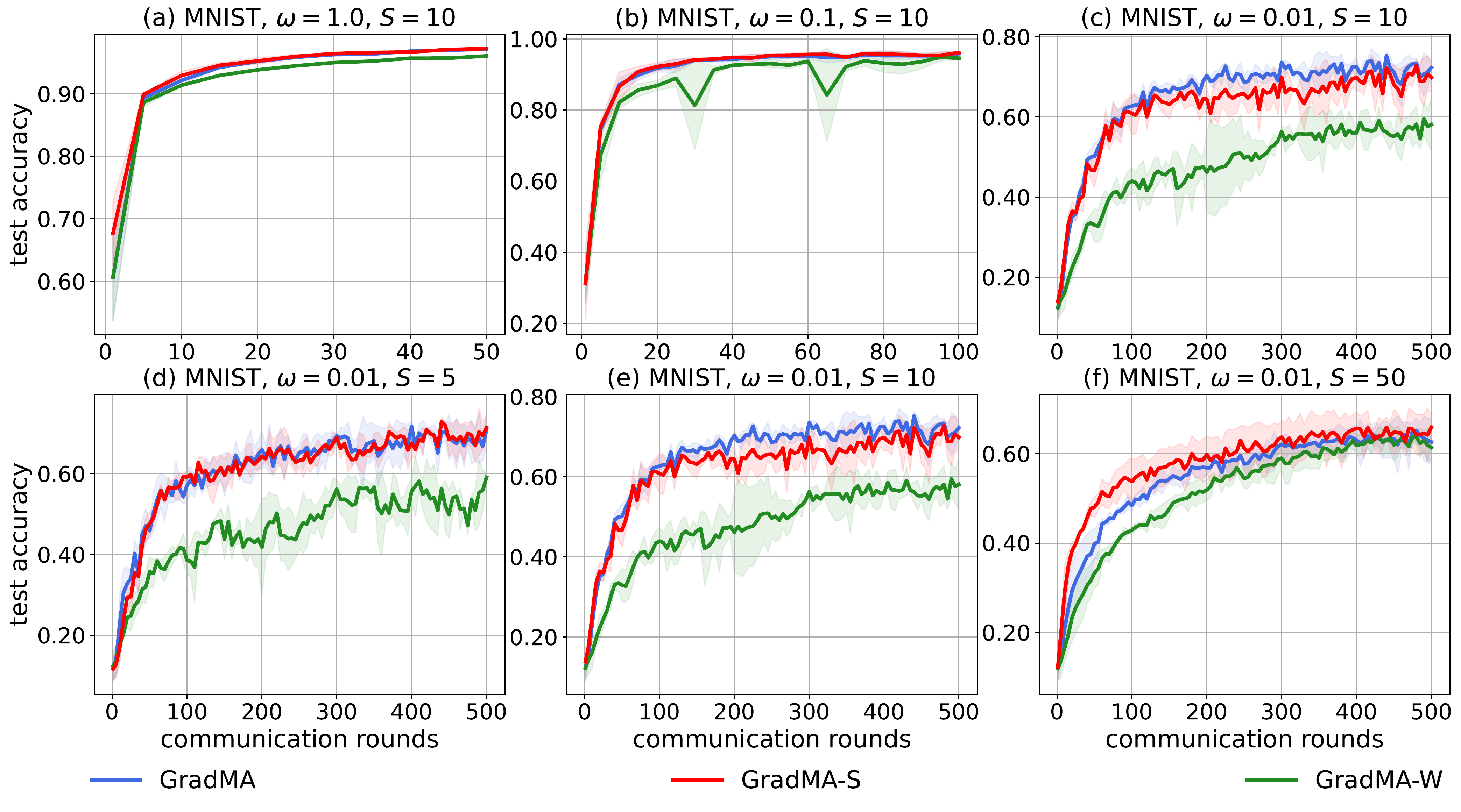} 
\caption{Full test accuracy curves for GradMA, GradMA-S and GradMA-W on MNIST.}
\end{figure}

\begin{figure}[H]
\centering
\includegraphics[width=1\textwidth]{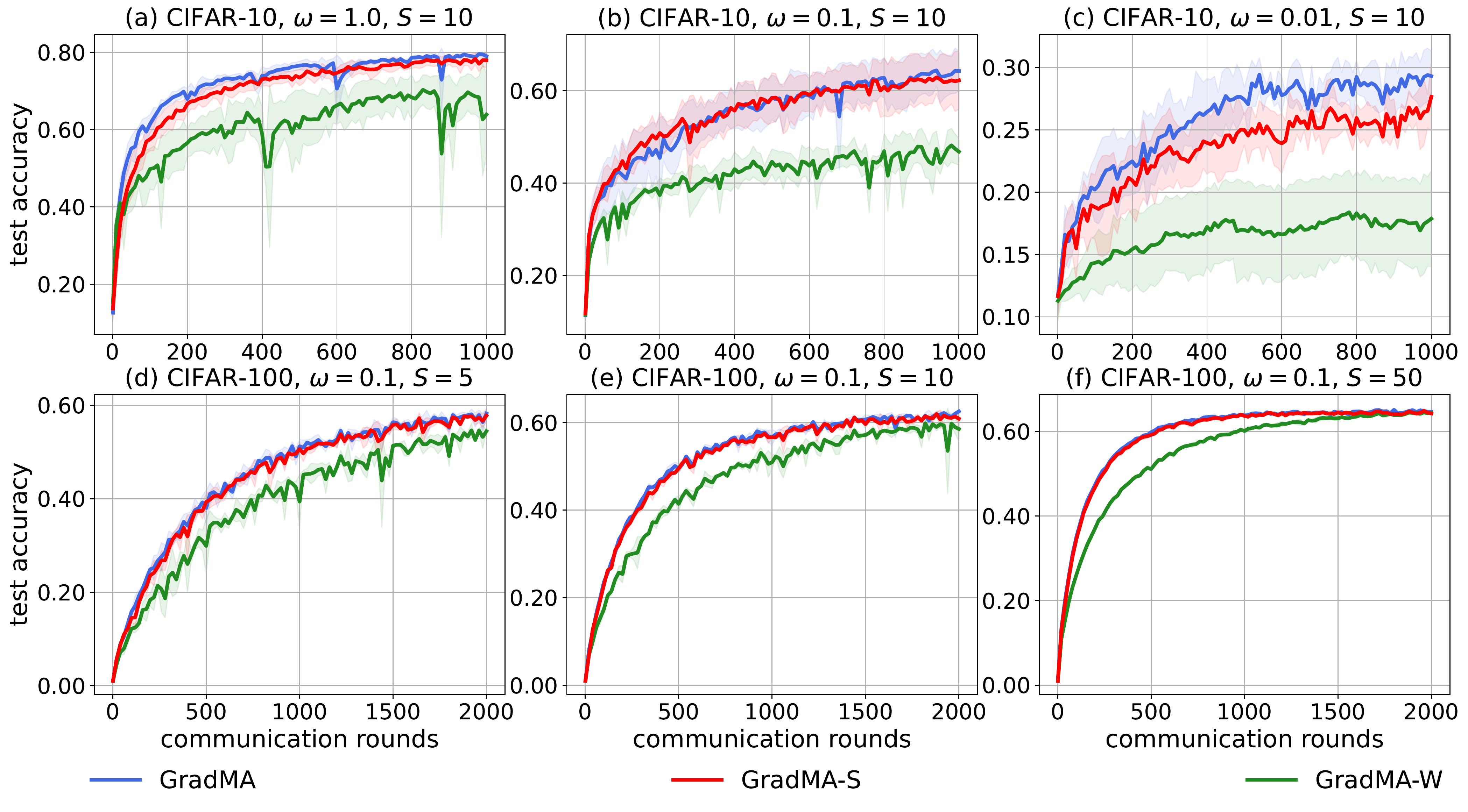} 
\caption{Full test accuracy curves for GradMA, GradMA-S and GradMA-W on CIFAR-10 and CIFAR-100.}
\end{figure}

\begin{figure}[H]
\centering
\includegraphics[width=1\textwidth]{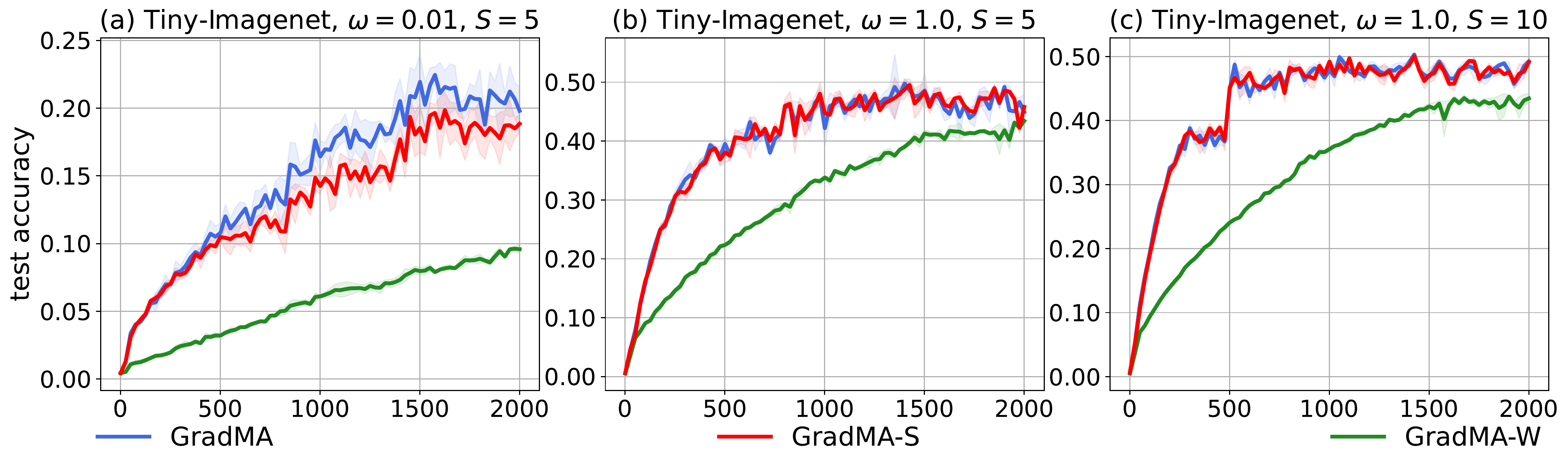} 
\caption{Full test accuracy curves for GradMA, GradMA-S and GradMA-W on Tiny-Imagenet.}
\end{figure}

\begin{figure}[H]
\centering
\includegraphics[width=1.0\textwidth]{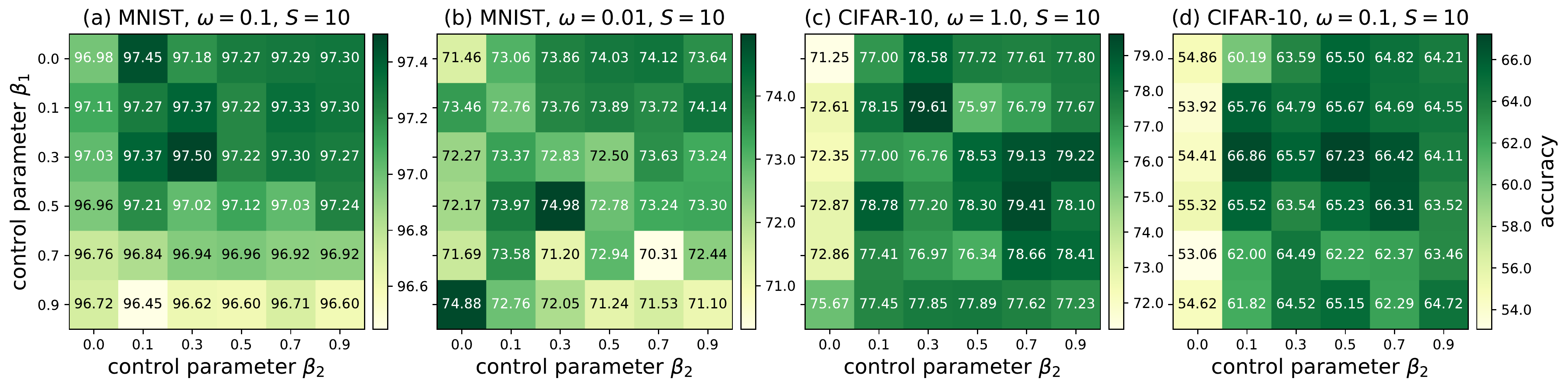} 
\caption{Top test accuracy~(\%) overview for varying control parameters ($\beta_1$,$\beta_2$) on MNIST and CIFAR-10.}
\label{beta_1_beta2_appendix:}
\end{figure}

\begin{figure}[H]
\centering
\includegraphics[width=0.7\textwidth]{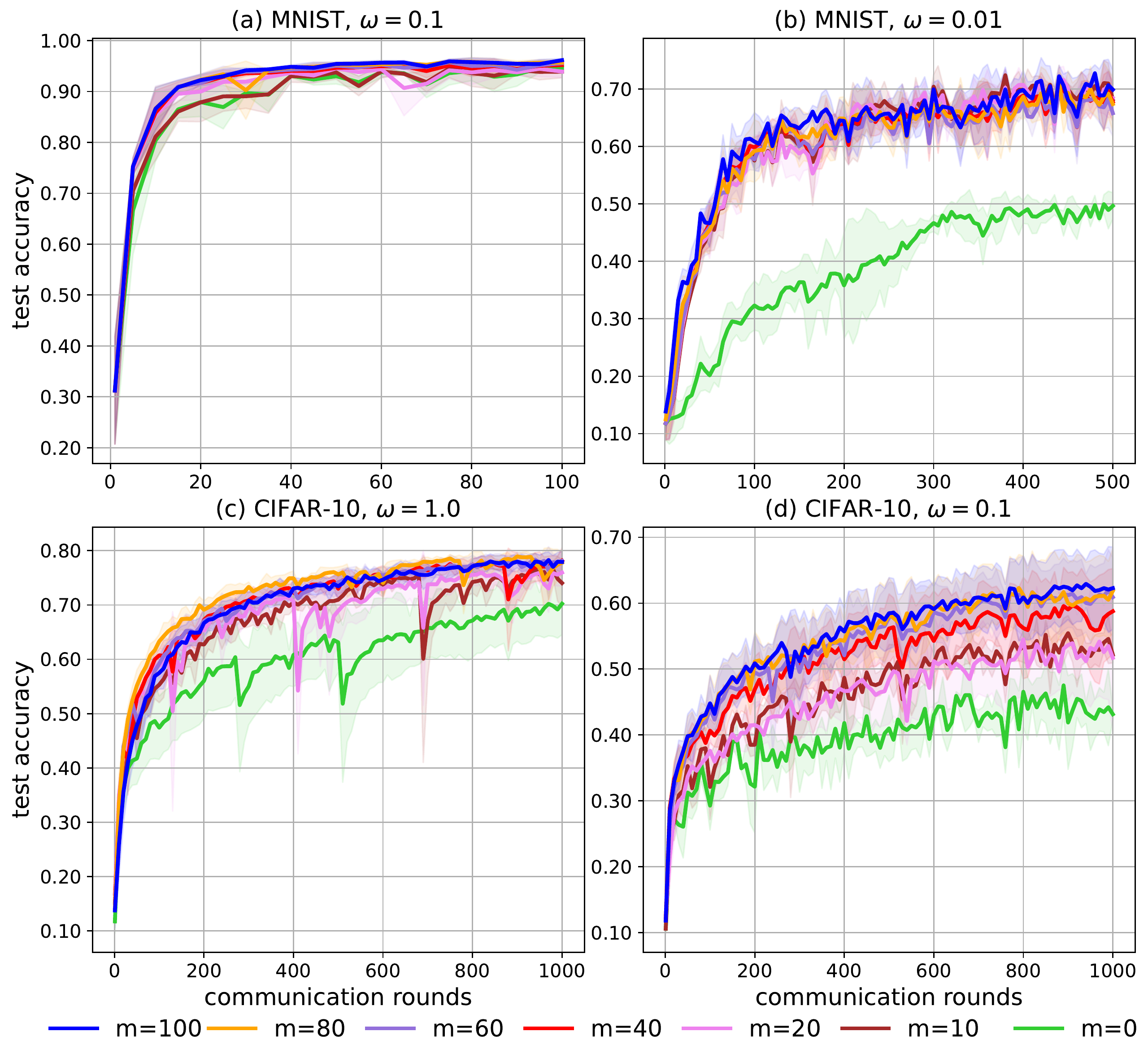} 
\caption{Full test accuracy curves for varying memory sizes $m$ on MNIST and CIFAR-10.}
\label{m_omega_comm_round:}
\end{figure}

\subsection{\quad Convergence Proof of GradMA}
\label{appendix_C:}
In this section, we provide the complete theoretical proof for convergence result of GranMA.

We first review the rule for $i$-th~($i\in [N]$) worker to update local model in Alg.~\ref{alg:1} and Alg.~\ref{local_update:}, as follows:
\begin{align}
    & \bm{g}_{t, \tau}^{(i)}=\nabla f_i(\bm{x}_{t, \tau}^{(i)}), \label{cal_g:}\\
    & \bm{G}_{t,\tau}^{(i)}=[ \nabla f_{i}(\bm{x}_{t,\tau-1}^{(i)}), \nabla f_{i}(\bm{x}_{t}), \bm{x}_{t, \tau}^{(i)}-\bm{x}_t], \label{cal_G:}\\
    & \tilde{\bm{g}}_{t,\tau}^{(i)}={\rm QP}_l(\bm{g}_{t,\tau}^{(i)}, \bm{G}_{t,\tau}^{(i)}),  \label{QP_g:}\\
    & \bm{x}_{t,\tau+1}^{(i)}=\bm{x}_{t,\tau}^{(i)}- \eta_l \tilde{\bm{g}}_{t,\tau}^{(i)}, \label{update_local_x:}
\end{align}
where $\tau \in [0, \dots, I-1]$ and $\bm{x}_{0, -1}^{(i)}=\bm{x}_{0}^{(i)}=\bm{x}_0$, $\bm{x}_{t, -1}^{(i)}=\bm{x}_{t}^{(i)}, \bm{x}_{t, 0}^{(i)}=\bm{x}_{t}$~($t>0$).

After receiving update directions sent by active workers, the server updates the centralized model according to the following update rule~(see Alg.~\ref{alg:1} and Alg.~\ref{server_update:}):
\begin{align}
    & \bm{d}_{t+1}=\frac{1}{S}\sum_{i\in \mathcal{S}_t} \bm{d}_{t+1}^{(i)}=\frac{\eta_l}{S}\sum_{i\in \mathcal{S}_t} \sum_{\tau=0}^{I-1} \tilde{\bm{g}}_{t, \tau}^{(i)}, \label{cal_d:}\\
    & \bm{m}_{t+1}=\beta_1 \tilde{\bm{m}}_{t}+\bm{d}_{t+1}, \label{cal_m:}\\
    & \tilde{\bm{m}}_{t+1}={\rm QP}_g(\bm{m}_{t+1}, \bm{D}), \label{QP_d:}\\
    & \bm{x}_{t+1} = \bm{x}_{t}- \eta_g\tilde{\bm{m}}_{t+1}, \label{update_x:}
\end{align}
where $t \in [0, \cdots, T-1]$ and $\tilde{\bm{m}}_{0} = \bm{0}$. 
Here, we omit the update rule of $\bm{D}$ in that Assumption~\ref{UBE_QP_g:} holds as long as the information contained in $\bm{D}$ is meaningful, without needing to focus on the specific content of $\bm{D}$.

Furthermore, we set $\tilde{\bm{d}}_{t+1} =\bm{d}_{t+1}+\tilde{\bm{m}}_{t+1}-\bm{m}_{t+1}$ yields:
\begin{align}
    & \tilde{\bm{m}}_{t+1}= \beta_1 \tilde{\bm{m}}_{t}+\tilde{\bm{d}}_{t+1}, \label{tilde_m_update:}\\
    & \bm{x}_{t+1} = \bm{x}_{t}- \eta_g\tilde{\bm{m}}_{t+1}. \label{update_x_1:}
\end{align}

Now, we define an auxiliary sequence such that
\begin{align}
    \bm{u}_{t}=\frac{1}{1-\beta_1} \bm{x}_{t} - \frac{\beta_1}{1-\beta_1} \bm{x}_{t-1}, \label{update_u:}
\end{align}
where $t>0$. If $t=0$ then $\bm{u}_{t}=\bm{x}_{t}$. 

\begin{lemm} 
    \label{Lemm_1:}
    Define the sequence $\{\bm{u}_t\}_{t\geq 0}$ as in Eq.~(\ref{update_u:}). According to Alg.~\ref{alg:1}, we have the following relationship
    \begin{align*}
        \bm{u}_{t+1}-\bm{u}_{t} = -\frac{\eta_g}{1-\beta_1} \tilde{\bm{d}}_{t+1}.
    \end{align*}
\end{lemm}
\begin{proof}
    Using mathematical induction on Eq.~(\ref{update_u:}), we get:
    
    case $t=0$,
    \begin{align*}
        \bm{u}_{t+1}-\bm{u}_{t} 
        & = \bm{u}_{1}-\bm{u}_{0} \\
        & = \frac{1}{1-\beta_1} \bm{x}_{1} - \frac{\beta_1}{1-\beta_1} \bm{x}_{0} -\bm{x}_{0} = \frac{1}{1-\beta_1} (\bm{x}_{1} - \bm{x}_{0})\\
        & = -\frac{\eta_g}{1-\beta_1} \tilde{\bm{m}}_{1}=-\frac{\eta_g}{1-\beta_1} \tilde{\bm{d}}_{1},
    \end{align*}
    
    and case $t>0$,
    \begin{align*}
        \bm{u}_{t+1}-\bm{u}_{t} 
        & = \frac{1}{1-\beta_1} \bm{x}_{t+1} - \frac{\beta_1}{1-\beta_1} \bm{x}_{t} - \frac{1}{1-\beta_1} \bm{x}_{t} + \frac{\beta_1}{1-\beta_1} \bm{x}_{t-1} \\
        & = \frac{1}{1-\beta_1} \left((\bm{x}_{t+1} - \bm{x}_{t})-\beta_1 (\bm{x}_{t} - \bm{x}_{t-1}) \right)\\
        & = -\frac{\eta_g}{1-\beta_1} \left(\bm{m}_{t+1} -\beta_1 \bm{m}_{t} \right)=-\frac{\eta_g}{1-\beta_1} \tilde{\bm{d}}_{t+1}.
    \end{align*}
    
    Hence, the lemma is proved.
\end{proof}

\begin{lemm}
\label{Lemm_2:} 
Under Assumptions~\ref{L_smooth:}-\ref{UBE_QP_l:}, then the following relationship   generated according to Alg.~\ref{alg:1} holds with $\eta_l \leq \frac{1}{4\sqrt{10}LI}$: for any $t\in[0,\cdots,T-1]$ and $\tau \in [0, \cdots, I-1]$, 
\begin{align*}
\frac{1}{N}\sum_{i\in[N]}\mathbbm{E}\left[\left\|\bm{x}_{t, \tau}^{(i)}-\bm{x}_{t}\right\|^2\right]   \overset{}{\leq}
 40I^2\eta_l^2\varepsilon_l^2 +  40I^2\eta_l^2\rho^2 + 40I^2\eta_l^2\mathbbm{E}\left[\|\nabla f(\bm{x}_{t})\|^2\right] ,
\end{align*}
where the expectation $\mathbbm{E}$ is w.r.t the sampled active workers per communication round.
\end{lemm}

\begin{proof}
For any worker $i\in[N]$ and $\tau \in [1, \cdots, I-1]$, we have:
\begin{align}
\mathbbm{E}&\left[\left\|\bm{x}_{t, \tau}^{(i)}-\bm{x}_{t}\right\|^2\right]\notag\\
& \overset{(a)}{=} \mathbbm{E}\left[\left\|\bm{x}_{t, \tau-1}^{(i)}-\bm{x}_{t}-\eta_l \tilde{\bm{g}}_{t, \tau-1}^{(i)}\right\|^2\right]\notag \\
& \overset{}{=} 
    \mathbbm{E}\left[\left\|\bm{x}_{t, \tau-1}^{(i)}-\bm{x}_{t}-\eta_l\left( \tilde{\bm{g}}_{t, \tau-1}^{(i)}-\bm{g}_{t, \tau-1}^{(i)}+\nabla f_i(\bm{x}_{t, \tau-1}^{(i)})-\nabla f_i(\bm{x}_{t})+\nabla f_i(\bm{x}_{t})-\nabla f(\bm{x}_{t})+\nabla f(\bm{x}_{t})\right)\right\|^2\right]\notag\\
& \overset{(b)}{\leq}
    \left(1+\frac{1}{2I-1}\right)\mathbbm{E}\left[\|\bm{x}_{t, \tau-1}^{(i)}-\bm{x}_{t}\|^2\right]+8I\eta_l^2\bigg[\mathbbm{E}\left[\|\tilde{\bm{g}}_{t, \tau-1}^{(i)}-\bm{g}_{t, \tau-1}^{(i)}\|^2\right] + \mathbbm{E}\left[\|\nabla f_i(\bm{x}_{t, \tau-1}^{(i)})-\nabla f_i(\bm{x}_{t})\|^2\right] \notag\\ 
    & \quad + \mathbbm{E}\left[\|\nabla f_i(\bm{x}_{t})-\nabla f(\bm{x}_{t})\|^2\right] + \mathbbm{E}\left[\|\nabla f(\bm{x}_{t})\|^2\right]\bigg]\notag\\
& \overset{}{\leq}
    \left(1+\frac{1}{2I-1}+8I\eta_l^2L^2\right)\mathbbm{E}\left[\|\bm{x}_{t, \tau-1}^{(i)}-\bm{x}_{t}\|^2\right]+8I\eta_l^2\varepsilon_l^2 +  8I\eta_l^2\rho^2 + 8I\eta_l^2\mathbbm{E}\left[\|\nabla f(\bm{x}_{t})\|^2\right]\notag\\
& \overset{(c)}{\leq}
    \left(1+\frac{1}{I-1}\right)\mathbbm{E}\left[\|\bm{x}_{t, \tau-1}^{(i)}-\bm{x}_{t}\|^2\right]+8I\eta_l^2\varepsilon_l^2 +  8I\eta_l^2\rho^2 + 8I\eta_l^2\mathbbm{E}\left[\|\nabla f(\bm{x}_{t})\|^2\right], \label{Lem_2_1:}
\end{align}
where ($a$) holds by using the Eq.~(\ref{update_local_x:}),  ($b$) follows from the inequalities $\|\bm{x}\pm\bm{y}\|^2 \leq(1+\frac{1}{2I-1})\|\bm{x}\|^2+2I\|\bm{y}\|^2, \bm{x}, \bm{y} \in \mathbbm{R}^d$ and $\left\|\sum_{i=1}^{N}\bm{x}_i\right\|^2 \leq N\sum_{i=1}^{N}\|\bm{x}_i\|^2, \bm{x}_i \in \mathbbm{R}^d$, 
and ($c$) holds by using the fact that $\frac{1}{I-1} \geq \frac{1}{2I-1}+8I\eta_l^2L^2$ holds if $\eta_l \leq \frac{1}{4\sqrt{10}LI}$.

Then, recursively unrolling inequality~(\ref{Lem_2_1:}), we get:
\begin{align*}
\frac{1}{N}\sum_{i\in[N]}\mathbbm{E}\left[\left\|\bm{x}_{t, \tau}^{(i)}-\bm{x}_{t}\right\|^2\right]
& \overset{}{\leq}
    \sum_{k=0}^{\tau-1}\left(1+\frac{1}{I-1}\right)^k\left[8I\eta_l^2\varepsilon_l^2 +  8I\eta_l^2\rho^2 + 8I\eta_l^2\mathbbm{E}\left[\|\nabla f(\bm{x}_{t})\|^2\right]\right]\\
& \overset{}{\leq}
    (I-1)\left[\left(1+\frac{1}{I-1}\right)^{I} -1\right]\left[8I\eta_l^2\varepsilon_l^2 +  8I\eta_l^2\rho^2 + 8I\eta_l^2\mathbbm{E}\left[\|\nabla f(\bm{x}_{t})\|^2\right]\right]\\
& \overset{(a)}{\leq}
    5I\left[8I\eta_l^2\varepsilon_l^2 +  8I\eta_l^2\rho^2 + 8I\eta_l^2\mathbbm{E}\left[\|\nabla f(\bm{x}_{t})\|^2\right]\right]\\
& \overset{}{\leq}
    40I^2\eta_l^2\varepsilon_l^2 +  40I^2\eta_l^2\rho^2 + 40I^2\eta_l^2\mathbbm{E}\left[\|\nabla f(\bm{x}_{t})\|^2\right],
\end{align*}
where ($a$) holds by using $\sum_{k=0}^{\tau-1}\left(1+\frac{1}{I-1}\right)^k=\frac{1-\left(1+\frac{1}{I-1}\right)^\tau}{1-\left(1+\frac{1}{I-1}\right)}=(I-1)\left(\left(1+\frac{1}{I-1}\right)^\tau-1\right)\leq(I-1)\left(\left(1+\frac{1}{I-1}\right)^I-1\right)\leq 5I$.

So far, we complete the proof.
\end{proof}

\begin{lemm}
\label{Lemm_3:}
Under Assumptions~\ref{L_smooth:}-\ref{UBE_QP_l:},
then the following relationship generated according to Alg.~\ref{alg:1} holds: for any $t\in[0,\cdots,T-1]$,
\begin{align*}
    \frac{1}{N}\sum_{i\in [N]}\mathbbm{E}\left[\left\|\sum_{\tau=0}^{I-1}\tilde{\bm{g}}_{t, \tau}^{(i)}\right\|^2\right]  \overset{}{\leq} 4I^2(\varepsilon_l^2+\rho^2)(1+40I^2\eta_l^2L^2) + 4I^2(1+40I^2\eta_l^2L^2)\mathbbm{E}\left[\left\|\nabla f(\bm{x}_{t})\right\|^2\right],
\end{align*}
where the expectation $\mathbbm{E}$ is w.r.t the sampled active workers per communication round.
\end{lemm}

\begin{proof}

\begin{align*}
\frac{1}{N}\sum_{i\in [N]}&\mathbbm{E}\left[\left\|\sum_{\tau=0}^{I-1}\tilde{\bm{g}}_{t, \tau}^{(i)}\right\|^2\right] \\
& \overset{(a)}{\leq} I\frac{1}{N}\sum_{i\in [N]}\sum_{\tau=0}^{I-1}\mathbbm{E}\left[\left\|\tilde{\bm{g}}_{t, \tau}^{(i)}\right\|^2\right] \\
& \overset{}{\leq}
    I\frac{1}{N}\sum_{i\in [N]}\sum_{\tau=0}^{I-1}\mathbbm{E}\left[\left\|\tilde{\bm{g}}_{t, \tau}^{(i)}-\bm{g}_{t, \tau}^{(i)}+\nabla f_i(\bm{x}_{t,\tau}^{(i)})-\nabla f_i(\bm{x}_{t})+\nabla f_i(\bm{x}_{t})-\nabla f(\bm{x}_{t})+\nabla f(\bm{x}_{t})\right\|^2\right]\\
& \overset{(b)}{\leq}
    4I\frac{1}{N}\sum_{i\in [N]}\sum_{\tau=0}^{I-1}\left[\left\|\tilde{\bm{g}}_{t, \tau}^{(i)}-\bm{g}_{t, \tau}^{(i)}\right\|^2 + \mathbbm{E}\left[\left\|\nabla f_i(\bm{x}_{t,\tau}^{(i)})-\nabla f_i(\bm{x}_{t})\right\|^2\right] +\left\|\nabla f_i(\bm{x}_{t})-\nabla f(\bm{x}_{t})\right\|^2 +  \mathbbm{E}\left[\left\|\nabla f(\bm{x}_{t})\right\|^2\right] \right]\\
& \overset{}{\leq}
    4I\frac{1}{N}\sum_{i\in [N]}\sum_{\tau=0}^{I-1}\left[\varepsilon_l^2 + L^2\mathbbm{E}\left[\left\|\bm{x}_{t,\tau}^{(i)}-\bm{x}_{t}\right\|^2\right] +\rho^2 +  \mathbbm{E}\left[\left\|\nabla f(\bm{x}_{t})\right\|^2\right] \right]\\
& \overset{}{\leq}
    4I^2\left(\varepsilon_l^2 +\rho^2 + \mathbbm{E}\left[\left\|\nabla f(\bm{x}_{t})\right\|^2\right]\right) + 4IL^2\frac{1}{N}\sum_{i\in [N]}\sum_{\tau=0}^{I-1} \mathbbm{E}\left[\left\|\bm{x}_{t,\tau}^{(i)}-\bm{x}_{t}\right\|^2\right] \\
& \overset{(c)}{\leq}
    4I^2\left(\varepsilon_l^2 +\rho^2 + \mathbbm{E}\left[\left\|\nabla f(\bm{x}_{t})\right\|^2\right]\right) + 4I^2L^2 \left(40I^2\eta_l^2\varepsilon_l^2 +  40I^2\eta_l^2\rho^2 + 40I^2\eta_l^2\mathbbm{E}\left[\|\nabla f(\bm{x}_{t})\|^2\right]\right)\\
& \overset{}{\leq}
    4I^2(\varepsilon_l^2+\rho^2)(1+40I^2\eta_l^2L^2) + 4I^2(1+40I^2\eta_l^2L^2)\mathbbm{E}\left[\left\|\nabla f(\bm{x}_{t})\right\|^2\right]
\end{align*}
where ($a$) and ($b$) result from the fact that  $\left\|\sum_{i=1}^{N}\bm{x}_i\right\|^2 \leq N\sum_{i=1}^{N}\|\bm{x}_i\|^2, \bm{x}_i \in \mathbbm{R}^d$, and ($c$) uses the statement from Lemma~\ref{Lemm_2:}.

So far, the lemma is proved.
\end{proof}

\begin{lemm}
\label{Lemm_4:}
Under Assumption~\ref{UBE_QP_g:}, then the following relationship generated according to Alg.~\ref{alg:1} holds: for any $t\in[0,\cdots,T-1]$,
\begin{align*}
    \mathbbm{E}\left[\left\|\tilde{\bm{d}}_{t+1}\right\|^2\right]  \overset{}{\leq} \frac{2\eta_l^2}{S^2}\mathbbm{E}\left[\left\|\sum_{i\in [N]}\mathbbm{I}\{i\in \mathcal{S}_t\}\sum_{\tau=0}^{I-1}\tilde{\bm{g}}_{t, \tau}^{(i)}\right\|^2\right]+\frac{2\varepsilon_g^2}{1-\beta_2},
\end{align*}
where the expectation $\mathbbm{E}$ is w.r.t the sampled active workers per communication round.
\end{lemm}

\begin{proof}

\begin{align*}
\mathbbm{E}\left[\left\|\tilde{\bm{d}}_{t+1}\right\|^2\right] & \overset{(a)}{\leq} \mathbbm{E}\left[\left\|\bm{d}_{t+1}+\tilde{\bm{m}}_{t+1}-\bm{m}_{t+1}\right\|^2\right]\\
& \overset{}{\leq}
    2\mathbbm{E}\left[\left\|\bm{d}_{t+1}\right\|^2\right]+2\mathbbm{E}\left[\left\|\tilde{\bm{m}}_{t+1}-\bm{m}_{t+1}\right\|^2\right]\\
& \overset{(b)}{\leq}
    2\mathbbm{E}\left[\left\|\frac{1}{S}\sum_{i\in \mathcal{S}_t}\bm{d}_{t+1}^{(i)}\right\|^2\right]+\frac{2\varepsilon_g^2}{1-\beta_2}\\
& \overset{}{=}
    \frac{2\eta_l^2}{S^2}\mathbbm{E}\left[\left\|\sum_{i\in [N]}\mathbbm{I}\{i\in \mathcal{S}_t\}\sum_{\tau=0}^{I-1}\tilde{\bm{g}}_{t, \tau}^{(i)}\right\|^2\right]+\frac{2\varepsilon_g^2}{1-\beta_2},
\end{align*}
where ($a$) uses the fact that $\tilde{\bm{d}}_{t+1} = \bm{d}_{t+1}+\tilde{\bm{m}}_{t+1}-\bm{m}_{t+1}$, and ($b$) results from the Eq.~(\ref{cal_d:}).
Hence, the lemma is proved.
\end{proof}

\begin{lemm} 
    \label{Lemm_5:}
    Define the sequence $\{\bm{u}_t\}_{t\geq 0}$ as in Eq.~(\ref{update_u:}), the following relationship generated according to Alg.~\ref{alg:1} holds: for any $t \in [0,\cdots, T-1]$, 
    \begin{align*}
        \sum_{t=0}^{T-1}\mathbbm{E}\left[\left\|\bm{u}_{t} - \bm{x}_{t}\right\|^2\right]
        & \overset{}{\leq}
          \frac{2\beta_1^2\eta_g^2\eta_l^2}{(1-\beta_1)^4S^2}  \sum_{t=0}^{T-1}\mathbbm{E}\left[\left\|\sum_{i\in [N]}\mathbbm{I}\{i\in \mathcal{S}_t\}\sum_{\tau=0}^{I-1}\tilde{\bm{g}}_{t, \tau}^{(i)}\right\|^2\right]+\frac{2\beta_1^2\eta_g^2\varepsilon_g^2T}{(1-\beta_1)^4(1-\beta_2)},
    \end{align*}
    where the expectation $\mathbbm{E}$ is w.r.t the sampled active workers per communication round.
\end{lemm}
\begin{proof}
    Recursively applying Eq.~(\ref{tilde_m_update:}) to achieve the update rule for $\tilde{\bm{m}}_{t}$ yields: 
    \begin{align}
        \label{u_x_1:}
        \tilde{\bm{m}}_{t}
        & \overset{(a)}{=} \sum_{k=1}^{t}\beta_1^{t-k}\tilde{\bm{d}}_k, \forall t \geq 1,
    \end{align}
    where ($a$) holds by $\tilde{\bm{m}}_0=\bm{0}$.
    Furthermore, building on equations (\ref{update_u:}) and (\ref{u_x_1:}), we get:
    \begin{align}
        \label{u_x_2:}
        \bm{u}_{t} - \bm{x}_{t}
        & = \frac{\beta_1}{1-\beta_1} \left(\bm{x}_{t} - \bm{x}_{t-1}\right)=-\frac{\beta_1\eta_g}{1-\beta_1} \tilde{\bm{m}}_{t}=-\frac{\beta_1\eta_g}{1-\beta_1} \sum_{k=1}^{t}\beta_1^{t-k}\tilde{\bm{d}}_k.
    \end{align}
    Now, we define $z_t=\sum_{k=1}^{t}\beta_1^{t-k}=\frac{1-\beta_1^t}{1-\beta_1}, \forall t\geq1$.  Using Eq. (\ref{u_x_2:}) we obtain:
    \begin{align}
        \mathbbm{E}\left[\left\|\bm{u}_{t} - \bm{x}_{t}\right\|^2\right]
        & \overset{}{=}
            \frac{\beta_1^2\eta_g^2}{(1-\beta_1)^2}z_t^2 \mathbbm{E}\left[\left\|\sum_{k=1}^{t}\frac{\beta_1^{t-k}}{z_t}\tilde{\bm{d}}_k\right\|^2\right]\notag\\
        & \overset{(a)}{\leq}
            \frac{\beta_1^2\eta_g^2}{(1-\beta_1)^2}z_t \sum_{k=1}^{t}\beta_1^{t-k}\mathbbm{E}\left[\left\|\tilde{\bm{d}}_k\right\|^2\right]\notag\\
        & \overset{(b)}{\leq}
            \frac{\beta_1^2\eta_g^2}{(1-\beta_1)^3} \sum_{k=1}^{t}\beta_1^{t-k}\mathbbm{E}\left[\left\|\tilde{\bm{d}}_k\right\|^2\right], \label{u_x_3:}
    \end{align}
    where ($a$) follows from the fact that $\|\sum_{k=1}^{T}\frac{c_k}{c}\bm{a}_k\|^2\leq\sum_{k=1}^{T}\frac{c_i}{c}\|\bm{a}_k\|^2( \bm{a}_k\in\mathbbm{R}^d)$ holds if $c=\sum_{k=1}^{T}c_k$, and ($b$) results from $1-\beta^t\leq 1$.
    
    Next, summing Eq.~(\ref{u_x_3:}) over $t\in \{0, \cdots, T-1\}(T\geq1)$, we have:
    \begin{align}
        \sum_{t=0}^{T-1}\mathbbm{E}\left[\left\|\bm{u}_{t} - \bm{x}_{t}\right\|^2\right]
        & \overset{(a)}{\leq}
            \frac{\beta_1^2\eta_g^2}{(1-\beta_1)^3} \sum_{t=1}^{T-1}\sum_{k=1}^{t}\beta_1^{t-k}\mathbbm{E}\left[\left\|\tilde{\bm{d}}_k\right\|^2\right] \notag\\
        & \overset{}{=}
            \frac{\beta_1^2\eta_g^2}{(1-\beta_1)^3} \left(\sum_{t=k}^{T-1}\beta_1^{t-k}\right) \sum_{k=1}^{T-1}\mathbbm{E}\left[\left\|\tilde{\bm{d}}_k\right\|^2\right]
            \notag\\
        & \overset{(b)}{\leq}
            \frac{\beta_1^2\eta_g^2}{(1-\beta_1)^4}  \sum_{t=1}^{T-1}\mathbbm{E}\left[\left\|\tilde{\bm{d}}_t\right\|^2\right] \notag\\
        & \overset{(c)}{\leq}
             \frac{2\beta_1^2\eta_g^2\eta_l^2}{(1-\beta_1)^4S^2}  \sum_{t=0}^{T-1}\mathbbm{E}\left[\left\|\sum_{i\in [N]}\mathbbm{I}\{i\in \mathcal{S}_t\}\sum_{\tau=0}^{I-1}\tilde{\bm{g}}_{t, \tau}^{(i)}\right\|^2\right]+\frac{2\beta_1^2\eta_g^2\varepsilon_g^2T}{(1-\beta_1)^4(1-\beta_2)},
    \end{align}
    where ($a$) uses $\bm{u}_0=\bm{x}_0$,  ($b$) holds by using the inequality $\sum_{t=k}^{T-1}\beta_1^{t-k}=\frac{1-\beta_1^t}{1-\beta_1}\leq\frac{1}{1-\beta_1}$, and ($c$) follows from the statement of Lemma~\ref{Lemm_4:}.
    
    Hence, the lemma is proved.
\end{proof}

\begin{lemm}
\label{Lemm_6:}
According to Assumptions~\ref{Global_Function_Below_Bounds:}-\ref{UBE_QP_g:} and setting $\eta_l \leq \frac{1}{4\sqrt{10}LI}$, $\eta_g\eta_l\leq\frac{(1-\beta_1)^2S(N-1)}{IL(\beta_1S(N-1)+4N(S-1))}$ and $ 320I^2\eta_l^2L^2+\frac{64I\eta_g\eta_lL(1+40I^2\eta_l^2L^2)}{(1-\beta_1)^2}\frac{N-S}{S(N-1)}\leq 1$, then the iterates generated by Alg.~\ref{alg:1} satisfy: for all $t \in [0,\cdots, T-1]$,
\begin{align*}
    \frac{1}{T}\sum_{t=0}^{T-1}\mathbbm{E}\left[\left\|\nabla f(\bm{x}_{t})\right\|^2\right]
    & \overset{}{\leq} 
         \frac{8(1-\beta_1)(f(\bm{x}_{0}) -f^\star)}{I\eta_g\eta_lT} +8\varepsilon_l^2  +320I^2\eta_l^2L^2\varepsilon_l^2 + \frac{64I\eta_g\eta_lL\varepsilon_l^2(1+40I^2\eta_l^2L^2)}{(1-\beta_1)^2}\frac{N-S}{S(N-1)}\\
        &   + \frac{20 \eta_g L \varepsilon_g^2}{(1-\beta_1)^2(1-\beta_2)I\eta_l} + \frac{8 \varepsilon_g^2}{(1-\beta_2)I^2\eta_l^2}  
         + 320I^2\eta_l^2L^2\rho^2 + \frac{64I\eta_g\eta_lL\rho^2(1+40I^2\eta_l^2L^2)}{(1-\beta_1)^2}\frac{N-S}{S(N-1)},
\end{align*}
where the expectation $\mathbbm{E}$ is w.r.t the sampled active workers per communication round.
\end{lemm}

\begin{proof}
    Based on $L$-smooth of $f$ and expectation w.r.t. the sampled active workers per communication  round, we have: 
    \begin{align}
        \label{Lemm_6_1:}
        \mathbbm{E}[f(\bm{u}_{t+1})] 
        & \overset{}{\leq} \mathbbm{E}[f(\bm{u}_{t})]  + \mathbbm{E}\left[\left\langle\nabla f(\bm{u}_{t}), \bm{u}_{t+1}-\bm{u}_{t}\right\rangle\right] + \frac{L}{2}\mathbbm{E}\left[\left\|\bm{u}_{t+1}-\bm{u}_{t}\right\|^{2}\right] \notag \\
         & \overset{(a)}{=} \mathbbm{E}[f(\bm{u}_{t})] -\frac{\eta_g}{1-\beta_1}\mathbbm{E}\left[\left\langle\nabla f(\bm{u}_{t}),  \tilde{\bm{d}}_{t+1}\right\rangle\right] + \frac{L\eta_g^2}{2(1-\beta_1)^2}\mathbbm{E}\left[\left\|\tilde{\bm{d}}_{t+1}\right\|^{2}\right] \notag \\
         & \overset{}{=} \mathbbm{E}[f(\bm{u}_{t})]  \underbrace{-\frac{\eta_g}{1-\beta_1}\mathbbm{E}\left[\left\langle\nabla f(\bm{u}_{t})-\nabla f(\bm{x}_{t}),  \tilde{\bm{d}}_{t+1}\right\rangle\right]}_{T_1} \underbrace{-\frac{\eta_g}{1-\beta_1}\mathbbm{E}\left[\left\langle\nabla f(\bm{x}_{t}),  \tilde{\bm{d}}_{t+1}\right\rangle\right]}_{T_2} + \underbrace{\frac{L\eta_g^2}{2(1-\beta_1)^2} \mathbbm{E}\left[\left\|\tilde{\bm{d}}_{t+1}\right\|^{2}\right]}_{T_3},
    \end{align}
    where ($a$) holds  because of the statement of Lemma~\ref{Lemm_1:}.

    Firstly, we note that
    \begin{align}
        T_1 & = -\frac{\eta_g}{1-\beta_1}\mathbbm{E}\left[\left\langle\nabla f(\bm{u}_{t})-\nabla f(\bm{x}_{t}),  \tilde{\bm{d}}_{t+1}\right\rangle\right] \notag \\
        & \overset{}{=} 
            -\frac{\eta_g}{1-\beta_1}\mathbbm{E}\left[\left\langle\nabla f(\bm{u}_{t})-\nabla f(\bm{x}_{t}),  \frac{1}{S}\sum_{i\in \mathcal{S}_t}\bm{d}_{t+1}^{(i)}\right\rangle\right] -\frac{\eta_g}{1-\beta_1}\mathbbm{E}\left[\left\langle\nabla f(\bm{u}_{t})-\nabla f(\bm{x}_{t}),  \tilde{\bm{m}}_{t+1}-\bm{m}_{t+1}\right\rangle\right]\notag\\
        & \overset{}{=} 
            \underbrace{-\frac{\eta_g}{1-\beta_1} \mathbbm{E}\left[\left\langle\nabla f(\bm{u}_{t})-\nabla f(\bm{x}_{t}),  \frac{1}{N} \sum_{i\in [N]}\bm{d}_{t+1}^{(i)}\right\rangle\right]}_{T_{1, 1}} \underbrace{-\frac{\eta_g}{1-\beta_1}\mathbbm{E}\left[\left\langle\nabla f(\bm{u}_{t})-\nabla f(\bm{x}_{t}),  \tilde{\bm{m}}_{t+1}-\bm{m}_{t+1}\right\rangle\right]}_{T_{1, 2}}. \notag
    \end{align}
    
    We proceed by analysis $T_{1, 1}$,
    \begin{align}
        T_{1, 1} 
        & \overset{}{=} 
            -\frac{\eta_g}{1-\beta_1}\mathbbm{E}\left[\left\langle\nabla f(\bm{u}_{t})-\nabla f(\bm{x}_{t}),  \frac{1}{N}\sum_{i\in [N]}\bm{d}_{t+1}^{(i)}\right\rangle\right]\notag\\
        & \overset{(a)}{=} 
            -\frac{\eta_g}{1-\beta_1}\mathbbm{E}\left[\left\langle\nabla f(\bm{u}_{t})-\nabla f(\bm{x}_{t}), \frac{\eta_l}{N}\sum_{i\in [N]} \sum_{\tau=0}^{I-1} \tilde{\bm{g}}_{t, \tau}^{(i)}\right\rangle\right]\notag\\
        & \overset{(b)}{\leq} 
            \frac{1-\beta_1}{2\beta_1 L}\mathbbm{E}\left[\|\nabla f(\bm{u}_t)- \nabla f(\bm{x}_t)\|^2\right] + \frac{\beta_1 L \eta_g^2\eta_l^2}{2(1-\beta_1)^3}\mathbbm{E}\left[\left\|\frac{1}{N}\sum_{i\in [N]}\sum_{\tau=0}^{I-1} \tilde{\bm{g}}_{t, \tau}^{(i)} \right\|^2\right]\notag\\
        & \overset{}{\leq} 
            \frac{(1-\beta_1)L}{2\beta_1 }\mathbbm{E}\left[\|\bm{u}_t- \bm{x}_t\|^2\right] + \frac{\beta_1 L \eta_g^2\eta_l^2}{2(1-\beta_1)^3N^2}\mathbbm{E}\left[\left\|\sum_{i\in [N]}\sum_{\tau=0}^{I-1} \tilde{\bm{g}}_{t, \tau}^{(i)}\right\|^2\right], \notag
    \end{align}
    where ($a$) follows by using Eq.~(\ref{cal_d:}), and ($b$) holds because of the fact that $\pm\langle\bm{a}, \bm{b}\rangle \leq \frac{1}{2}\|\bm{a}\|^2+\frac{1}{2}\|\bm{a}\|^2~ (\bm{a}, \bm{b} \in \mathbbm{R}^d)$ where $\bm{a}=-\frac{\sqrt{1-\beta_1}}{\sqrt{\beta_1 L}}(\nabla f(\bm{u}_{t})-\nabla f(\bm{x}_{t}))$ and $\bm{b}=\frac{\sqrt{\beta_1 L}\eta_g\eta_l}{(1-\beta_1)^{3/2}}\frac{1}{N}\sum_{i\in [N]}\sum_{\tau=0}^{I-1} \tilde{\bm{g}}_{t, \tau}^{(i)}$. And we proceed by analysis $T_{1, 2}$,
    
    \begin{align}
        T_{1, 2} 
        & \overset{}{=}  
            -\frac{\eta_g}{1-\beta_1}\mathbbm{E}\left[\left\langle\nabla f(\bm{u}_{t})-\nabla f(\bm{x}_{t}),  \tilde{\bm{m}}_{t+1}-\bm{m}_{t+1}\right\rangle\right] \notag\\
        & \overset{(a)}{\leq} 
            \frac{1-\beta_1}{2\beta_1 L}\mathbbm{E}\left[\left\|\nabla f(\bm{u}_{t})-\nabla f(\bm{x}_{t})\right\|^2\right] + \frac{\beta_1 L \eta_g^2}{2(1-\beta_1)^3}\mathbbm{E}\left[\left\|\tilde{\bm{m}}_{t+1}-\bm{m}_{t+1}\right\|^2\right]\notag\\
        & \overset{}{\leq} 
            \frac{(1-\beta_1)L}{2\beta_1 }\mathbbm{E}\left[\left\|\bm{u}_{t}-\bm{x}_{t}\right\|^2\right] + \frac{\beta_1 L \eta_g^2 \varepsilon_g^2}{2(1-\beta_1)^3(1-\beta_2)}, \notag
    \end{align}
    where ($a$) results from the fact that $\pm\langle\bm{a}, \bm{b}\rangle \leq \frac{1}{2}\|\bm{a}\|^2+\frac{1}{2}\|\bm{b}\|^2~ (\bm{a}, \bm{b} \in \mathbbm{R}^d)$ where $\bm{a}=-\frac{\sqrt{1-\beta_1}}{\sqrt{\beta_1 L}}(\nabla f(\bm{u}_{t})-\nabla f(\bm{x}_{t}))$ and $\bm{b}=\frac{\sqrt{\beta_1 L}\eta_g}{(1-\beta_1)^{3/2}}\left(\tilde{\bm{m}}_{t+1}-\bm{m}_{t+1}\right)$. 
    
    
    Secondly, we observe that
    \begin{align}
        T_2 
        & \overset{}{=}  
            -\frac{\eta_g}{1-\beta_1}\mathbbm{E}\left[\left\langle\nabla f(\bm{x}_{t}),  \tilde{\bm{d}}_{t+1}\right\rangle\right]\notag\\
        & \overset{}{=}  
            \underbrace{-\frac{\eta_g}{1-\beta_1}\mathbbm{E}\left[\left\langle\nabla f(\bm{x}_{t}),  \bm{d}_{t+1}\right\rangle\right]}_{T_{2, 1}}\underbrace{-\frac{\eta_g}{1-\beta_1}\mathbbm{E}\left[\left\langle\nabla f(\bm{x}_{t}),  \tilde{\bm{m}}_{t+1}-\bm{m}_{t+1}\right\rangle\right]}_{T_{2, 2}}. \notag
    \end{align}
    
    We proceed by analysis $T_{2, 1}$,
    \begin{align}
        T_{2, 1}
        & \overset{}{=}  
            -\frac{\eta_g}{1-\beta_1}\mathbbm{E}\left[\left\langle\nabla f(\bm{x}_{t}),  \bm{d}_{t+1}\right\rangle\right]\notag\\
        & \overset{}{=}  
            -\frac{\eta_g}{1-\beta_1}\mathbbm{E}\left[\left\langle\nabla f(\bm{x}_{t}),  \frac{\eta_l}{N}\sum_{i\in [N]} \sum_{\tau=0}^{I-1} \tilde{\bm{g}}_{t, \tau}^{(i)}\right\rangle\right]\notag\\
        & \overset{}{=}  
            -\frac{\eta_g}{1-\beta_1}\mathbbm{E}\left[\left\langle\nabla f(\bm{x}_{t}),  \frac{\eta_l}{N}\sum_{i\in [N]} \sum_{\tau=0}^{I-1}  \tilde{\bm{g}}_{t, \tau}^{(i)} + \eta_l I \nabla f(\bm{x}_{t}) -\eta_l I \nabla f(\bm{x}_{t})\right\rangle\right]\notag\\
        & \overset{}{=}  
            -\frac{I\eta_g\eta_l}{1-\beta_1}\mathbbm{E}\left[\left\|\nabla f(\bm{x}_{t})\right\|^2\right]+\frac{\eta_g}{1-\beta_1}\mathbbm{E}\left[\left\langle-\sqrt{\eta_l I}\nabla f(\bm{x}_{t}),  \frac{\sqrt{\eta_l}}{\sqrt{I}N}\sum_{i\in [N]} \sum_{\tau=0}^{I-1} \left(\tilde{\bm{g}}_{t, \tau}^{(i)}-\nabla f_i(\bm{x}_{t})\right)\right\rangle\right]\notag\\
        & \overset{(a)}{=}  
            -\frac{I\eta_g\eta_l}{1-\beta_1}\mathbbm{E}\left[\left\|\nabla f(\bm{x}_{t})\right\|^2\right]+\frac{\eta_g}{1-\beta_1}\mathbbm{E}\bigg[\frac{\eta_l I}{2}\|\nabla f(\bm{x}_{t})\|^2 + \frac{\eta_l}{2IN^2}\left\|\sum_{i\in [N]} \sum_{\tau=0}^{I-1} \left(\tilde{\bm{g}}_{t, \tau}^{(i)}-\nabla f_i(\bm{x}_{t})\right)\right\|^2 \notag\\
            & \quad \quad \quad \quad \quad \quad \quad \quad \quad \quad \quad \quad \quad \quad \quad \quad \quad \quad \quad \quad \quad \quad \quad \quad \quad \quad \quad \quad \quad \quad \quad - \frac{\eta_l}{2IN^2}
            \left\|\sum_{i\in [N]} \sum_{\tau=0}^{I-1} \tilde{\bm{g}}_{t, \tau}^{(i)}\right\|^2\bigg]\notag\\
        & \overset{}{=}  
            -\frac{I\eta_g\eta_l}{2(1-\beta_1)}\mathbbm{E}\left[\left\|\nabla f(\bm{x}_{t})\right\|^2\right]+\frac{\eta_g\eta_l}{2I(1-\beta_1)N^2}\mathbbm{E}\left\|\sum_{i\in [N]} \sum_{\tau=0}^{I-1} \left(\tilde{\bm{g}}_{t, \tau}^{(i)}-\bm{g}_{t, \tau}^{(i)}+\nabla f_i(\bm{x}_{t, \tau}^{(i)})-\nabla f_i(\bm{x}_{t})\right)\right\|^2\notag\\
            & \quad \quad \quad \quad \quad \quad \quad \quad \quad \quad \quad \quad \quad \quad \quad \quad \quad \quad \quad \quad \quad \quad \quad \quad \quad \quad \quad \quad - \frac{\eta_g\eta_l}{2I(1-\beta_1)N^2}
            \mathbbm{E}\left\|\sum_{i\in [N]} \sum_{\tau=0}^{I-1} \tilde{\bm{g}}_{t, \tau}^{(i)}\right\|^2\notag\\
        & \overset{(b)}{\leq}  
            -\frac{I\eta_g\eta_l}{2(1-\beta_1)}\mathbbm{E}\left[\left\|\nabla f(\bm{x}_{t})\right\|^2\right]+\frac{\eta_g\eta_l}{(1-\beta_1)N}\sum_{i\in [N]} \sum_{\tau=0}^{I-1}\left[\mathbbm{E}\left\|\tilde{\bm{g}}_{t, \tau}^{(i)}-\bm{g}_{t, \tau}^{(i)}\right\|^2+\mathbbm{E}\left\|\nabla f_i(\bm{x}_{t, \tau}^{(i)})-\nabla f_i(\bm{x}_{t})\right\|^2\right]\notag\\
            & \quad \quad \quad \quad \quad \quad \quad \quad \quad \quad \quad \quad \quad \quad \quad \quad \quad \quad \quad \quad \quad \quad \quad \quad \quad \quad \quad \quad - \frac{\eta_g\eta_l}{2I(1-\beta_1)N^2}
            \mathbbm{E}\left\|\sum_{i\in [N]} \sum_{\tau=0}^{I-1}\tilde{\bm{g}}_{t, \tau}^{(i)}\right\|^2\notag\\
        & \overset{}{\leq}  
            -\frac{I\eta_g\eta_l}{2(1-\beta_1)}\mathbbm{E}\left[\left\|\nabla f(\bm{x}_{t})\right\|^2\right]+\frac{I\eta_g\eta_l\varepsilon_l^2}{(1-\beta_1)} +\frac{\eta_g\eta_lL^2}{(1-\beta_1)N}\sum_{i\in [N]} \sum_{\tau=0}^{I-1}\mathbbm{E}\left[\left\|\bm{x}_{t, \tau}^{(i)}-\bm{x}_{t}\right\|^2\right]\notag\\
            & \quad \quad \quad \quad \quad \quad \quad \quad \quad \quad \quad \quad \quad \quad \quad \quad \quad \quad \quad \quad \quad \quad \quad \quad \quad \quad \quad \quad  - \frac{\eta_g\eta_l}{2I(1-\beta_1)N^2}
            \mathbbm{E}\left\|\sum_{i\in [N]} \sum_{\tau=0}^{I-1}\tilde{\bm{g}}_{t, \tau}^{(i)}\right\|^2, \notag
    \end{align}
    where ($a$) follows from the fact that $\pm\langle\bm{a}, \bm{b} \rangle=\frac{1}{2}(\|\bm{a}\|^2+\|\bm{b}\|^2-\|\bm{a}-\bm{b}\|^2)$, and ($b$) uses the inequality $\left\|\sum_{i=1}^{N}\bm{x}_i\right\|^2 \leq N\sum_{i=1}^{N}\|\bm{x}_i\|^2, \bm{x}_i \in \mathbbm{R}^d$.
    And we proceed by analysis $T_{2, 2}$,
    \begin{align}
        T_{2, 2}
        & \overset{}{=}  
            -\frac{\eta_g}{1-\beta_1}\mathbbm{E}\left[\left\langle\nabla f(\bm{x}_{t}),  \tilde{\bm{m}}_{t+1}-\bm{m}_{t+1}\right\rangle\right]\notag\\
        & \overset{(a)}{\leq}  
            \frac{I\eta_g \eta_l}{4(1-\beta_1)}\mathbbm{E}\left[\|\nabla f(\bm{x}_{t})\|^2\right] +\frac{\eta_g}{(1-\beta_1)I\eta_l}\mathbbm{E} \left[ \|\tilde{\bm{m}}_{t+1}-\bm{m}_{t+1}\|^2\right]\notag\\
        & \overset{}{\leq}  
            \frac{I\eta_g \eta_l}{4(1-\beta_1)}\mathbbm{E}\left[\|\nabla f(\bm{x}_{t})\|^2\right] +\frac{\eta_g \varepsilon_g^2}{(1-\beta_1)(1-\beta_2)I\eta_l}, \notag
    \end{align}
    where ($a$) using the fact that $\pm\langle\bm{a}, \bm{b}\rangle \leq \frac{1}{2}\|\bm{a}\|^2+\frac{1}{2}\|\bm{b}\|^2~ (\bm{a}, \bm{b} \in \mathbbm{R}^d)$ where $\bm{a}=-\frac{\sqrt{I\eta_l}}{\sqrt{2}}\nabla f(\bm{x}_{t})$ and $\bm{b}=\frac{\sqrt{2}}{\sqrt{I\eta_l}}\left(\tilde{\bm{m}}_{t+1}-\bm{m}_{t+1}\right)$.

    Next, we utilize the statement of Lemma~\ref{Lemm_4:} to derive the following upper bound of $T_3$,
    
    \begin{align*}
        T_3 
        & \overset{}{=}  
            \frac{L\eta_g^2}{2(1-\beta_1)^2}\mathbbm{E}\left[\left\|\tilde{\bm{d}}_{t+1}\right\|^{2}\right] \overset{}{\leq}  
             \frac{L\eta_g^2\eta_l^2}{(1-\beta_1)^2S^2}\mathbbm{E}\left[\left\|\sum_{i\in [N]}\mathbbm{I}\{i\in \mathcal{S}_t\}\sum_{\tau=0}^{I-1}\tilde{\bm{g}}_{t, \tau}^{(i)}\right\|^2\right]+\frac{L\eta_g^2\varepsilon_g^2}{(1-\beta_1)^2(1-\beta_2)}. 
    \end{align*}

    Substituting the upper bounds of $T_{1, 1}$, $T_{1, 2}$ into $T_1$ and $T_{2, 1}$, $T_{2, 2}$ into $T_2$ and $T_1$, $T_2$, $T_3$ into (\ref{Lemm_6_1:}) yields:
    
    \begin{align}
        & \mathbbm{E}[f(\bm{u}_{t+1})] \notag\\
        & \leq \mathbbm{E}[f(\bm{u}_{t})]  \underbrace{-\frac{\eta_g}{1-\beta_1}\mathbbm{E}\left[\left\langle\nabla f(\bm{u}_{t})-\nabla f(\bm{x}_{t}),  \tilde{\bm{d}}_{t+1}\right\rangle\right]}_{T_1} \underbrace{-\frac{\eta_g}{1-\beta_1}\mathbbm{E}\left[\left\langle\nabla f(\bm{x}_{t}),  \tilde{\bm{d}}_{t+1}\right\rangle\right]}_{T_2} + \underbrace{\frac{L\eta_g^2}{2(1-\beta_1)^2} \mathbbm{E}\left[\left\|\tilde{\bm{d}}_{t+1}\right\|^{2}\right]}_{T_3} \notag \\
        & \overset{}{\leq} 
            \mathbbm{E}[f(\bm{u}_{t})] -\frac{I\eta_g\eta_l}{(1-\beta_1)}\left(\frac{1}{4}-40I^2\eta_l^2L^2\right)\mathbbm{E}\left[\left\|\nabla f(\bm{x}_{t})\right\|^2\right] + \frac{(1-\beta_1)L}{\beta_1 }\mathbbm{E}\left[\|\bm{u}_t- \bm{x}_t\|^2\right] + \frac{\beta_1 L \eta_g^2 \varepsilon_g^2}{2(1-\beta_1)^3(1-\beta_2)} \notag\\
            & +\frac{I\eta_g\eta_l\varepsilon_l^2}{(1-\beta_1)} +\frac{\eta_g \varepsilon_g^2}{(1-\beta_1)(1-\beta_2)I\eta_l} +\frac{40I^3\eta_g\eta_l^3L^2\varepsilon_l^2}{(1-\beta_1)} + \frac{40I^3\eta_g\eta_l^3L^2\rho^2}{(1-\beta_1)} +\frac{L\eta_g^2\varepsilon_g^2}{(1-\beta_1)^2(1-\beta_2)} \notag\\
            &  - \frac{\eta_g\eta_l}{2I(1-\beta_1)N^2}\left(1
             - \frac{\beta_1 I \eta_g\eta_lL}{(1-\beta_1)^2}\right)\mathbbm{E}\left[\left\|\sum_{i\in [N]}\sum_{\tau=0}^{I-1} \tilde{\bm{g}}_{t, \tau}^{(i)}\right\|^2\right]  + \frac{L\eta_g^2\eta_l^2}{(1-\beta_1)^2S^2}\mathbbm{E}\left[\left\|\sum_{i\in [N]}\mathbbm{I}\{i\in \mathcal{S}_t\}\sum_{\tau=0}^{I-1}\tilde{\bm{g}}_{t, \tau}^{(i)}\right\|^2\right]. \label{Lemm_6_2:}
    \end{align}
    
    Using the inequality~(\ref{Lemm_6_2:}), making a simple arrangement and doing the summation operation from $t=0$ to $T-1$, we get:
    \begin{align}
        & \mathbbm{E}[f(\bm{u}_{T})] -\mathbbm{E}[f(\bm{u}_{0})] \notag\\
        & \overset{}{\leq} 
            -\frac{I\eta_g\eta_l}{(1-\beta_1)}\left(\frac{1}{4}-40I^2\eta_l^2L^2\right)\sum_{t=0}^{T-1}\mathbbm{E}\left[\left\|\nabla f(\bm{x}_{t})\right\|^2\right] + \frac{(1-\beta_1)L}{\beta_1 }\sum_{t=0}^{T-1}\mathbbm{E}\left[\|\bm{u}_t- \bm{x}_t\|^2\right] + \frac{\beta_1 L \eta_g^2 \varepsilon_g^2T}{2(1-\beta_1)^3(1-\beta_2)} \notag\\
            & \quad \quad \quad \quad \quad \quad \quad \quad +\frac{I\eta_g\eta_l\varepsilon_l^2T}{(1-\beta_1)} +\frac{\eta_g \varepsilon_g^2T}{(1-\beta_1)(1-\beta_2)I\eta_l} +\frac{40I^3\eta_g\eta_l^3L^2\varepsilon_l^2T}{(1-\beta_1)} + \frac{40I^3\eta_g\eta_l^3L^2\rho^2T}{(1-\beta_1)} +\frac{L\eta_g^2\varepsilon_g^2T}{(1-\beta_1)^2(1-\beta_2)} \notag\\
            & \quad  - \frac{\eta_g\eta_l}{2I(1-\beta_1)N^2}\left(1
             - \frac{\beta_1 I \eta_g\eta_lL}{(1-\beta_1)^2}\right)\sum_{t=0}^{T-1}\mathbbm{E}\left[\left\|\sum_{i\in [N]}\sum_{\tau=0}^{I-1} \tilde{\bm{g}}_{t, \tau}^{(i)}\right\|^2\right]  + \frac{L\eta_g^2\eta_l^2}{(1-\beta_1)^2S^2}\sum_{t=0}^{T-1}\mathbbm{E}\left[\left\|\sum_{i\in [N]}\mathbbm{I}\{i\in \mathcal{S}_t\}\sum_{\tau=0}^{I-1}\tilde{\bm{g}}_{t, \tau}^{(i)}\right\|^2\right] \notag\\
        & \overset{(a)}{\leq} 
            -\frac{I\eta_g\eta_l}{(1-\beta_1)}\left(\frac{1}{4}-40I^2\eta_l^2L^2\right)\sum_{t=0}^{T-1}\mathbbm{E}\left[\left\|\nabla f(\bm{x}_{t})\right\|^2\right] + \frac{5\beta_1  \eta_g^2 L \varepsilon_g^2T}{2(1-\beta_1)^3(1-\beta_2)} \notag\\
            & \quad \quad \quad \quad \quad \quad \quad \quad +\frac{I\eta_g\eta_l\varepsilon_l^2T}{(1-\beta_1)} +\frac{\eta_g \varepsilon_g^2T}{(1-\beta_1)(1-\beta_2)I\eta_l} +\frac{40I^3\eta_g\eta_l^3L^2\varepsilon_l^2T}{(1-\beta_1)} + \frac{40I^3\eta_g\eta_l^3L^2\rho^2T}{(1-\beta_1)} +\frac{\eta_g^2L\varepsilon_g^2T}{(1-\beta_1)^2(1-\beta_2)} \notag\\
            & \quad  \underbrace{- \frac{\eta_g\eta_l}{2I(1-\beta_1)N^2}\left(1
             - \frac{\beta_1 I \eta_g\eta_lL}{(1-\beta_1)^2}\right)\sum_{t=0}^{T-1}\mathbbm{E}\left[\left\|\sum_{i\in [N]}\sum_{\tau=0}^{I-1} \tilde{\bm{g}}_{t, \tau}^{(i)}\right\|^2\right]  +  \frac{2\eta_g^2\eta_l^2L}{(1-\beta_1)^3S^2}\sum_{t=0}^{T-1}\mathbbm{E}\left[\left\|\sum_{i\in [N]}\mathbbm{I}\{i\in \mathcal{S}_t\}\sum_{\tau=0}^{I-1}\tilde{\bm{g}}_{t, \tau}^{(i)}\right\|^2\right]}_{T_4}, \label{Lemm_6_3:}
    \end{align}
     where ($a$) holds by using the statement of Lemma~\ref{Lemm_5:}. Next, we derive the upper bound for $T_4$.
    To simplify the proof process, we set  $\bm{q}_{t,i}=\sum_{\tau=0}^{I-1}\tilde{\bm{g}}_{t, \tau}^{(i)}$ yields:
    
    \begin{align}
         \mathbbm{E}\left[\left\|\sum_{i\in [N]}\sum_{\tau=0}^{I-1} \tilde{\bm{g}}_{t, \tau}^{(i)}\right\|^2\right] & \overset{}{=} 
            \mathbbm{E}\left[\left\|\sum_{i\in [N]}\bm{q}_{t,i}\right\|^2\right] \notag\\
        & \overset{}{=} 
           \sum_{i\in [N]} \mathbbm{E}\left[\left\|\bm{q}_{t,i}\right\|^2\right]+\sum_{i \neq j} \mathbbm{E}\left[\langle \bm{q}_{t,i}, \bm{q}_{t,j} \rangle\right] \notag\\
        & \overset{}{=} 
           \sum_{i\in [N]} N\mathbbm{E}\left[\left\|\bm{q}_{t,i}\right\|^2\right]-\frac{1}{2}\sum_{i \neq j} \mathbbm{E}\left[\| \bm{q}_{t,i}- \bm{q}_{t,j} \|^2\right], \label{Lemm_6_4:}
    \end{align}
    and
    \begin{align}
        \mathbbm{E}\left[\left\|\sum_{i\in [N]}\mathbbm{I}\{i\in \mathcal{S}_t\}\sum_{\tau=0}^{I-1}\tilde{\bm{g}}_{t, \tau}^{(i)}\right\|^2\right]  & \overset{}{=} 
            \mathbbm{E}\left[\left\|\sum_{i\in [N]}\mathbbm{P}\{i\in \mathcal{S}_t\}\bm{q}_{t,i}\right\|^2\right] \notag\\
        & \overset{}{=} 
            \sum_{i\in [N]}\mathbbm{P}\{i\in \mathcal{S}_t\}\mathbbm{E}\left[\left\|\bm{q}_{t,i}\right\|^2\right] + \sum_{i \neq j} \mathbbm{P}\{i, j\in \mathcal{S}_t\} \mathbbm{E}\left[\langle \bm{q}_{t,i}, \bm{q}_{t,j} \rangle\right]\notag\\
        & \overset{}{=} 
            \frac{S}{N}\sum_{i\in [N]}\mathbbm{E}\left[\left\|\bm{q}_{t,i}\right\|^2\right] + \frac{S(S-1)}{N(N-1)}\sum_{i \neq j} \mathbbm{E}\left[\langle \bm{q}_{t,i}, \bm{q}_{t,j} \rangle\right]\notag\\
        & \overset{}{=} 
            \frac{S^2}{N}\sum_{i\in [N]}\mathbbm{E}\left[\left\|\bm{q}_{t,i}\right\|^2\right] - \frac{S(S-1)}{2N(N-1)}\sum_{i \neq j} \mathbbm{E}\left[\| \bm{q}_{t,i} - \bm{q}_{t,j} \|^2\right], \label{Lemm_6_5:}
    \end{align}
    where $\mathbbm{P}\{i\in \mathcal{S}_t\}=\frac{S}{N}$. 
    
    Substituting equalities~(\ref{Lemm_6_4:}) and~(\ref{Lemm_6_5:}) into $T_4$ yields:
    \begin{align}
         T_4 & \overset{}{=} 
              - \frac{\eta_g\eta_l}{2I(1-\beta_1)N^2}\left(1
             - \frac{\beta_1 I \eta_g\eta_lL}{(1-\beta_1)^2}\right)\sum_{t=0}^{T-1}\mathbbm{E}\left[\left\|\sum_{i\in [N]}\sum_{\tau=0}^{I-1} \tilde{\bm{g}}_{t, \tau}^{(i)}\right\|^2\right]  +  \frac{2\eta_g^2\eta_l^2L}{(1-\beta_1)^3S^2}\sum_{t=0}^{T-1}\mathbbm{E}\left[\left\|\sum_{i\in [N]}\mathbbm{I}\{i\in \mathcal{S}_t\}\sum_{\tau=0}^{I-1}\tilde{\bm{g}}_{t, \tau}^{(i)}\right\|^2\right] \notag\\
        & \overset{}{=} 
              - \frac{\eta_g\eta_l}{2I(1-\beta_1)N^2}\left(1
             - \frac{\beta_1 I \eta_g\eta_lL}{(1-\beta_1)^2}\right)\sum_{t=0}^{T-1}\left[\sum_{i\in [N]} N\mathbbm{E}\left[\left\|\bm{q}_{t,i}\right\|^2\right]-\frac{1}{2}\sum_{i \neq j} \mathbbm{E}\left[\| \bm{q}_{t,i}- \bm{q}_{t,j} \|^2\right]\right]  \notag\\
             & \quad \quad \quad \quad \quad \quad \quad \quad \quad \quad \quad \quad \quad \quad \quad +  \frac{2\eta_g^2\eta_l^2L}{(1-\beta_1)^3S^2}\sum_{t=0}^{T-1}\left[\frac{S^2}{N}\sum_{i\in [N]}\mathbbm{E}\left[\left\|\bm{q}_{t,i}\right\|^2\right] - \frac{S(S-1)}{2N(N-1)}\sum_{i \neq j} \mathbbm{E}\left[\| \bm{q}_{t,i} - \bm{q}_{t,j} \|^2\right]\right] \notag\\
        & \overset{}{=} 
              - \frac{\eta_g\eta_l}{2I(1-\beta_1)N}\left(1
             - \frac{\beta_1 I \eta_g\eta_lL}{(1-\beta_1)^2}-\frac{4I\eta_g\eta_lL}{(1-\beta_1)^2}\right)\sum_{t=0}^{T-1}\sum_{i\in [N]} \mathbbm{E}\left[\left\|\bm{q}_{t,i}\right\|^2\right]  \notag\\
             & \quad \quad \quad \quad \quad \quad \quad \quad \quad \quad \quad \quad + \frac{\eta_g\eta_l}{4I(1-\beta_1)N^2}\left(1
             - \frac{\beta_1 I \eta_g\eta_lL}{(1-\beta_1)^2}-\frac{4I\eta_g\eta_lL}{(1-\beta_1)^2}\frac{N(S-1)}{S(N-1)}\right)\sum_{t=0}^{T-1}\sum_{i \neq j} \mathbbm{E}\left[\| \bm{q}_{t,i}- \bm{q}_{t,j} \|^2\right] \notag\\
        & \overset{(a)}{=} 
              - \frac{\eta_g\eta_l}{2I(1-\beta_1)N}\left(1
             - \frac{\beta_1 I \eta_g\eta_lL}{(1-\beta_1)^2}-\frac{4I\eta_g\eta_lL}{(1-\beta_1)^2}\right)\sum_{t=0}^{T-1}\sum_{i\in [N]} \mathbbm{E}\left[\left\|\bm{q}_{t,i}\right\|^2\right]  \notag\\
             & \quad \quad + \frac{\eta_g\eta_l}{4I(1-\beta_1)N^2}\left(1
             - \frac{\beta_1 I \eta_g\eta_lL}{(1-\beta_1)^2}-\frac{4I\eta_g\eta_lL}{(1-\beta_1)^2}\frac{N(S-1)}{S(N-1)}\right)\sum_{t=0}^{T-1}\left[2N\sum_{i\in [N]}\mathbbm{E}\left[\|\bm{q}_{t,i}\|^2\right]-2\mathbbm{E}\left[\|\sum_{i\in[N]}\bm{q}_{t,i}\|^2\right]\right] \notag\\
        & \overset{}{=} 
               \frac{2\eta_g^2\eta_l^2L}{N(1-\beta_1)^3}\left(1-\frac{N(S-1)}{S(N-1)}\right)\sum_{t=0}^{T-1}\sum_{i\in [N]} \mathbbm{E}\left[\left\|\bm{q}_{t,i}\right\|^2\right]  \notag\\
             & \quad \quad \quad \quad \quad \quad \quad \quad \quad \quad \quad \quad \quad \quad - \frac{\eta_g\eta_l}{2I(1-\beta_1)N^2}\left(1
             - \frac{\beta_1 I \eta_g\eta_lL}{(1-\beta_1)^2}-\frac{4I\eta_g\eta_lL}{(1-\beta_1)^2}\frac{N(S-1)}{S(N-1)}\right)\sum_{t=0}^{T-1}\mathbbm{E}\left[\|\sum_{i\in[N]}\bm{q}_{t,i}\|^2\right]\notag\\
        & \overset{}{=} 
               \frac{2\eta_g^2\eta_l^2L}{(1-\beta_1)^3}\frac{N-S}{NS(N-1)}\sum_{t=0}^{T-1}\sum_{i\in [N]} \mathbbm{E}\left[\left\|\bm{q}_{t,i}\right\|^2\right]  \notag\\
             & \quad \quad \quad \quad \quad \quad \quad \quad \quad \quad \quad \quad \quad \quad - \frac{\eta_g\eta_l}{2I(1-\beta_1)N^2}\left(1
             - \frac{\beta_1 I \eta_g\eta_lL}{(1-\beta_1)^2}-\frac{4I\eta_g\eta_lL}{(1-\beta_1)^2}\frac{N(S-1)}{S(N-1)}\right)\sum_{t=0}^{T-1}\mathbbm{E}\left[\|\sum_{i\in[N]}\bm{q}_{t,i}\|^2\right] \notag \\
        & \overset{(b)}{\leq} 
               \frac{2\eta_g^2\eta_l^2L}{(1-\beta_1)^3}\frac{N-S}{NS(N-1)}\sum_{t=0}^{T-1}\sum_{i\in [N]} \mathbbm{E}\left[\left\|\bm{q}_{t,i}\right\|^2\right] \notag \\
        & \overset{(c)}{\leq} 
               \frac{2\eta_g^2\eta_l^2L}{(1-\beta_1)^3}\frac{N-S}{S(N-1)} \left[4I^2(\varepsilon_l^2+\rho^2)(1+40I^2\eta_l^2L^2)T + 4I^2(1+40I^2\eta_l^2L^2)\sum_{t=0}^{T-1}\mathbbm{E}\left[\left\|\nabla f(\bm{x}_{t})\right\|^2\right]\right] \notag\\
        & \overset{}{=} 
                \frac{8I^2\eta_g^2\eta_l^2L(\varepsilon_l^2+\rho^2)(1+40I^2\eta_l^2L^2)T}{(1-\beta_1)^3}\frac{N-S}{S(N-1)} + \frac{8I^2\eta_g^2\eta_l^2L(1+40I^2\eta_l^2L^2)}{(1-\beta_1)^3}\frac{N-S}{S(N-1)}\sum_{t=0}^{T-1}\mathbbm{E}\left[\left\|\nabla f(\bm{x}_{t})\right\|^2\right], \notag
    \end{align}
    where ($a$) holds by using $\sum_{i \neq j}\mathbbm{E}\left[\| \bm{q}_{t,i}- \bm{q}_{t,j} \|^2\right]=2N\sum_{i\in [N]}\mathbbm{E}\left[\|\bm{q}_{t,i}\|^2\right]-2\mathbbm{E}\left[\|\sum_{i\in[N]}\bm{q}_{t,i}\|^2\right]$, ($b$) results from the fact that $1 - \frac{\beta_1 I \eta_g\eta_lL}{(1-\beta_1)^2}-\frac{4I\eta_g\eta_lL}{(1-\beta_1)^2}\frac{N(S-1)}{S(N-1)} \geq 0$ holds if $\eta_g\eta_l\leq\frac{(1-\beta_1)^2S(N-1)}{IL(\beta_1S(N-1)+4N(S-1))}$, and ($c$) follows from the statement of Lemma~\ref{Lemm_3:}. 
    
    Furthermore, substituting the upper bound of $T_4$ into~(\ref{Lemm_6_3:}), we get:
    \begin{align}
        & \mathbbm{E}[f(\bm{u}_{T})] -\mathbbm{E}[f(\bm{u}_{0})] \notag\\
        & \overset{}{\leq} 
            -\frac{I\eta_g\eta_l}{(1-\beta_1)}\left(\frac{1}{4}-40I^2\eta_l^2L^2\right)\sum_{t=0}^{T-1}\mathbbm{E}\left[\left\|\nabla f(\bm{x}_{t})\right\|^2\right] + \frac{5\beta_1  \eta_g^2 L \varepsilon_g^2T}{2(1-\beta_1)^3(1-\beta_2)} \notag\\
            & \quad \quad \quad \quad \quad \quad \quad \quad +\frac{I\eta_g\eta_l\varepsilon_l^2T}{(1-\beta_1)} +\frac{\eta_g \varepsilon_g^2T}{(1-\beta_1)(1-\beta_2)I\eta_l} +\frac{40I^3\eta_g\eta_l^3L^2\varepsilon_l^2T}{(1-\beta_1)} + \frac{40I^3\eta_g\eta_l^3L^2\rho^2T}{(1-\beta_1)} +\frac{\eta_g^2L\varepsilon_g^2T}{(1-\beta_1)^2(1-\beta_2)} \notag\\
            & \quad \underbrace{- \frac{\eta_g\eta_l}{2I(1-\beta_1)N^2}\left(1
             - \frac{\beta_1 I \eta_g\eta_lL}{(1-\beta_1)^2}\right)\sum_{t=0}^{T-1}\mathbbm{E}\left[\left\|\sum_{i\in [N]}\sum_{\tau=0}^{I-1} \tilde{\bm{g}}_{t, \tau}^{(i)}\right\|^2\right]  +  \frac{2\eta_g^2\eta_l^2L}{(1-\beta_1)^3S^2}\sum_{t=0}^{T-1}\mathbbm{E}\left[\left\|\sum_{i\in [N]}\mathbbm{I}\{i\in \mathcal{S}_t\}\sum_{\tau=0}^{I-1}\tilde{\bm{g}}_{t, \tau}^{(i)}\right\|^2\right]}_{T_4} \notag\\
        & \overset{}{\leq} 
            -\frac{I\eta_g\eta_l}{(1-\beta_1)}\left(\frac{1}{4}-40I^2\eta_l^2L^2\right)\sum_{t=0}^{T-1}\mathbbm{E}\left[\left\|\nabla f(\bm{x}_{t})\right\|^2\right] + \frac{5\beta_1  \eta_g^2 L \varepsilon_g^2T}{2(1-\beta_1)^3(1-\beta_2)} \notag\\
            & \quad \quad \quad \quad \quad \quad \quad \quad +\frac{I\eta_g\eta_l\varepsilon_l^2T}{(1-\beta_1)} +\frac{\eta_g \varepsilon_g^2T}{(1-\beta_1)(1-\beta_2)I\eta_l} +\frac{40I^3\eta_g\eta_l^3L^2\varepsilon_l^2T}{(1-\beta_1)} + \frac{40I^3\eta_g\eta_l^3L^2\rho^2T}{(1-\beta_1)} +\frac{\eta_g^2L\varepsilon_g^2T}{(1-\beta_1)^2(1-\beta_2)} \notag\\
            & \quad + \frac{8I^2\eta_g^2\eta_l^2L(\varepsilon_l^2+\rho^2)(1+40I^2\eta_l^2L^2)T}{(1-\beta_1)^3}\frac{N-S}{S(N-1)} + \frac{8I^2\eta_g^2\eta_l^2L(1+40I^2\eta_l^2L^2)}{(1-\beta_1)^3}\frac{N-S}{S(N-1)}\sum_{t=0}^{T-1}\mathbbm{E}\left[\left\|\nabla f(\bm{x}_{t})\right\|^2\right] \notag\\
       & \overset{}{=} 
            -\frac{I\eta_g\eta_l}{(1-\beta_1)}\left(\frac{1}{4}-40I^2\eta_l^2L^2-\frac{8I\eta_g\eta_lL(1+40I^2\eta_l^2L^2)}{(1-\beta_1)^2}\frac{N-S}{S(N-1)}\right)\sum_{t=0}^{T-1}\mathbbm{E}\left[\left\|\nabla f(\bm{x}_{t})\right\|^2\right] + \frac{5\beta_1  \eta_g^2 L \varepsilon_g^2T}{2(1-\beta_1)^3(1-\beta_2)} \notag\\
            & \quad \quad \quad \quad \quad \quad \quad \quad +\frac{I\eta_g\eta_l\varepsilon_l^2T}{(1-\beta_1)} +\frac{\eta_g \varepsilon_g^2T}{(1-\beta_1)(1-\beta_2)I\eta_l} +\frac{40I^3\eta_g\eta_l^3L^2\varepsilon_l^2T}{(1-\beta_1)} + \frac{40I^3\eta_g\eta_l^3L^2\rho^2T}{(1-\beta_1)} +\frac{\eta_g^2L\varepsilon_g^2T}{(1-\beta_1)^2(1-\beta_2)} \notag\\
            &  \quad \quad \quad \quad \quad \quad \quad \quad \quad \quad \quad \quad \quad \quad \quad \quad \quad \quad \quad \quad \quad \quad \quad \quad \quad \quad \quad + \frac{8I^2\eta_g^2\eta_l^2L(\varepsilon_l^2+\rho^2)(1+40I^2\eta_l^2L^2)T}{(1-\beta_1)^3}\frac{N-S}{S(N-1)}  \notag\\
        & \overset{(a)}{\leq} 
            -\frac{I\eta_g\eta_l}{8(1-\beta_1)}\sum_{t=0}^{T-1}\mathbbm{E}\left[\left\|\nabla f(\bm{x}_{t})\right\|^2\right] + \frac{5\beta_1  \eta_g^2 L \varepsilon_g^2T}{2(1-\beta_1)^3(1-\beta_2)} +\frac{I\eta_g\eta_l\varepsilon_l^2T}{(1-\beta_1)} +\frac{\eta_g \varepsilon_g^2T}{(1-\beta_1)(1-\beta_2)I\eta_l} +\frac{40I^3\eta_g\eta_l^3L^2\varepsilon_l^2T}{(1-\beta_1)} \notag\\
            &  \quad \quad \quad \quad \quad \quad \quad \quad \quad \quad + \frac{40I^3\eta_g\eta_l^3L^2\rho^2T}{(1-\beta_1)} +\frac{\eta_g^2L\varepsilon_g^2T}{(1-\beta_1)^2(1-\beta_2)} + \frac{8I^2\eta_g^2\eta_l^2L(\varepsilon_l^2+\rho^2)(1+40I^2\eta_l^2L^2)T}{(1-\beta_1)^3}\frac{N-S}{S(N-1)},  \label{Lemm_6_6:}
    \end{align}
    where ($a$) results from the fact that $\frac{1}{4}-40I^2\eta_l^2L^2-\frac{8I\eta_g\eta_lL(1+40I^2\eta_l^2L^2)}{(1-\beta_1)^2}\frac{N-S}{S(N-1)}\geq\frac{1}{8}$ holds if $ 320I^2\eta_l^2L^2+\frac{64I\eta_g\eta_lL(1+40I^2\eta_l^2L^2)}{(1-\beta_1)^2}\frac{N-S}{S(N-1)}\leq 1$. Now, we use the statement of Assumption~\ref{Global_Function_Below_Bounds:} yields:
    \begin{align}
        f^\star -\mathbbm{E}[f(\bm{x}_{0})] \overset{}{\leq} \mathbbm{E}[f(\bm{u}_{T})] -\mathbbm{E}[f(\bm{u}_{0})]. \label{Lemm_6_7:}
    \end{align}
    This holds as $\bm{u}_{0}=\bm{x}_{0}$. Finally, the proof of the lemma is completed by substituting inequality~(\ref{Lemm_6_7:}) into inequality~(\ref{Lemm_6_6:}) and making a simple arrangement.
\end{proof}

\end{document}